\documentclass{article}

\PassOptionsToPackage{numbers,compress}{natbib}
 \usepackage[preprint]{neurips_2026}

\usepackage[utf8]{inputenc} 
\usepackage{CJKutf8} 
\usepackage[T1]{fontenc}    
\usepackage{hyperref}       
\usepackage{url}            
\usepackage{booktabs}       
\usepackage{amsfonts}       
\usepackage{amsmath}        
\usepackage{amssymb}        
\usepackage{amsthm}         
\usepackage{enumitem}
\usepackage{nicefrac}       
\usepackage{microtype}      
\usepackage[table]{xcolor} 
\usepackage{graphicx}
\usepackage{tabularx}
\usepackage{algorithm}
\usepackage{multirow}
\usepackage{algpseudocode}
\usepackage{listings}

\definecolor{tablegroup}{HTML}{E7D9A8}
\definecolor{tableavg}{HTML}{D8E7E2}
\definecolor{tableours}{HTML}{DCE6EF}

\newcolumntype{Y}{>{\centering\arraybackslash}X}

\newcolumntype{V}{>{\columncolor{tableavg}\centering\arraybackslash}X}

\newcommand{\groupbar}[1]{%
    \multicolumn{10}{@{}>{\columncolor{tablegroup}}c@{}}{\textbf{#1}}%
}
\newcolumntype{L}[1]{>{\raggedright\arraybackslash}p{#1}}
\newcolumntype{C}[1]{>{\centering\arraybackslash}p{#1}}
\newcolumntype{A}[1]{>{\columncolor{tableavg}[\tabcolsep][\tabcolsep]}C{#1}}

\algrenewcommand\algorithmicrequire{\textbf{Require:}}
\algrenewcommand\algorithmicensure{\textbf{Ensure:}}
\algrenewcommand\algorithmiccomment[1]{\hfill$\triangleright$ #1}

\lstdefinestyle{promptstyle}{
  basicstyle=\ttfamily\scriptsize,
  breaklines=true,
  breakatwhitespace=false,
  columns=fullflexible,
  keepspaces=true,
  frame=single,
  xleftmargin=0.5em,
  framexleftmargin=0.5em,
  aboveskip=0.6em,
  belowskip=0.6em
}

\title{From Instance Selection to Fixed-Pool Data Recipe Search for Supervised Fine-Tuning}

\author{%
  Haodong Wu\textsuperscript{1}
  \quad Jiahao Zhang\textsuperscript{2}
  \quad Lijie Hu\textsuperscript{2}
  \quad Yongqi Zhang\textsuperscript{1}
  \\
  \textsuperscript{1}The Hong Kong University of Science and Technology (Guangzhou)
  \\
  \textsuperscript{2}Mohamed bin Zayed University of Artificial Intelligence
  \\
  \texttt{hwu315@connect.hkust-gz.edu.cn}
  \quad
  \texttt{yongqizhang@hkust-gz.edu.cn}
  \\
  \texttt{\{jiahao.zhang,lijie.hu\}@mbzuai.ac.ae}
}

\begin{document}

\maketitle

\begin{abstract}
Supervised fine-tuning (SFT) data selection is commonly formulated as instance ranking: score each example and retain a top-$k$ subset. 
However, effective SFT training subsets are often produced through ordered curation recipes, where filtering, mixing, and deduplication operators jointly shape the final data distribution. 
We formulate this problem as \emph{fixed-pool data recipe search}: given a raw instruction pool and a library of grounded operators, the goal is to discover an executable recipe that constructs a high-quality selected subset under a limited budget of full SFT evaluations, without generating, rewriting, or augmenting training samples. 
We introduce \textbf{AutoSelection}, a two-layer solver that decouples fixed-pool materialization based on cached task-, data-, and model-side signals from expensive full evaluation, using warmup probes, realized subset states, local recipe edits, Gaussian-process-assisted ranking, and stagnation-triggered reseeding. 
Experiments on a 90K instruction pool show that AutoSelection achieves the strongest in-distribution reasoning average across three base models, outperforming full-data training, random recipe search, random top-$k$, and single-operator selectors. 
Additional Out-of-distribution graph-reasoning results, search-stability analyses, structural ablations, and 1.5B-to-7B transfer checks further show that recipe structure matters beyond individual selection operators.
Code is available at \url{https://github.com/w253/AutoSelection}.

\end{abstract}

\section{Introduction}
\label{sec:intro}

\begin{figure*}[t]
    \centering
    \vspace{-14px}
    \includegraphics[width=\textwidth]{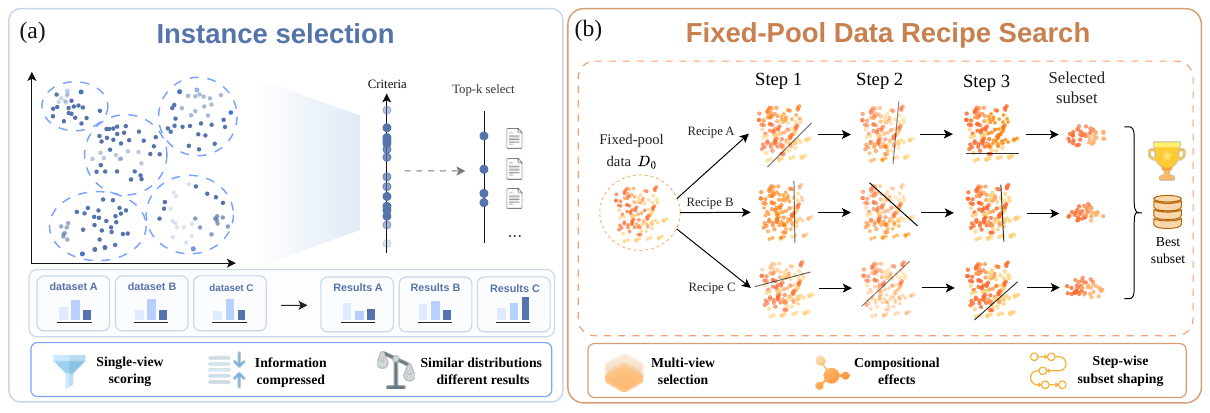}
    \vspace{-15px}
    \caption{Conceptual contrast between instance-level selection and fixed-pool data recipe search for SFT data curation.
    (a) Instance selection compresses curation into one scoring view and a top-$k$ retention step over individually scored instances. 
    Appendix~\ref{app:metric_distributions} illustrates that subsets with similar one-dimensional metric distributions can still yield different benchmark results.
    (b) Fixed-pool recipe search evaluates multiple ordered recipes over the same raw pool $\mathcal{D}_0$; different operator choices, parameters, and ordering transform the same sample-ID space into different selected subsets.
    }
    \vspace{-15px}
    \label{fig:positioning}
\end{figure*}

The effectiveness of supervised fine-tuning (SFT) depends critically on the quality, diversity, and composition of the instruction data used for adaptation~\citep{zhou2023lima, liu2024alignment, chen2023datajuicer}.
Many automatic approaches formulate this problem as \emph{instance-level data selection} (see Figure~\ref{fig:positioning}(a)): they score each sample with quality heuristics, influence estimates, learned representations, or iterative utility signals, and then retain a subset~\citep{liu2024alignment, xia2024less, ma2025mona, lin2025lead}.
This score-and-select view is useful and has produced strong targeted selectors~\citep{xia2024less, ma2025mona, lin2025lead}.
However, it abstracts away how SFT datasets are often assembled in practice, where filtering, source mixing, deduplication, and light cleaning are applied as a multi-stage curation workflow rather than as a top-k instance-retention decision~\citep{chen2026datachef}.


We study \emph{fixed-pool data recipe search} for SFT. In this setting, the raw data pool is fixed, every operator is grounded in cached signals over that pool, and a candidate recipe is an ordered sequence of filtering, selection, deduplication, or set-composition operations whose execution returns a selected subset (see Figure \ref{fig:positioning}(b)). The goal is not to assign one final score to each instance, but to decide which executable recipe should be evaluated next so that the best selected subset after a small number of full SFT evaluations performs as well as possible.
Under this view, a conventional top-$k$ selector is a length-one degenerate recipe: it applies one scoring rule followed by one retention rule. Fixed-pool recipe search strictly enlarges this view by making operator choice, ordering, parameters, and intermediate subset states part of the optimization object.


Fixed-pool data recipe search is complementary to recent workflow-level and LLM-driven systems.
DataChef~\citep{chen2026datachef} formulates end-to-end data recipe generation for LLM adaptation, and LLM-AutoDP~\citep{ huang2026llmautodp} uses LLM agents to generate and iteratively refine data-processing strategies with feedback from model training and evaluation.
These methods show that generate-and-evaluate loops are powerful for automating data workflows.
AutoSelection studies a different controlled setting: all candidates operate on the same raw pool, and language models are not used to rewrite, synthesize, or augment training samples.
This boundary is useful because it helps attribute measured differences primarily to grounded operator choices over a fixed raw pool, while reducing confounds from generation, rewriting, or newly introduced samples.

The resulting problem is a structured search problem under expensive validation.
The candidate space is compositional because a recipe must choose which operators to activate, where to place them, and how to set their parameters.
Operator choices interact through the intermediate subsets they produce, so the value of one step cannot be understood independently of the rest of the recipe.
Moreover, each reliable observation requires recipe execution, model fine-tuning, and benchmark evaluation.
The central algorithmic challenge is therefore budget allocation over interacting operator choices rather than scoring samples once.

In this work, we present \textbf{AutoSelection}, a framework for fixed-pool data recipe search in supervised fine-tuning.
AutoSelection separates cheap search-side reasoning from expensive full evaluation: it caches grounded task-, data-, and model-side signals; probes multiple retention regimes during warmup; represents realized subsets with state vectors; edits current seed recipes through a Summarizer, Proposer, and Ranker; and refreshes the seed only after stagnation.
When language models are used, they act only as search-side assistants for summarizing history, proposing grounded recipe edits, ranking candidates, and reseeding.
Under matched full evaluation budgets, we evaluate whether this design can discover higher-scoring selected subsets than fixed-pool baselines, including Random recipe search and single-operator alternatives. 

Our contributions are threefold.
\begin{itemize}[left=0pt]
    \item \textbf{Problem formulation.}
    We formalize SFT data curation as \emph{fixed-pool data recipe search}, a budgeted black-box problem over executable recipes on a fixed raw pool; this framing subsumes top-$k$ instance selection and makes operator composition, ordering, and realized subset states explicit.
    \item \textbf{Method.}
    We propose \textbf{AutoSelection}, a two-layer solver that decouples cached fixed-pool candidate materialization from expensive full SFT evaluation, using retention warmup, state-aware local edits, GP-assisted ranking, and stagnation-triggered reseeding.
    \item \textbf{Empirical evidence.}
    On a 90K instruction pool across three base models, AutoSelection improves over full-data training, random recipe search, random top-$k$, and single-operator selectors, with additional OOD, stability, structural-ablation, and 1.5B-to-7B transfer analyses.
\end{itemize}

\section{Related Works}

\subsection{Instance-level data selection}

Many automatic data selection methods for instruction tuning operate on individual samples or fixed subsets. One line of work selects data by task relevance, including gradient-based influence signals in LESS \citep{xia2024less} and model-centric activation signals in MONA \citep{ma2025mona}. Influence-based attribution has also been extended to bilevel meta-learning settings, where task and instance effects propagate through both inner and outer optimization loops \citep{ren2025evaluating}. Another line estimates instance utility during training, as in LEAD \citep{lin2025lead}. Related preprocessing-style methods use instruction difficulty, exact or semantic redundancy, and instruction-structure diversity as selection signals \citep{li2024cherry,lee2022dedup,abbas2023semdedup,bukharin2024data}. Although these methods differ in their scoring criteria, they typically instantiate standalone selectors or fixed preprocessing rules. Thus, their common optimization object is an instance or subset ranking, rather than a jointly optimized multi-step data recipe.

\subsection{Recipe and pipeline optimization}

Recent work increasingly treats workflows, recipes, and pipelines as first-class optimization objects. Data-Juicer \citep{chen2023datajuicer} provides configurable operator pipelines, while AutoPipe \citep{chwa2026autopipe} searches broader LLM post-training pipelines under compute constraints. DataChef \citep{chen2026datachef} and LLM-AutoDP \citep{huang2026llmautodp} move toward recipe-level automation, but rely on open-ended recipe generation or LLM-driven processing modules. These methods are adjacent to ours, but many target broader pipeline optimization, external LLM processing, or generated data workflows rather than controlled search over a fixed raw pool. By contrast, our work fixes the raw pool, searches finer-grained sample-level operator compositions, and uses full evaluation as the central search signal. This narrower scope enables controlled budget-matched comparison while preserving the core challenge of fixed-pool data recipe search.

\paragraph{Hyperparameter optimization and AutoML.}
Our setting is also related to black-box hyperparameter optimization and AutoML, where random search~\citep{bergstra2012random}, Bayesian optimization~\citep{snoek2012practical}, Hyperband-style resource allocation \citep{li2018hyperband}, and automated pipeline search allocate limited evaluations over structured configuration spaces \citep{falkner2018bohb, feurer2015autosklearn}. AutoSelection borrows the budgeted-search perspective, but the optimized configurations are ordered data-curation programs whose execution changes the SFT training subset. Thus, each observation reflects recipe execution, supervised fine-tuning, and downstream evaluation, rather than validation loss under a fixed training set. This distinction makes subset state, operator ordering, and fixed-pool attribution central to our formulation.
\section{Method}
\label{sec:method}

\begin{figure*}[t]
    \centering
    \vspace{-15pt}
    \includegraphics[width=\textwidth]{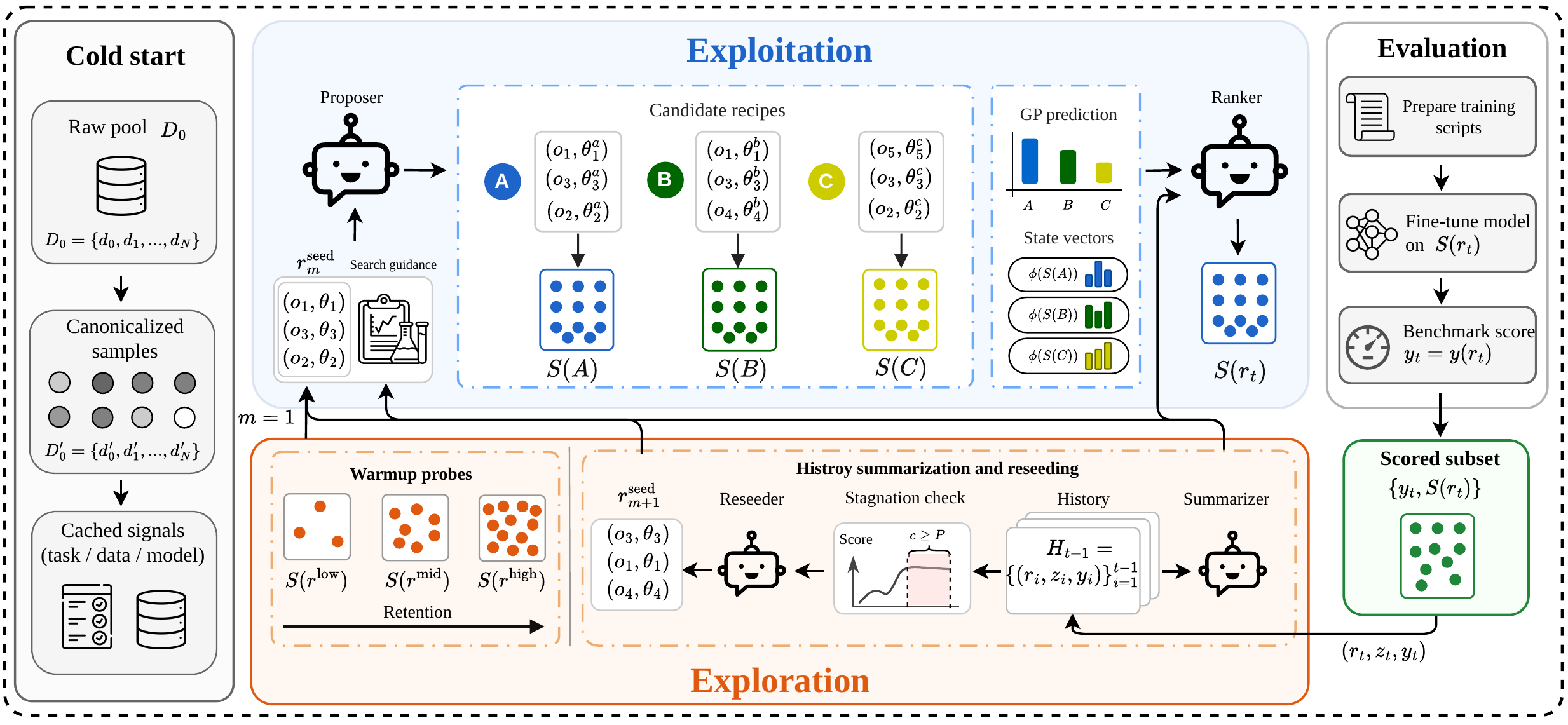}
    \vspace{-15pt}
    \caption{AutoSelection as a solver for fixed-pool data recipe search.
    All candidate recipes operate on the same canonicalized pool derived from the fixed raw pool.
    Warmup probes initialize the search across retention regimes and set the first seed anchor.
    During search, the Summarizer converts evaluated history into guidance, the Proposer generates local recipe edits, the Ranker chooses one materialized candidate for full evaluation, and the Reseeder refreshes the seed anchor after stagnation.}
    \label{fig:method_overview}
    \vspace{-5pt}
\end{figure*}

We first define fixed-pool data recipe search independently of any particular solver, and then instantiate it with AutoSelection. 
We use \textbf{AutoSelection} to refer to the full budgeted solver described in Sections~\ref{sec:method_part1}-\ref{sec:method_part2}. 
Conceptually, AutoSelection has two coupled layers: a fixed-pool materialization layer that caches grounded signals and summarizes executed candidate subsets, and a search-controller layer that allocates full SFT evaluations through warmup, local edits, ranking, and reseeding.
Figure~\ref{fig:method_overview} summarizes this search loop.

\subsection{Fixed-pool data recipe search}
\paragraph{Definition.}
Let $\mathcal{D}_0=\{d_1,\ldots,d_N\}$ be a fixed raw data pool and let $\mathcal{D}_0'=\mathrm{Canon}(\mathcal{D}_0)$ be its canonicalized executable representation.
Canonicalization normalizes fields, attaches source and execution metadata, and preserves stable sample identifiers, so $\mathcal{D}_0'$ contains the same $N$ underlying samples as $\mathcal{D}_0$.
Let $f_0$ be the base model, $\mathcal{E}$ the evaluation suite, and $\mathcal{O}$ a shared library of grounded operators.
Each grounded operator is a subset transformer over the fixed canonicalized pool:
\[
    o(\cdot;\theta): 2^{\mathcal{D}_0'} \rightarrow 2^{\mathcal{D}_0'},
    \qquad o\in\mathcal{O},\ \theta\in\Theta_o ,
\]
where $2^{\mathcal{D}_0'}$ denotes the set of all subsets of $\mathcal{D}_0'$.
The parameter $\theta$ contains the operator-specific execution choices, such as thresholds, retained sizes.
Thus, an operator may filter, select, deduplicate, or recombine samples, but it never creates sample identifiers outside the fixed pool.

A data recipe is a bounded variable-length ordered program
\begin{equation}
    r = \big((o_\ell,\theta_\ell)\big)_{\ell=1}^{L(r)},
    \qquad
    o_\ell \in \mathcal{O},\quad \theta_\ell\in\Theta_{o_\ell},\quad L(r)\le L_{\max},
    \label{eq:recipe_definition}
\end{equation}
where $L(r)$ is the recipe length and $L_{\max}$ is the maximum admissible length.
Let $\mathcal{R}$ denote the recipe space induced by $\mathcal{O}$ and the admissible parameter sets.
Executing a recipe produces a subset, fine-tuned model, and observed utility:

\begin{equation}
    S(r) = \mathrm{Exec}(\mathcal{D}_0', r),
    \qquad
    \hat f_r = \mathcal{A}_{\mathrm{SFT}}(f_0,S(r)),
    \qquad
    y(r) = \mathrm{Eval}_{\mathcal{E}}(\hat f_r).
    \label{eq:recipe_to_score}
\end{equation}
Here $y(r)$ is the observed downstream utility of the recipe after full evaluation.

The fixed-pool recipe search problem is to adaptively choose at most $B$ recipes to evaluate.
For the $t$-th full evaluation, we write $r_t$ for the queried recipe, $S_t=S(r_t)$ for its selected subset, and $y_t=y(r_t)$ for its observed score.
A finite-budget solver returns the best observed recipe and subset:
\begin{equation}
    t^\star \in \arg\max_{1 \le t \le B} y_t,
    \qquad
    r^\star = r_{t^\star},
    \qquad
    S^\star = S_{t^\star}.
    \label{eq:incumbent_objective}
\end{equation}
One full evaluation means executing a recipe, fine-tuning $f_0$ on the selected subset, and evaluating the fine-tuned model on $\mathcal{E}$.
The budget $B$ is therefore the primary resource constraint.

\paragraph{Single-operator selection as a degenerate recipe.}
Let $\mathcal{U}$ be the set of single-operator selectors included in $\mathcal{O}$, and let $\Theta_u$ be the admissible parameter set for $u\in\mathcal{U}$.
If $\mathcal{R}$ contains all length-one recipes, then each single-operator baseline is a recipe $r_{u,\theta}=((u,\theta))\in\mathcal{R}$.
Therefore,
\[
    \max_{r\in\mathcal{R}} y(r)
    \ge
    \max_{u\in\mathcal{U},\,\theta\in\Theta_u}
    y(r_{u,\theta}).
\]
This containment statement does not imply that AutoSelection attains the global maximum under a finite budget $B$; it only formalizes that the recipe space is at least as expressive as the included single-operator selectors.
Appendix~\ref{app:formal_notes} provides the proof and a concrete top-$k$ example.

\subsection{Making the fixed pool searchable: cached signals and realized subset states}
\label{sec:method_part1}
Given the formulation above, a practical solver must address two issues.
First, many candidate recipes need to be materialized without repeatedly recomputing expensive sample-level signals.
Second, candidate recipes that look similar syntactically can produce very different realized subsets.
AutoSelection addresses these issues through a cold-start cache and a state-vector abstraction.

During cold start, the raw samples are canonicalized into $\mathcal{D}_0'$ with normalized instruction-response fields, source metadata, and stable identifiers.
The solver then precomputes reusable task-, data-, and model-side signals over this canonicalized pool.
Task-side signals include benchmark-conditioned activation-similarity statistics following MONA \citep{ma2025mona}.
Data-side signals summarize intrinsic properties such as lexical or instruction-structure diversity.
Model-side signals summarize cached responses of the base model, such as instruction-following difficulty, varentropy, and sparse-activation statistics.
Because these signals are grounded in $\mathcal{D}_0'$, different recipes can reuse the same cached measurements while still producing different selected subsets.

For any materialized candidate recipe $r$, evaluated or not, AutoSelection summarizes the resulting subset with a state vector
\begin{equation}
    z(r)=\phi(S(r))
    =
    \left[
    z^{\mathrm{task}}(r);
    z^{\mathrm{data}}(r);
    z^{\mathrm{model}}(r)
    \right].
    \label{eq:state_vector}
\end{equation}
The task block contains benchmark-conditioned relevance statistics, the data block contains realized scale statistics such as retained-example and token ratios, and the model block contains cached model-side summaries such as IFD, varentropy, and sparse-activation distribution drift.
For an evaluated recipe $r_t$, we write $z_t=z(r_t)$.
State vectors let the search controller compare realized subset properties before spending a full SFT evaluation on a candidate.
Appendix~\ref{app:state_vector} gives the field-level definition and diagnostics.

\subsection{Navigating the recipe space: warmup, local refinement, and reseeding}
\label{sec:method_part2}

The search-controller layer addresses the exploration--exploitation trade-off induced by expensive full evaluations.
Given a limited evaluation budget, AutoSelection should not exploit around an arbitrary initial recipe too early, but it also cannot spend the budget on uniform exploration over the entire compositional recipe space.
Therefore, it begins with a small warmup exploration stage that probes 3 data-scale regimes before switching to seed-centered local refinement.

This warmup stage resolves an early high-impact uncertainty: prior instruction-tuning studies show that downstream behavior can change substantially with the amount of retained data, and that smaller, better-curated subsets can sometimes match or outperform much larger ones \citep{jha2023limit, liu2024alignment, ivison2025largescale}.
We sample candidate recipes from $\mathcal{R}$, monitor the retained-example ratio after execution on $\mathcal{D}_0'$, and keep three probe recipes that fall into low-, medium-, and high-retention bins.
Evaluating these probes provides initial evidence about which retention regime is promising, sets the first seed recipe $r_1^{\mathrm{seed}}$, and initializes the search history $\mathcal{H}_3=\{(r_i,z_i,y_i)\}_{i=1}^{3}$.
The initial seed anchor is set to the best warmup recipe,
\[
    r_1^{\mathrm{seed}}
    \in
    \arg\max_{(r,z,y)\in\mathcal{H}_3} y .
\]

For each later evaluation step $t>3$, AutoSelection uses the accumulated history
$\mathcal{H}_{t-1}=\{(r_i,z_i,y_i)\}_{i=1}^{t-1}$.
The \emph{Summarizer} reads $\mathcal{H}_{t-1}$ and produces search guidance $g_{t-1}=\mathrm{Summarizer}(\mathcal{H}_{t-1})$, a short set of data-backed hypotheses about which operators, retention levels, or compositions appear promising or risky.
These findings are summaries of the evaluated recipes, not edits to the raw pool, and they only steer the next local proposal.

The Proposer then samples a sibling candidate set around the current seed anchor:
\[
    \mathcal{C}_t=\{r'_{t,j}\}_{j=1}^{M_t},
    \qquad
    r'_{t,j}\sim Q_{\mathrm{prop}}(\cdot\mid r_m^{\mathrm{seed}},g_{t-1},\mathcal{H}_{t-1}).
\]
Here $M_t=|\mathcal{C}_t|$, $r'$ denotes a candidate recipe, and $Q_{\mathrm{prop}}$ is a local-edit proposal policy over operations such as inserting, deleting, swapping, or retuning recipe steps.
All proposed recipes are constrained to the operator catalog and are validated before execution.
Each candidate recipe $r'\in\mathcal{C}_t$ is executed on the cached pool to obtain $S(r')$ and $z(r')$, but it is not yet used for SFT.
Following standard Bayesian optimization practice for expensive ML evaluations \citep{snoek2012practical}, a GP surrogate fitted on previous recipe encodings provides a cheap score prior; Appendix~\ref{app:recipe_vector} gives the compact recipe-vector example used for this surrogate feature.

\begin{equation}
    \hat\mu_{t-1}(r'),\hat\sigma_{t-1}(r')
    =
    \mathrm{GP}_{t-1}(\psi(r')).
    \label{eq:gp_surrogate}
\end{equation}
The surrogate uses only the recipe encoding; realized state summaries are passed to the Ranker instead of entering the surrogate.
The \emph{Ranker} combines recipe structure, state-vector information, the GP prior, and search history to choose one candidate for full evaluation:
\begin{equation}
    \rho_t(r') =
    f_{\mathrm{rank}}\!\left(
    r', z(r'), \hat\mu_{t-1}(r'), \hat\sigma_{t-1}(r'), \mathcal{H}_{t-1}
    \right),
    \qquad
    r_t \in \arg\max_{r' \in \mathcal{C}_t} \rho_t(r').
    \label{eq:ranker_score}
\end{equation}
Only $r_t$ is evaluated, and the resulting triple $(r_t,z_t,y_t)$ is appended to the history.

The seed anchor and incumbent are tracked separately.
The incumbent $(r^\star,S^\star,y^\star)$ is the best observed result, whereas $r_m^{\mathrm{seed}}$ defines the local proposal neighborhood for the current search phase.
Keeping the seed fixed within a phase avoids moving the neighborhood after every noisy single evaluation.
When the search has not improved the incumbent for $P$ consecutive evaluations, the Reseeder refreshes the anchor:
\[
    r_{m+1}^{\mathrm{seed}}
    \sim
    Q_{\mathrm{seed}}(\cdot\mid\mathcal{H}_t),
\]

where $Q_{\mathrm{seed}}$ is a history-conditioned policy that selects a new promising motif or recipe region.
This mechanism provides exploration after stagnation while keeping most full evaluations focused on local exploitation.
Algorithm~\ref{alg:search_pseudocode} in Appendix~\ref{app:search_pseudocode} gives the full pseudocode, and Appendix~\ref{app:formal_notes} summarizes the search-side complexity.


\vspace{-10pt}
\section{Experiments}
\label{sec:experiments}

\subsection{Experimental setup}
\label{subsec:exp_setup}

We evaluate AutoSelection in the fixed-pool setting defined in Section~\ref{sec:method}.
All methods receive the same raw SFT pool, operate without synthetic data generation, LLM-based sample rewriting, or pool augmentation, and are compared through the quality of the selected subset.
The raw pool is a 90K-sample merged instruction-tuning pool, constructed by sampling 30K samples from each of OpenHermes-2.5 \citep{teknium2023openhermes}, the LESS instruction-tuning data pool \citep{xia2024less}, and Alpaca-52K \citep{wang2023selfinstruct}.
The validation suite contains GPQA \citep{rein2024gpqa}, GSM8K \citep{cobbe2021training}, BBH \cite{srivastava2023beyond}, and MMLU \citep{hendrycks2021measuring}.
We evaluate our method on three base models--Qwen2.5-1.5B \citep{qwen2025qwen25technicalreport}, Llama3.2-1B \citep{grattafiori2024llama3herdmodels}, and Qwen2.5-3B \citep{qwen2025qwen25technicalreport}--to examine its effectiveness across model families and scales.
The search budget is $B=15$ full evaluations; each evaluation executes one recipe, fine-tunes the base model on the selected subset, and evaluates the resulting model. We use a stagnation patience of $P=4$ for reseeding and fix the proposer candidate set size to $|\mathcal{M}_t|=5$. When multiple AutoSelection runs are available, the main table reports the median selected-subset result. We also test the selected subsets and their recipes on two OOD graph benchmarks, GraphWiz \citep{chen2024graphwiz} and NLGraph \citep{wang2023can}. Implementation details and prompt templates are provided in Appendices~\ref{app:implementation_details} and~\ref{app:prompt_templates}.

We use one shared operator library for all search methods. The library covers task relevance (MONA~\citep{ma2025mona}), model-internal difficulty and uncertainty (IFD~\citep{li2024cherry} and varentropy~\citep{6665143,Maadani2020ANG,li-etal-2026-entropy,he2026rethinking}), intrinsic data diversity (N-gram entropy~\citep{Shannon1948AMT,wu2025the} and action-object branching~\citep{zhao2023preliminary, bukharin2024data}), redundancy reduction (SemDedup~\citep{abbas2023semdedup}), stochastic exploration (Random top-$k$~\citep{bergstra2012random,diddee2025chasing,ivison2025large, nayak2026critical}), and set composition (Mix). The same library defines the AutoSelection search space and the Random recipe search baseline.
For single-operator baselines, we choose retained-scale operating points from the Qwen2.5-1.5B check in Appendix~\ref{app:baseline_param_protocol} and reuse them across model sizes for a consistent cross-scale comparison. Appendix~\ref{app:operator_library} gives the full operator definitions, parameters, and implementation details. Appendix~\ref{app:llm_autodp} reports a boundary-case LLM-AutoDP pilot and explains why it is not included as a main fixed-pool baseline.
\begin{table*}[t]
\centering
\vspace{-20pt}
\caption{Main results across model scales. \textit{Group} separates full-pool training, recipe-level methods, top-$k$ single-operator selection, and deduplication baselines. The best score in each metric column is bolded and the second-best score is underlined.}
\label{tab:main_results}
{\small
\setlength{\tabcolsep}{2.4pt}
\begin{tabularx}{\textwidth}{@{}L{0.95cm}L{2.05cm}*{4}{Y}VYYV@{}}
\toprule
\multirow{2}{*}{Group}
& \multirow{2}{*}{Method}
& \multicolumn{5}{c}{In-distribution reasoning}
& \multicolumn{3}{c}{OOD graph} \\
\cmidrule(lr){3-7}
\cmidrule(l){8-10}
& & GPQA & GSM8K & BBH & MMLU & \textbf{Avg} & GraphWiz & NLGraph & \textbf{Avg} \\
\midrule

\groupbar{Llama3.2-1B} \\
Full & Full data
& 17.19 & \textbf{16.60} & 4.78 & 33.06 & 17.91
& \textbf{33.62} & \textbf{57.25} & \textbf{45.44} \\
Recipe & Random
& 15.62 & 13.49 & \underline{9.45} & 32.33 & 17.72
& 32.16 & 53.55 & 42.86 \\
Top-$k$ & Random
& 19.41 & \underline{14.10} & 8.47 & 33.33 & \underline{18.83}
& 32.43 & 53.22 & 42.83 \\
Top-$k$ & MONA
& 18.97 & 10.38 & \textbf{9.89} & \underline{33.61} & 18.21
& 32.09 & 54.67 & 43.38 \\
Top-$k$ & AO
& 21.43 & 9.55 & 9.04 & 33.33 & 18.34
& 31.21 & 55.00 & 43.11 \\
Top-$k$ & IFD
& 16.07 & 10.08 & 4.34 & 29.33 & 14.96
& 29.50 & 50.80 & 40.15 \\
Top-$k$ & N-gram
& \underline{24.77} & 10.76 & 3.26 & \textbf{34.55} & 18.34
& 32.65 & 50.48 & 41.57 \\
Dedup & SemDedup
& 17.41 & 6.36 & 0.76 & 26.77 & 12.83
& 30.56 & 49.35 & 39.96 \\
Top-$k$ & Varentropy
& 17.18 & 3.56 & 8.69 & 31.05 & 15.12
& 32.43 & 48.70 & 40.57 \\
\rowcolor{tableours}
Recipe & \textbf{AutoSelection}
& \textbf{26.78} & 11.52 & 6.08 & \textbf{34.55} & \textbf{19.73}
& \underline{33.59} & \underline{55.48} & \underline{44.54} \\

\midrule
\groupbar{Qwen2.5-1.5B} \\
Full & Full data
& 21.20 & \textbf{55.26} & 22.93 & 55.94 & 38.83
& 36.96 & 54.03 & 45.50 \\
Recipe & Random
& 23.43 & 52.23 & 27.39 & 56.83 & 39.97
& 36.88 & 50.16 & 43.52 \\
Top-$k$ & Random
& 21.20 & 52.16 & 27.82 & 56.38 & 39.39
& \underline{38.71} & 53.22 & 45.97 \\
Top-$k$ & MONA
& 21.20 & 54.35 & 24.89 & 55.77 & 39.05
& 37.18 & 56.45 & 46.82 \\
Top-$k$ & AO
& 23.66 & 49.81 & \textbf{33.36} & 55.33 & \underline{40.54}
& 37.65 & \underline{57.74} & \underline{47.70} \\
Top-$k$ & IFD
& 23.21 & 23.88 & 24.45 & 56.22 & 31.94
& 37.93 & 57.09 & 47.51 \\
Top-$k$ & N-gram
& \underline{24.10} & 51.63 & 19.13 & \textbf{57.38} & 38.06
& 37.28 & 52.25 & 44.77 \\
Dedup & SemDedup
& 14.73 & 40.40 & 20.43 & \underline{56.88} & 33.11
& 38.46 & 53.54 & 46.00 \\
Top-$k$ & Varentropy
& 20.31 & 38.51 & 25.76 & 55.50 & 35.02
& 35.71 & 52.90 & 44.31 \\
\rowcolor{tableours}
Recipe & \textbf{AutoSelection}
& \textbf{29.01} & \underline{54.58} & \underline{30.00} & 55.33 & \textbf{42.23}
& \textbf{38.91} & \textbf{58.54} & \textbf{48.73} \\

\midrule
\groupbar{Qwen2.5-3B} \\
Full & Full data
& \underline{23.66} & 64.82 & 31.95 & 61.00 & 45.36
& \underline{37.31} & 64.03 & 50.67 \\
Recipe & Random
& 21.43 & 64.44 & 30.22 & \underline{62.00} & 44.52
& 35.71 & \textbf{66.29} & \textbf{51.00} \\
Top-$k$ & Random
& 21.87 & 64.59 & 31.73 & 61.94 & 45.03
& 36.75 & 56.29 & 46.52 \\
Top-$k$ & MONA
& \underline{23.66} & \underline{70.58} & 28.69 & 61.83 & \underline{46.19}
& \textbf{38.56} & 59.19 & 48.88 \\
Top-$k$ & AO
& 23.21 & 62.69 & \underline{35.86} & 60.11 & 45.47
& 34.62 & \underline{66.12} & 50.37 \\
Top-$k$ & IFD
& 21.42 & 21.22 & 21.84 & 60.66 & 31.29
& 36.25 & 55.80 & 46.03 \\
Top-$k$ & N-gram
& 21.42 & 59.66 & 21.63 & 61.11 & 40.96
& 36.15 & 57.41 & 46.78 \\
Dedup & SemDedup
& \underline{23.66} & 60.19 & 29.23 & 61.88 & 43.74
& 35.87 & 61.93 & 48.90 \\
Top-$k$ & Varentropy
& \textbf{28.12} & 49.20 & 35.21 & 61.27 & 43.45
& 35.81 & 56.61 & 46.21 \\
\rowcolor{tableours}
Recipe & \textbf{AutoSelection}
& 22.99 & \textbf{72.78} & \textbf{36.84} & \textbf{63.72} & \textbf{49.08}
& 36.28 & 65.65 & \underline{50.97} \\
\bottomrule
\end{tabularx}
}
\vspace{-15pt}
\end{table*}
\begin{figure}[h]
    \centering
    \vspace{-10pt}
    \includegraphics[width=\linewidth]{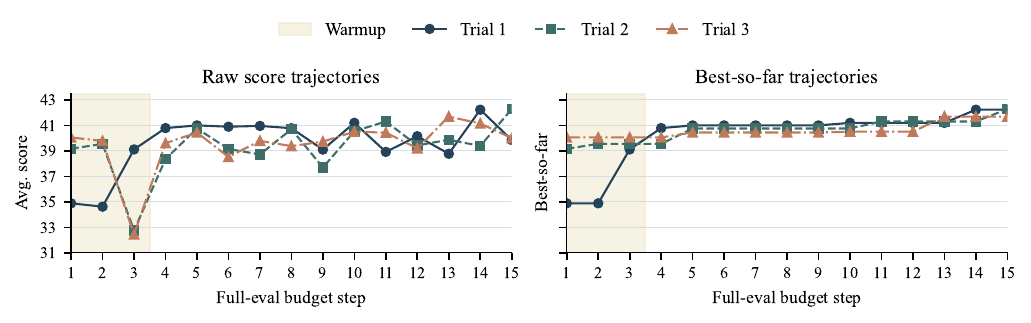}
    \vspace{-14pt}
     \caption{Raw-score and best-so-far curves for three 1.5B AutoSelection runs under the same 15 full evaluation budget.}
    \label{fig:seed_stability}
    \vspace{-15pt}
\end{figure}
\subsection{Main results}
\label{subsec:main_results}

As shown in Table~\ref{tab:main_results}, AutoSelection achieves the highest in-distribution reasoning average for all three base models. Compared with the strongest non-AutoSelection baseline in each block, it improves the reasoning average by 0.90, 1.69, and 2.89 points on Llama3.2-1B, Qwen2.5-1.5B, and Qwen2.5-3B, respectively; compared with full-data training, the gains are 1.82, 3.40, and 3.72 points. The best single-operator baseline varies across tasks and model scales, indicating selector fragility and supporting the need for recipe-level composition. On the held-out graph benchmarks, AutoSelection remains competitive: it ranks first on Qwen2.5-1.5B, nearly ties the best baseline on Qwen2.5-3B, and is the strongest non-full-data method on Llama3.2-1B.
Overall, the table suggests that the value of AutoSelection lies less in discovering a universally best selector, and more in allocating a small number of full evaluations to find a stronger composition of grounded curation decisions over the same fixed pool.

\subsection{Search stability under randomness}
\label{subsec:seed_stability}

To check whether AutoSelection's gain is caused by a lucky random search, we repeat the Qwen2.5-1.5B setting for three independent runs and report the selected subset score under the same 15 full evaluation budget in Table~\ref{tab:seed_stability} at Appendix~\ref{app:seed_stability}; the corresponding raw-score and best-so-far curves are shown in Figure~\ref{fig:seed_stability}. The Qwen2.5-1.5B AutoSelection row in Table~\ref{tab:main_results} reports the median of these repeated runs rather than the best run. The three selected subset scores are 42.23, 42.28, and 41.69, with a mean of 42.07 and a narrow range of 0.59 points. Even the lowest run remains above the strongest non-AutoSelection baseline in Table~\ref{tab:main_results} (41.69 vs. 40.54), while the best-so-far curves converge to a similar score band after the warmup stage. These three available 1.5B runs suggest that the observed improvement is not solely a single lucky run under this evaluation budget.

\subsection{Search-side ablations}
\label{subsec:search_ablations}

\begin{figure}[t]
    \centering
    \vspace{-20pt}
    \includegraphics[width=\linewidth]{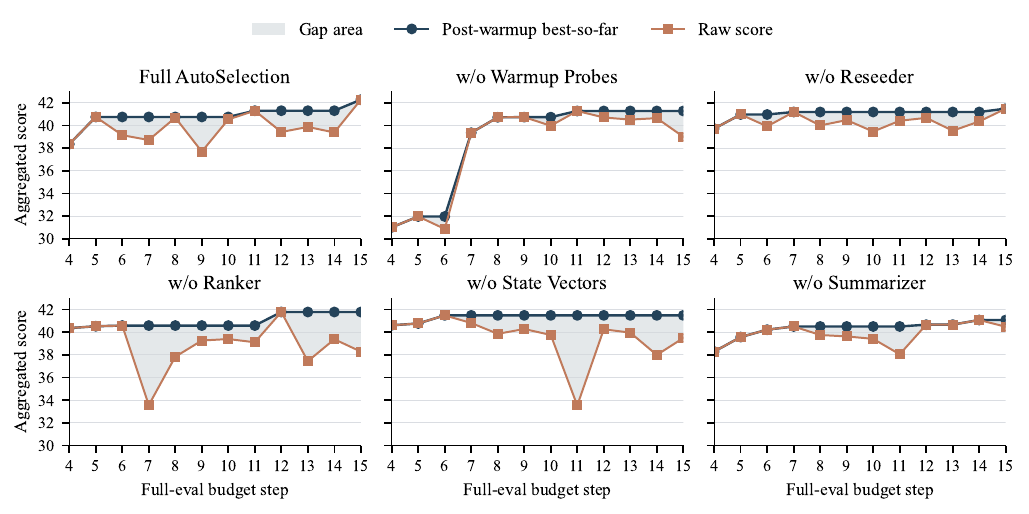}
    \vspace{-14pt}
    \caption{Trajectory diagnostics for the search-side ablations on the 1.5B setting. Curves show post-warmup raw scores and best-so-far scores over full-evaluation steps 4--15; shaded regions indicate the gap-area diagnostic used to summarize exposure to low-scoring edits.}
    \label{fig:search_ablation_panel}
\end{figure}

The ablation in Table~\ref{tab:search_ablation} is intended as a search-policy diagnostic rather than the main evidence for AutoSelection. 
We use the strongest complete 1.5B AutoSelection run as the full-reference setting. 
Except for w/o Warmup, which removes the initial retention-regime probes to test warmup sensitivity, each ablation starts from the same warmup recipes and then removes one post-warmup search component. 
For w/o Ranker, candidates are selected using only GP scores. 
Best@4--15 is the primary metric, while Mean@4--15 and GapArea@4--15 summarize trajectory smoothness and exposure to low-scoring edits. 
Under this objective, the full setting reaches the highest post-warmup score; the remaining columns and Figure~\ref{fig:search_ablation_panel} are used to interpret how different components affect the search trajectory.
\begin{table*}[h]
\centering
\caption{Search-side ablations on the 1.5B setting. The primary objective is the best observed post-warmup recipe under the fixed budget; benchmark columns decompose this selected recipe. Mean@4--15 and GapArea@4--15 are trajectory diagnostics over post-warmup steps 4--15, not optimization targets.}
\label{tab:search_ablation}
{\small
\begin{tabular*}{\textwidth}{@{}l@{\extracolsep{\fill}}ccccc cc@{}}
\toprule
\strut
& \multicolumn{5}{c}{Primary objective: best observed recipe}
& \multicolumn{2}{c}{Trajectory diagnostics} \\
\cmidrule(lr){2-6}
\cmidrule(l){7-8}
Method & GPQA & GSM8K & BBH & MMLU & Best@4--15 & Mean@4--15 & GapArea@4--15 \\
\midrule
Full AutoSelection & 24.55 & 58.00 & 29.02 & 57.56 & \textbf{42.28} & 39.85 & 1.02 \\
w/o Warmup & 25.45 & 53.90 & 29.24 & 56.56 & 41.29 & 38.07 & 0.51 \\
w/o Reseeder & 23.66 & 56.18 & 29.13 & 57.00 & 41.49 & 40.36 & 0.71 \\
w/o Ranker & 23.44 & 53.30 & 33.59 & 56.83 & 41.79 & 38.97 & 2.00 \\
w/o State Vectors & 28.35 & 50.57 & 30.33 & 56.72 & 41.49 & 39.57 & 1.79 \\
w/o Summarizer & 23.66 & 54.66 & 29.67 & 56.33 & 41.08 & 39.87 & 0.48 \\
Random Select & 25.89 & 53.90 & 28.37 & 56.50 & 41.17 & 39.66 & 1.27 \\
\bottomrule
\end{tabular*}
}
\vspace{-10pt}
\end{table*}

The diagnostics suggest three main component roles rather than monotonic improvements on every trajectory statistic. 
1) Warmup and summarization improve budget use in different ways: removing warmup makes the search spend early evaluations in a low-scoring region, while removing the Summarizer keeps a relatively smooth trajectory but finds a weaker best recipe, suggesting that history summaries help turn past outcomes into sharper local guidance. 
2) Reseeding trades short-term smoothness for peak discovery: w/o Reseeder is comparatively stable, but its best-so-far curve flattens below the full method, indicating that reseeding helps escape saturated local regions. 
3) Ranking and state vectors help screen risky edits: GP-only selection outperforms random candidate choice on Best@4--15, suggesting that the surrogate provides useful coarse direction, but the full Ranker is more stable; removing state vectors exposes the search to candidates whose recipe form may look promising but whose realized subset state is weaker. 
Additional diagnostics in Appendix~\ref{app:random_select_curve}, Appendix~\ref{app:gp_fit}, and Appendix~\ref{app:recipe_rank_case} further separate random candidate selection, GP surrogate behavior, and Ranker allocation quality. 
Overall, these ablations suggest that AutoSelection's components mainly improve budget allocation and candidate screening, while full SFT evaluation remains necessary for confirming the best recipe.

\subsection{Matched-seed Structural ablations}
\label{subsec:structural_ablations}

We further examine whether the selected recipe is sensitive to structure beyond its operator set. 
Because the reference Qwen2.5-1.5B recipe contains a random-$k$ operator, we use a matched-seed design: the reference recipe and each structural variant are executed with the same random seeds (42,256,1024), while keeping operator parameters and the evaluation protocol fixed. 
This controls the stochastic component of the random-$k$ step and makes the paired drop from the reference recipe the relevant comparison.
As shown in Table~\ref{tab:structural_ablation}, the reference recipe obtains the highest mean score, while all structural variants show positive mean paired drops. 
These results suggest that, for this reference recipe, ordering and composition
can affect performance beyond the operator set alone, supporting our recipe-level
formulation of fixed-pool data optimization.
The case studies in Appendix~\ref{app:operator_composition_cases} provide additional qualitative support.

\begin{table}[h]
\centering
\caption{Matched-seed structural ablation of the selected Qwen2.5-1.5B recipe. 
Each variant uses the same random-$k$ seeds as the reference recipe while keeping operator parameters and the evaluation protocol fixed. 
Drop reports the mean paired decrease relative to the reference recipe across the three seeds.}
\label{tab:structural_ablation}
{\small
\renewcommand{\arraystretch}{1.08}
\setlength{\tabcolsep}{4.0pt}
\begin{tabular*}{\linewidth}{@{}l@{\extracolsep{\fill}}lccc@{\hspace{0.45cm}}}
\toprule
Variant & Recipe & Avg & Std & Drop \\
\midrule
Reference recipe 
& N-gram$\rightarrow$MONA$\rightarrow$SemDedup$\rightarrow$random-$k$
& \textbf{41.61} & 0.47 & -- \\
Swap MONA/N-gram 
& MONA$\rightarrow$N-gram$\rightarrow$SemDedup$\rightarrow$random-$k$
& 40.68 & 0.59 & 0.93 \\
SemDedup early 
& N-gram$\rightarrow$SemDedup$\rightarrow$MONA$\rightarrow$random-$k$
& 41.10 & 0.57 & 0.51 \\
No SemDedup 
& N-gram$\rightarrow$MONA$\rightarrow$random-$k$
& 40.94 & 0.26 & 0.67 \\
Mix replacement 
& N-gram$\rightarrow$MONA$\rightarrow$Mix$\rightarrow$random-$k$
& 40.72 & 0.53 & 0.89 \\
\bottomrule
\end{tabular*}
}
\end{table}

\subsection{Transferability to larger models}
\label{subsec:transferability}

Motivated by prior evidence that small models can provide useful signals for larger-model data selection and mixture optimization~\citep{mekala-etal-2024-smaller,NEURIPS2024_97fe251c,NEURIPS2023_dcba6be9}, we conduct a 7B transfer check with Qwen2.5-7B \citep{qwen2025qwen25technicalreport} as supporting analysis. Specifically, selected subsets obtained during the Qwen2.5-1.5B search are used to fine-tune a larger model, covering subsets produced by both strong and weaker runs. The observed trend is coarse but informative: selected subsets from stronger 1.5B runs tend to remain competitive after transfer, while selected subsets from weaker runs do not reliably become strong solely because the target model is larger (see Table~\ref{tab:transfer_7b} in Appendix~\ref{app:transferability}). The 1.5B and 7B selected-subset rankings have a positive Spearman rank correlation of $\rho=0.82$, suggesting moderate cross-scale similarity but not exact rank preservation. We therefore treat cross-scale transfer as an approximate data-quality signal. This one 1.5B-to-7B transfer check suggests that some selected-subset motifs may remain useful across scale.
\section{Conclusion}

In this work, we introduce AutoSelection, a fixed-pool data recipe search framework for supervised fine-tuning. Instead of treating data selection as a single instance-level ranking problem, AutoSelection optimizes ordered combinations of filtering, mixing, deduplication, and selection operators under a limited full-evaluation budget. By caching reusable task-, data-, and model-side signals, using warmup probes, summarizing search history, ranking candidate recipes, and reseeding after stagnation, AutoSelection explores compositional data-curation strategies efficiently. Evaluations across multiple model scales show that recipe-level search improves the in-distribution reasoning average over full-data training, single-operator selection, Random top-$k$, and Random recipe search, while remaining competitive on held-out graph benchmarks. Additional analyses indicate that operator ordering and composition can affect downstream SFT performance, supporting the value of treating data recipes as first-class optimization objects.

\section*{Limitations and Future Work}
\label{sec:conandlimi}
This work focuses on fixed-pool data recipe search under a controlled evaluation protocol. While AutoSelection improves SFT performance across the studied settings, our evaluation is still limited to a moderate-size instruction pool, several model scales, and a finite set of reasoning benchmarks. Future work can extend the evaluation to larger raw pools, more heterogeneous data sources, and domain-specific SFT tasks to better understand the generality of recipe-level data optimization. In addition, AutoSelection relies on full evaluation, where each evaluation requires recipe execution, model fine-tuning, and benchmark evaluation. This makes scaling to larger models or longer search budgets computationally expensive. Developing cheaper proxy evaluations, multi-fidelity search strategies, or transferable recipe priors is therefore an important direction for making fixed-pool data recipe search more scalable.

\bibliographystyle{unsrtnat}
\bibliography{reference}

\newpage
\appendix

\section{Experimental Protocol and Operator Details}
\label{app:experimental_protocol}

\subsection{Operator Library}
\label{app:operator_library}

Table~\ref{tab:operator_library} summarizes the operators used in our
experiments. The library is designed to cover heterogeneous selection signals
rather than a single definition of data quality. These operators include
task-relevance signals, model-internal difficulty and uncertainty signals,
intrinsic diversity signals, redundancy-reduction operations, and recipe-level
combinators. For single-operator baselines, we apply one operator at a time. For
AutoSelection, the same operators form the search space from which candidate
recipes are constructed.

\begin{table}[h]
\centering
\small
\caption{
Summary of the operator library used in AutoSelection. Each operator provides
a distinct signal or composition rule for constructing candidate subsets.
}
\label{tab:operator_library}
\begin{tabular}{ll}
\toprule
\textbf{Operator} & \textbf{Signal / Criterion} \\
\midrule
MONA top-$k$
& Benchmark-conditioned activation similarity \\

IFD top-$k$
& Instruction-data difficulty \\

Varentropy top-$k$
& Token-level uncertainty fluctuation \\

N-gram top-$k$
& N-gram entropy / lexical diversity \\

AO top-$k$ (action-object)
& Action-object pattern diversity \\

SemDedup
& Embedding-space near-duplicate similarity \\

Random top-$k$
& Uniform sampling from the current candidate pool \\

Mix
& Set union of two selected subsets \\
\bottomrule
\end{tabular}
\end{table}

\paragraph{Operator roles.}
MONA provides a direct task-relevance signal from benchmark-conditioned model
representations. For multiple validation tasks, we compute task-specific
similarity scores, select top-$k$ samples for each task, and merge the selected
sets by union. IFD and varentropy are
model-internal signals: IFD measures how difficult an instruction-response pair is
for the base model, while varentropy captures fluctuations in token-level
predictive uncertainty. N-gram entropy and action-object branching (AO) are
lightweight data-side diversity signals, targeting lexical diversity and
instruction-structure diversity respectively; they are intended as interpretable
axes of variation rather than standalone claims about data quality. SemDedup
reduces near-duplicate samples in the embedding space.

Random top-$k$ acts as a stochastic escape operator. Because deterministic
filters can repeatedly return overlapping subsets, random sampling allows the
search to jump to alternative data regions under the same budget. This follows
the motivation of random search in structured configuration spaces
\citep{bergstra2012random} and is further supported by recent instruction-data
selection studies where random subsets remain strong and difficult to
consistently outperform \citep{diddee2025chasing,ivison2025large}.

In addition to filtering operators, AutoSelection includes a union-style
combinator for recipe construction. When this operation is selected, the current
candidate subset is mixed with a previously strong subset from the search
history. The operation is implemented as a set union over sample identifiers, so
duplicate samples are retained only once. This allows the search process to
reuse effective data regions while still exploring new operator compositions.
\paragraph{Operator computation.}
given a score $s(x)$ over
pool $D$, top-fraction $\alpha$ keeps
$\operatorname{Top}_{\lceil \alpha |D|\rceil}(D;s)$, top-$k$ keeps
$\operatorname{Top}_{k}(D;s)$, and thresholding keeps
$\{x:s(x)\ge \tau\}$. MONA computes a benchmark-conditioned relevance score in
sparse-autoencoder (SAE) space,
\[
s_b(x)=\frac{\sum_j \min(a_j(x), t_{b,j})}{\sum_j \max(a_j(x), t_{b,j})},
\]
where $a(x)$ is the sparse activation vector of sample $x$ and $t_b$ is the
target vector for benchmark $b$; for multiple benchmarks, we keep top samples
per benchmark and take their union. IFD uses the cached instruction-following
difficulty score
\[
s_{\mathrm{IFD}}(x)=
\frac{\mathcal{L}(\mathrm{response}\mid \mathrm{instruction})}
{\mathcal{L}(\mathrm{response})}.
\]
Varentropy is computed from the token distribution $p_j$ at token position $j$ as
$H_j=-\sum_v p_j(v)\log p_j(v)$ and
$V_j=\sum_v p_j(v)(\log p_j(v)+H_j)^2$, then averaged over valid tokens.
N-gram entropy is implemented as unigram Shannon entropy over normalized word
tokens, $H(x)=-\sum_w p_x(w)\log_2 p_x(w)$. AO scores dependency branching by
$s_{\mathrm{AO}}(x)=0.5\,|\mathrm{verbs}(x)|+
0.5\,\operatorname{mean}_{v\in \mathrm{verbs}(x)}|\mathrm{subtree}(v)|$.
SemDedup builds L2-normalized SAE sparse vectors, clusters them with MiniBatch
K-Means, and greedily drops $x_i$ within its cluster if
$\max_{x_j\in K_c}\cos(\hat{a}_i,\hat{a}_j)\ge \tau$ for already kept samples
$K_c$.

Random top-$k$ draws $k$ samples uniformly from the current candidate pool,
which acts as a stochastic escape operator and follows the motivation of random
search in structured configuration spaces~\citep{bergstra2012random}. The Mix
operator is a set union over sample identifiers,
$\mathrm{Mix}(C,S)=C\cup(S\setminus C)$, where $C$ is the current subset and
$S$ is a selected subset from search history.
\paragraph{Use in baselines and AutoSelection.}
The single-operator baselines evaluate whether any individual signal is
sufficient on its own. Random top-$k$ evaluates whether improvements exceed simple
random subset selection. Random recipe search samples compositions from the same
operator space without the proposed search strategy. AutoSelection uses the full
operator library to search over data recipes, allowing heterogeneous
signals and recipe-level combinators to be combined under a fixed validation
budget.

\subsection{Implementation Details}
\label{app:implementation_details}

All reported evaluations use the same fixed raw pool described in Section~\ref{subsec:exp_setup}.
Before search, samples are canonicalized into a shared representation with stable sample identifiers, normalized instruction-response fields, source metadata, and cached operator-side signals.
Candidate recipes therefore differ only in how they transform this fixed pool; they do not add synthetic samples, rewrite responses, or query external generators during data construction.

For MONA-style task-relevance features, we follow the task-vector construction and related feature-extraction settings from MONA \citep{ma2025mona}.
The sparse-autoencoder (SAE) model used to obtain sparse activation features is trained with the EleutherAI \texttt{sparsify} library\footnote{\url{https://github.com/EleutherAI/sparsify}} on RedPajama-Data-1T.\footnote{\url{https://huggingface.co/datasets/togethercomputer/RedPajama-Data-1T}}
All other MONA-specific hyperparameters are kept consistent with the original MONA setting.

One full evaluation consists of executing a recipe on the canonicalized pool, fine-tuning the selected base model on the selected subset, and evaluating the fine-tuned model on the validation suite.
We run SFT training with LLaMA-Factory and use vLLM as the inference backend for benchmark evaluation.
The primary search budget is $B=15$ full evaluations.
Full evaluations are conducted on a single-node server equipped with 16 Ascend 910C accelerators.
Warmup occupies the first three evaluations by probing low-, medium-, and high-retention regimes; the later search-side ablations align comparisons on the post-warmup budget, i.e., steps 4--15.
In the ablation table, metrics named Best@4--15, Mean@4--15, and GapArea@4--15 are computed over this post-warmup window rather than over all 15 evaluations.

All SFT jobs use the same training configuration summarized in Table~\ref{tab:training_hyperparams}.
We keep these hyperparameters fixed across candidate recipes so that measured differences are attributable to the selected subset rather than optimizer or systems settings.

\begin{table}[h]
\centering
\caption{Main SFT training hyperparameters used for full evaluation.}
\label{tab:training_hyperparams}
\small
\begin{tabular}{ll}
\toprule
\textbf{Hyperparameter} & \textbf{Value} \\
\midrule
Training stage & SFT \\
Precision & bf16 \\
Maximum sequence length & 2048 \\
Epochs & 3 \\
Learning rate & $2.0 \times 10^{-5}$ \\
Scheduler & Cosine \\
Warmup ratio & 0.1 \\
Train batch size & 256 \\
Flash attention & FA2 \\
\bottomrule
\end{tabular}
\end{table}

The validation score aggregates GPQA, GSM8K, BBH, and MMLU.
GraphWiz and NLGraph are held out from the search objective and are used as OOD graph benchmarks in the main result table.
To simplify the evaluation pipeline while preserving broad benchmark coverage, we evaluate MMLU on a fixed randomly sampled 10\% subset shared by all methods.
For NLGraph, we filter out topology tasks and retain 3,200 samples for the final graph-reasoning evaluation set, again using the same filtered set for all methods.
The prompt templates used for benchmark evaluation are reported in Appendix~\ref{app:prompt_templates}.

Language models used inside AutoSelection act only as search-side assistants: they summarize previous evaluations, propose grounded recipe edits, rank candidate recipes after recipe execution and state-vector extraction but before full evaluation, and reseed after stagnation.
In our implementation, these search-side LLM calls use DeepSeek-R1 \citep{guo2025deepseek} as the backend model.
They do not inspect held-out answers, generate new training samples, or modify the raw pool.

\newpage
\subsection{Baseline Parameter Setting}
\label{app:baseline_param_protocol}

Table~\ref{tab:baseline_param_check_1p5b} reports the Qwen2.5-1.5B
operating-point check used to set the single-operator baselines in the main
table. We use them only to avoid arbitrary baseline parameters:
for single selectors, the main-table operating point is chosen to keep
the resulting retained-data scale close to the best AutoSelection subset when
that operator exposes direct size control. The best AutoSelection subset
reported in the main table contains roughly 45K retained examples, about half of the raw
training pool; therefore, size-controlled top-$k$ baselines use $0.5$ as the
default operating point. Since these parameters are operator-specific thresholds
or fractions, their numeric values are not always identical to the final retained
ratio. MONA top-$k$ keeps the $0.05$ fraction from the original MONA setting.
For evaluation efficiency and a consistent cross-scale comparison, we reuse
these operating points on the other model sizes.
\begin{table*}[h]
\centering
\caption{Qwen2.5-1.5B operating-point check for single-operator baselines. The reasoning average is computed over GPQA, GSM8K, BBH, and MMLU. Bold parameter and Avg entries indicate the operating points used for the in-distribution columns in the main table.}
\label{tab:baseline_param_check_1p5b}
{\small
\setlength{\tabcolsep}{3.4pt}
\renewcommand{\arraystretch}{1.06}
\begin{tabular}{lcccccc}
\toprule
Method & Param & GPQA & GSM8K & BBH & MMLU & \textbf{Avg} \\
\midrule
Random top-$k$ & \textbf{0.5} & 21.20 & 52.16 & 27.82 & 56.38 & \textbf{39.39} \\
\midrule
MONA top-$k$ & \textbf{0.05} & 21.20 & 54.35 & 24.89 & 55.77 & \textbf{39.05} \\
MONA top-$k$ & 0.1 & 23.88 & 51.02 & 26.30 & 55.00 & 39.05 \\
MONA top-$k$ & 0.3 & 22.32 & 53.44 & 23.58 & 55.77 & 38.78 \\
\midrule
AO top-$k$ & 0.1 & 18.08 & 48.80 & 26.30 & 51.94 & 36.28 \\
AO top-$k$ & 0.3 & 20.75 & 48.29 & 27.71 & 52.77 & 37.38 \\
AO top-$k$ & \textbf{0.5} & 23.66 & 49.81 & 33.36 & 55.33 & \textbf{40.54} \\
\midrule
IFD top-$k$ & 0.1 & 12.50 & 17.05 & 20.86 & 53.16 & 25.89 \\
IFD top-$k$ & 0.3 & 16.51 & 17.28 & 14.89 & 54.27 & 25.74 \\
IFD top-$k$ & \textbf{0.5} & 23.21 & 23.88 & 24.45 & 56.22 & \textbf{31.94} \\
\midrule
N-gram top-$k$ & 0.1 & 16.74 & 36.39 & 19.23 & 56.05 & 32.10 \\
N-gram top-$k$ & 0.3 & 22.76 & 51.25 & 16.63 & 56.72 & 36.84 \\
N-gram top-$k$ & \textbf{0.5} & 24.10 & 51.63 & 19.13 & 57.38 & \textbf{38.06} \\
\midrule
SemDedup & 0.975 & 17.41 & 35.10 & 23.26 & 52.88 & 32.16 \\
SemDedup & \textbf{0.985} & 14.73 & 40.40 & 20.43 & 56.88 & \textbf{33.11} \\
\midrule
Varentropy top-$k$ & 0.1 & 16.29 & 43.36 & 17.17 & 50.88 & 31.93 \\
Varentropy top-$k$ & 0.3 & 21.20 & 40.71 & 22.17 & 55.05 & 34.78 \\
Varentropy top-$k$ & \textbf{0.5} & 20.31 & 38.51 & 25.76 & 55.50 & \textbf{35.02} \\
\bottomrule
\end{tabular}
}
\end{table*}

\section{Search Procedure and Formal Details}
\label{app:search_procedure}

\subsection{Recipe Vector}
\label{app:recipe_vector}

Suppose the recipe space contains three operators: task-relevance filtering, deduplication, and size control.
A recipe that enables the first and third operators, disables deduplication, sets a relevance threshold of $0.70$, and targets a retained-example ratio of $0.30$ can be encoded as
\[
    \psi(r) = [1, 0.70, 0, 0.00, 1, 0.30],
\]
where each operator contributes an on/off indicator followed by its normalized $\theta$ value.
In the implementation, $\psi(r)$ is a compact fixed-dimensional surrogate feature derived from operator presence and normalized operator parameters.
The ordered recipe itself is still supplied to the search controller and Ranker, so order-sensitive reasoning is not claimed to be fully represented by $\psi(r)$ alone.
\newpage
\subsection{Search Pseudocode}
\label{app:search_pseudocode}

\begin{algorithm}[h]
    \caption{AutoSelection search via local edits and history-based reseeding.}
    \label{alg:search_pseudocode}
    \small
    \begin{algorithmic}[1]
        \Require Raw pool $\mathcal{D}_0$, base model $f_0$, recipe space $\mathcal{R}$, evaluation suite $\mathcal{E}$, evaluation budget $B$, stagnation patience $P$
        \Ensure Best discovered selected subset $S^\star$, generating recipe $r^\star$, and score $y^\star$
        \State Canonicalize $\mathcal{D}_0$ into $\mathcal{D}_0'$ and cache reusable task-, data-, and model-side signals
        \State Sample warmup recipes from $\mathcal{R}$, keep low-, medium-, and high-retention probes by retained-example ratio, and evaluate them to initialize $\mathcal{H}_3$
        \State Set $r_1^{\mathrm{seed}} \gets \arg\max_{(r, z, y) \in \mathcal{H}_3} y$ and initialize incumbent $r^\star,S^\star,y^\star$ from the best warmup evaluation
        \State Set $m \gets 1$, $t \gets |\mathcal{H}_3| + 1$, and $c \gets 0$
        \While{$t \le B$}
            \State $g_{t-1} \gets \mathrm{Summarizer}(\mathcal{H}_{t-1})$
            \State $\mathcal{C}_t \sim Q_{\mathrm{prop}}(\cdot \mid r_m^{\mathrm{seed}}, g_{t-1}, \mathcal{H}_{t-1})$
            \For{each $r' \in \mathcal{C}_t$}
                \State Execute $r'$ to compute $S(r')=\mathrm{Exec}(\mathcal{D}_0',r')$ and $z(r')=\phi(S(r'))$
                \State Encode $\psi(r')$ and estimate $\hat\mu_{t-1}(r'),\hat\sigma_{t-1}(r')=\mathrm{GP}_{t-1}(\psi(r'))$
                \State Compute $\rho_t(r')=f_{\mathrm{rank}}(r',z(r'),\hat\mu_{t-1}(r'),\hat\sigma_{t-1}(r'),\mathcal{H}_{t-1})$
            \EndFor
            \State Let the \emph{Ranker} choose $r_t \in \arg\max_{r' \in \mathcal{C}_t} \rho_t(r')$
            \State Set $S_t=S(r_t)$ and $z_t=z(r_t)$; fine-tune $f_0$ on $S_t$ and evaluate on $\mathcal{E}$ to obtain $y_t$
            \State Update $\mathcal{H}_t \gets \mathcal{H}_{t-1} \cup \{(r_t, z_t, y_t)\}$
            \If{$y_t > y^\star$}
                \State $r^\star \gets r_t$, $S^\star \gets S_t$, $y^\star \gets y_t$, $c \gets 0$
            \Else{} \State $c \gets c + 1$
            \EndIf
            \If{$c \ge P$}
                \State $r_{m+1}^{\mathrm{seed}} \sim Q_{\mathrm{seed}}(\cdot \mid \mathcal{H}_t)$, $m \gets m + 1$, $c \gets 0$ \Comment{Refresh the seed after stagnation}
            \EndIf
            \State $t \gets t + 1$
        \EndWhile
        \State \Return $S^\star$, $r^\star$, and $y^\star$
    \end{algorithmic}
\end{algorithm}

\subsection{Formal Notes and Search-Side Complexity}
\label{app:formal_notes}

\paragraph{Proof for the single-operator containment.}
Every recipe $r_{u,\theta}=((u,\theta))$ on the right-hand side of the
containment statement in Section~\ref{sec:method} is an element of
$\mathcal{R}$ by assumption. Therefore, maximizing over all recipes in
$\mathcal{R}$ cannot be worse than maximizing over this subset of length-one
recipes.

For a concrete example, a scoring-based selector $u$ with score $s$ and
parameter $\theta=\alpha$ maps a subset $\mathcal{D}\subseteq\mathcal{D}_0'$ to
\[
    u(\mathcal{D};\alpha)
    =
    \mathrm{Top}_{\lceil \alpha|\mathcal{D}|\rceil}(\mathcal{D};s).
\]
When used alone, this selector corresponds to the length-one recipe
$((u,\alpha))$.

\paragraph{Search-side complexity.}
Let $W$ be the number of warmup probes, $M=\max_t|\mathcal{C}_t|$ the maximum
number of sibling candidates generated per search step, and $L_{\max}$ the
maximum recipe length. AutoSelection performs exactly $B$ full
evaluations. Apart from these expensive evaluations, it executes at most
\[
     W + M(B-W)
\]
candidate recipes on the cached pool for subset materialization and state-vector
extraction, treating warmup probes as singleton candidate sets. If
$c_{o,\theta}(n)$ denotes the cost of applying operator $o$ with parameter
$\theta$ to a subset of size $n$, the
search-side execution cost is bounded by
\[
    O\!\left(
    \sum_{t=1}^{B}
    \sum_{r\in \mathcal{C}_t}
    \sum_{\ell=1}^{L(r)}
    c_{o_\ell,\theta_\ell}(|S_{\ell-1}|)
    \right),
\]
where $S_{\ell-1}$ is the intermediate subset before the $\ell$-th operator.
Exact GP refitting over $t$ evaluated recipes costs $O(t^3)$ per step, and
prediction over $|\mathcal{C}_t|$ candidates costs $O(|\mathcal{C}_t|t^2)$,
which is negligible in our setting because $B=15$. The dominant cost remains
\[
    \sum_{t=1}^{B}
    \left[
    C_{\mathrm{SFT}}(|S_t|)
    +
    C_{\mathrm{Eval}}(\mathcal{E})
    \right],
\]
namely fine-tuning and benchmark evaluation. The cold-start cache is therefore
important not because it removes full evaluations, but because it
prevents repeated recomputation of task-, data-, and model-side signals across
many candidate recipes.

The empirical accounting in Table~\ref{tab:compute_diag} is consistent with
this complexity picture. We summarize representative 15-evaluation search runs
and report the consistently available component-level accounting: accumulated
recipe execution time, combined SFT-and-evaluation time, and search-side LLM
time. The available records do not reliably split SFT training from benchmark
inference, so the two are reported jointly. Some logs contain broader
wall-clock totals that include additional overhead, but these fields are not
available uniformly across runs; Table~\ref{tab:compute_diag} therefore
uses the shared component fields only. Under this accounting, search-side LLM
calls remain small relative to full evaluation: LLM time is 0.37--0.61 hours,
whereas recipe execution plus SFT-and-evaluation accounts for 8.25--12.15 hours
over the same 15 evaluated recipes.

\begin{table}[h]
\centering
\caption{Representative component-level compute accounting for 15-evaluation search runs. All times are accumulated hours over the first 15 full evaluations.}
\label{tab:compute_diag}
\small
\setlength{\tabcolsep}{4.5pt}
\begin{tabular}{lrrrrr}
\toprule
Run & Evals & Recipe exec h & SFT+eval h & LLM h & Itemized total h \\
\midrule
A & 15 & 1.35 & 6.90 & 0.52 & 8.77 \\
B & 15 & 2.85 & 7.95 & 0.52 & 11.32 \\
C & 15 & 3.75 & 8.40 & 0.61 & 12.76 \\
D & 15 & 4.35 & 7.50 & 0.37 & 12.22 \\
\bottomrule
\end{tabular}
\end{table}
\newpage
\section{Search Diagnostics and Ablation Details}
\label{app:search_diagnostics}

\subsection{Random Select Curve}
\label{app:random_select_curve}
\begin{figure}[h]
    \centering
    \includegraphics[width=0.95\linewidth]{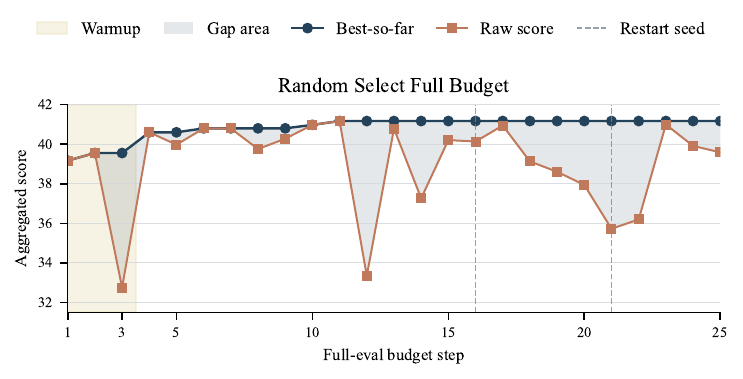}
    \caption{Extended-budget Random Select curve on the 1.5B setting. Shaded steps denote warmup probes, and dashed vertical markers denote restart seeds.}
    \label{fig:random_select_full_budget_curve}
\end{figure}

Random Select is included in Table~\ref{tab:search_ablation} as a policy-control ablation because it changes candidate selection rather than removing a single internal module.
After the same warmup and candidate-generation stages, this control chooses one candidate uniformly at random for full evaluation instead of using the GP prior and Ranker to select the next candidate.
Figure~\ref{fig:random_select_full_budget_curve} shows all 25 full evaluation records available in the Random Select run, including the three warmup probes and the later restart seeds.
Extending the budget gives Random Select more chances to improve its best result: the best-so-far score rises to 41.17 by step 11.
However, the raw curve remains highly volatile, with large drops after apparently strong evaluations and another collapse after the later restart.
Thus, additional random budget can raise the best-so-far curve, but it does not provide a stable search policy.

\subsection{State Vector Definition and Diagnostics}
\label{app:state_vector}

The state vector summarizes the realized subset $S$ after a candidate recipe has
been executed on the fixed reference pool $\mathcal{D}_0'$. We group the fields into task,
data, and model components,
\[
z(S)=\big[z^{\mathrm{task}}(S);z^{\mathrm{data}}(S);z^{\mathrm{model}}(S)\big].
\]
Table~\ref{tab:state_vector_fields} lists the fields used in our search.
SNAR denotes sparse-neuron activation rate.

\begin{table}[h]
\centering
\caption{State-vector fields used to summarize an executed candidate subset.}
\label{tab:state_vector_fields}
\scriptsize
\renewcommand{\arraystretch}{1.16}
\begin{tabularx}{\linewidth}{p{1.05cm}p{2.35cm}X X}
\toprule
\textbf{Group} & \textbf{Field} & \textbf{Computation} & \textbf{Meaning} \\
\midrule
\multirow{3}{*}{Task}
& \texttt{score\_mean}
& $\operatorname{mean}_{x,b}s_b(x)$ over $\mathcal{E}$
& Average MONA task relevance. \\
& \texttt{score\_std}
& $\operatorname{std}_{x,b}s_b(x)$
& Heterogeneity of task-relevance scores. \\
& \texttt{score\_per\_task}
& $\{\operatorname{mean}_{x\in S}s_b(x)\}_{b\in \mathcal{E}}$
& Benchmark-wise relevance profile. \\
\midrule
\multirow{2}{*}{Data}
& \texttt{retain\_ratio}
& $|S|/|\mathcal{D}_0'|$
& Retained-example scale change after filtering. \\
& \texttt{token\_ratio}
& $T(S)/T(\mathcal{D}_0')$, using whitespace-tokenized instruction and response text
& Training-token scale retained by the recipe. \\
\midrule
\multirow{3}{*}{Model}
& \texttt{distribution\_drift}
& $\|\mathrm{SNAR}(S)-\mathrm{SNAR}(\mathcal{D}_0')\|_2/\sqrt{D_{\mathrm{SAE}}}$
& Sparse-activation shift from the reference pool. \\
& \texttt{mean\_ifd}
& $\operatorname{mean}_{x\in S}s_{\mathrm{IFD}}(x) / \operatorname{mean}_{x\in \mathcal{D}_0'}s_{\mathrm{IFD}}(x)$
& Relative instruction-following difficulty. \\
& \texttt{mean\_varentropy}
& $\operatorname{mean}_{x\in S}V(x) / \operatorname{mean}_{x\in \mathcal{D}_0'}V(x)$
& Relative predictive-uncertainty complexity. \\
\bottomrule
\end{tabularx}
\end{table}

For the distribution-drift row, $\mathrm{SNAR}(\cdot)$ denotes a dataset-level
sparse-neuron activation-rate vector over cached SAE/MONA features. Let $A(x)$
be the active SAE feature-index set of sample $x$ and let $S_{\mathrm{valid}}$
be samples with cached sparse features. We compute
\[
    \mathrm{SNAR}_j(S)=
    \frac{1}{|S_{\mathrm{valid}}|}
    \sum_{x\in S_{\mathrm{valid}}}\mathbf{1}[j\in A(x)] .
\]
Thus, SNAR records how often each SAE feature appears in the subset, ignoring
activation magnitudes. This follows the use of sparse autoencoder features as
sparse activation coordinates in prior interpretability work
\citep{huben2024sparse} and in MONA task
relevance modeling~\citep{ma2025mona}. Here $s_{\mathrm{IFD}}$ is the IFD
score, $V(x)$ is sample-level varentropy, and $s_b(x)$ is the MONA similarity to
benchmark $b$. Candidate recipes are executed cheaply before SFT, so these
fields let the Ranker inspect realized subset scale, distribution shift,
difficulty, uncertainty, and task relevance before selecting one candidate for
full evaluation.

As an additional diagnostic, Table~\ref{tab:state_corr_diag} reports
correlations between saved state-vector fields and downstream full evaluation
scores over 107 state-score points from six complete search runs. The
strongest all-point Spearman correlations are token ratio, retain ratio, GSM8K
task relevance, and score mean, while model-side uncertainty and difficulty
fields are negatively associated with score in this set. A leave-one-out ridge
probe over the state vector gives Pearson 0.798 and Spearman 0.406. These
results support the use of state vectors as search context for screening
candidates, but they should not be read as evidence that state vectors are
standalone predictors that replace full evaluation.

\begin{table}[h]
\centering
\caption{State-feature correlations with downstream full evaluation score over 107 saved state-score points.}
\label{tab:state_corr_diag}
\small
\begin{tabular}{lrrr}
\toprule
State feature & All Spearman & Within-run Spearman & Run-centered Pearson \\
\midrule
\texttt{token\_ratio} & 0.420 & 0.436 & 0.716 \\
\texttt{retain\_ratio} & 0.374 & 0.362 & 0.676 \\
\texttt{score\_per\_task.gsm8k} & 0.343 & 0.274 & 0.675 \\
\texttt{mean\_varentropy} & -0.339 & -0.263 & -0.456 \\
\texttt{score\_mean} & 0.320 & 0.275 & 0.649 \\
\texttt{mean\_ifd} & -0.211 & -0.249 & -0.584 \\
\bottomrule
\end{tabular}
\end{table}

\subsection{GP Fit}
\label{app:gp_fit}

\begin{figure*}[h]
    \centering
    \includegraphics[width=\textwidth]{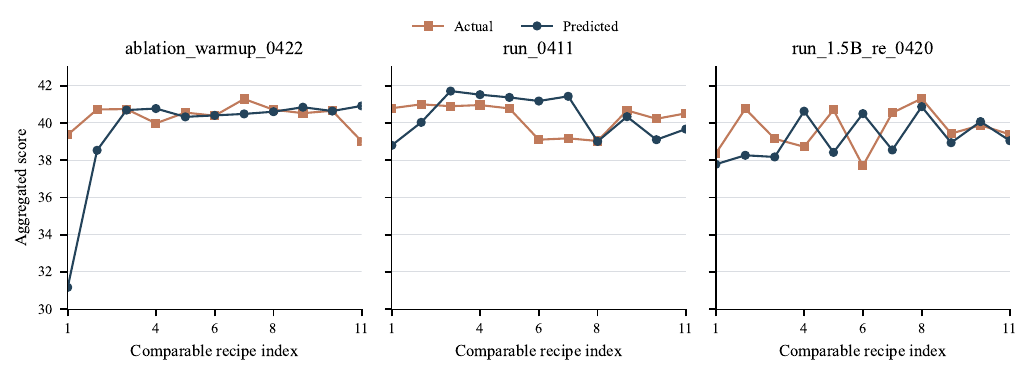}
    \caption{Rolling GP fit over the three analyzed runs. Each panel compares saved surrogate predictions with downstream full evaluation scores where both are available, then reindexes the comparable points onto the same x-axis.}
    \label{fig:gp_fit}
\end{figure*}

The GP surrogate is used to estimate candidate quality from recipe encodings
before full evaluation is run. Its role is to provide a
cheap historical prior over the search space, not to replace full
evaluation. Realized state vectors are used by the Ranker, not by the GP
surrogate. Figure~\ref{fig:gp_fit} shows the rolling fit between GP-predicted
scores and observed full evaluation scores over the analyzed search runs. The
predictions provide a coarse historical signal over evaluated recipes, but the
fit remains mixed and should not be treated as a replacement for full evaluation.
This is expected because the search has few observations and each full evaluation
is noisy. We therefore use the GP score only as a directional signal for
candidate prioritization while still relying on full evaluation for
final selection.

\newpage
\subsection{Operator-Composition Case Studies}
\label{app:operator_composition_cases}

\begin{figure*}[h]
    \centering
    \includegraphics[width=\textwidth]{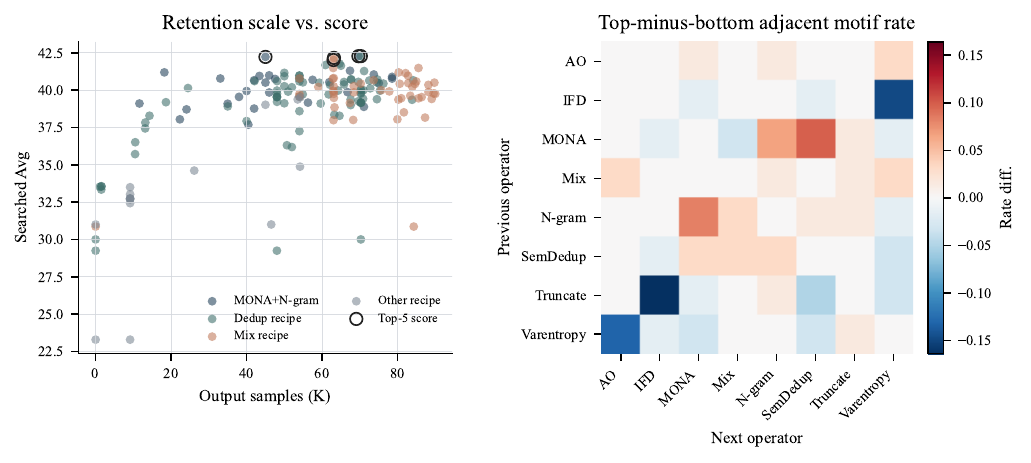}
    \caption{Run-level search diagnostics. Left: retained-example scale versus validation score. Right: adjacent-operator motif differences between top- and bottom-tertile recipe records.}
    \label{fig:search_diagnostics_panel}
\end{figure*}

Before the paired case studies, we first summarize two run-level
diagnostics over 181 full evaluation recipe records from nine evaluated search
runs. Figure~\ref{fig:search_diagnostics_panel} visualizes both the
retained-example-scale scatter and the adjacent-operator motif comparison. These
diagnostics are descriptive, not causal ablations, but they help contextualize
why retained scale and operator composition are both treated as search variables.

\paragraph{Retention scale is not sufficient.}
Table~\ref{tab:retention_bins_diag} shows that retained subset scale matters,
but does not explain downstream score by itself. For example, the 40K--60K
output range contains 51 records with a 12.98-point score range, while the
60K--80K range contains 67 records with a 12.27-point range. Thus, retention
scale is a useful state variable, but recipes with comparable realized sizes can
still differ substantially.

\begin{table}[h]
\centering
\caption{Binned relationship between retained-example scale and validation score over 181 full evaluation recipe records.}
\label{tab:retention_bins_diag}
\small
\begin{tabular}{lrrrrr}
\toprule
Retained examples & N & Min & Mean & Max & Range \\
\midrule
0--20K & 26 & 23.30 & 33.42 & 41.20 & 17.90 \\
20--40K & 8 & 34.62 & 38.74 & 40.79 & 6.17 \\
40--60K & 51 & 29.26 & 39.34 & 42.23 & 12.98 \\
60--80K & 67 & 30.01 & 40.03 & 42.29 & 12.27 \\
80--100K & 29 & 30.87 & 39.48 & 41.49 & 10.62 \\
\bottomrule
\end{tabular}
\end{table}

\paragraph{Operator motifs recur in stronger records.}
We split the same records into top- and bottom-tertile groups by
validation score and compare adjacent operator pairs. Table~\ref{tab:motif_diag}
lists the largest positive differences. The enrichment of composed motifs such
as MONA$\rightarrow$SemDedup and N-gram$\rightarrow$MONA supports the
fixed-pool data recipe search framing: high-scoring records are not explained only by the
presence of one isolated operator. These motif counts should be read as
retrospective search diagnostics rather than proof that any pair is universally
optimal.

\begin{table}[h]
\centering
\caption{Adjacent-operator motif differences between top- and bottom-tertile full evaluation records.}
\label{tab:motif_diag}
\small
\begin{tabular}{lrrr}
\toprule
Adjacent pair & Top count & Bottom count & Delta \\
\midrule
MONA$\rightarrow$SemDedup & 11 & 5 & +6 \\
N-gram$\rightarrow$MONA & 10 & 5 & +5 \\
MONA$\rightarrow$N-gram & 7 & 3 & +4 \\
SemDedup$\rightarrow$Mix & 4 & 2 & +2 \\
SemDedup$\rightarrow$N-gram & 4 & 2 & +2 \\
\bottomrule
\end{tabular}
\end{table}

The search runs provide qualitative evidence that operator effects are context-dependent even when retained scale is roughly controlled.
Table~\ref{tab:operator_composition_cases} therefore reports paired recipes with comparable realized subset sizes, so the comparisons focus on recipe structure after coarse retained-scale matching.
These comparisons should be read as search-run case studies, not isolated one-factor ablations: they support the fixed-pool data recipe search formulation by showing that order, thresholds, deduplication choices, and seed composition can change subset quality beyond retained scale alone.
Several patterns are visible from the scale-matched pairs: deduplication thresholds are not monotonic, task-relevance and diversity filters interact with their order, and mixing is most useful when followed by an additional diversity-oriented operator rather than used alone.
For compactness, the recipe strings use operator names consistently: N-gram, random-$k$, and Mix denote N-gram top-$k$, Random top-$k$, and Mix, respectively.

\begin{table*}[h]
\centering
\caption{Scale-matched recipe case studies from the search runs. Scores are validation averages on the 1.5B setting, and $n$ is the realized number of retained examples.}
\label{tab:operator_composition_cases}
\scriptsize
\setlength{\tabcolsep}{4pt}
\renewcommand{\arraystretch}{1.12}
\begin{tabularx}{\textwidth}{@{}C{0.65cm} C{0.95cm} X C{1.05cm} C{0.95cm}@{}}
\toprule
Case & Result & Recipe & $n$ & Avg \\
\midrule
\multirow{2}{*}{1} & High & MONA(0.70)$\rightarrow$SemDedup(0.73) & 71.1K & 41.29 \\
 & Low & MONA(0.65)$\rightarrow$SemDedup(0.70) & 67.6K & 39.01 \\
\addlinespace[0.25em]
\multirow{2}{*}{2} & High & SemDedup(0.75)$\rightarrow$random-$k$(54K) & 54.0K & 40.78 \\
 & Low & SemDedup(0.85)$\rightarrow$random-$k$(54K) & 54.0K & 38.60 \\
\addlinespace[0.25em]
\multirow{2}{*}{3} & High & MONA(0.87)$\rightarrow$Varentropy(0.87)$\rightarrow$SemDedup(0.79)$\rightarrow$random-$k$(50K) & 50.0K & 41.17 \\
 & Low & MONA(0.85)$\rightarrow$Varentropy(0.85)$\rightarrow$SemDedup(0.81)$\rightarrow$random-$k$(50K) & 50.0K & 39.13 \\
\addlinespace[0.25em]
\multirow{2}{*}{4} & High & N-gram(0.90)$\rightarrow$MONA(0.85)$\rightarrow$SemDedup(0.88)$\rightarrow$random-$k$(45K) & 45.0K & 42.23 \\
 & Low & N-gram(0.85)$\rightarrow$MONA(0.80)$\rightarrow$SemDedup(0.85)$\rightarrow$random-$k$(42K) & 42.0K & 38.78 \\
\addlinespace[0.25em]
\multirow{2}{*}{5} & High & random-$k$(80K)$\rightarrow$MONA(0.92)$\rightarrow$N-gram(0.92) & 69.6K & 42.28 \\
 & Low & random-$k$(80K)$\rightarrow$N-gram(0.90)$\rightarrow$MONA(0.90) & 67.4K & 40.80 \\
\addlinespace[0.25em]
\multirow{2}{*}{6} & High & random-$k$(68K)$\rightarrow$SemDedup(0.80)$\rightarrow$N-gram(0.90) & 60.9K & 41.69 \\
 & Low & random-$k$(59.6K)$\rightarrow$SemDedup(0.82) & 59.3K & 40.51 \\
\addlinespace[0.25em]
\multirow{2}{*}{7} & High & random-$k$(48K)$\rightarrow$SemDedup(1600, 0.96)$\rightarrow$MONA(0.58)$\rightarrow$Mix & 63.1K & 42.17 \\
 & Low & random-$k$(48K)$\rightarrow$SemDedup(1600, 0.96)$\rightarrow$MONA(0.70)$\rightarrow$Mix & 63.3K & 38.85 \\
\addlinespace[0.25em]
\bottomrule
\end{tabularx}
\end{table*}

\subsection{Search Stability}
\label{app:seed_stability}

Table~\ref{tab:seed_stability} gives the full three-run stability summary used in Section~\ref{subsec:seed_stability}. The main text reports the median run in the main result table and then gives the three selected scores, mean, and range in the stability discussion, while this appendix keeps the curated budget positions for reproducibility.

\begin{table*}[h]
\centering
\caption{Three-run search stability under the 15 full evaluation budget. Benchmark columns report the GPQA, GSM8K, BBH, and MMLU scores of the recipe that attains Best@15 in each run.}

\label{tab:seed_stability}
{
\renewcommand{\arraystretch}{1.08}
\begin{tabular}{lccccccc}
\toprule
Run & Pts & \shortstack{Best@15} & GPQA & GSM8K & BBH & MMLU & \shortstack{Early Mean @3} \\
\midrule
Run 1 & 15 & 42.23 & \textbf{29.02} & 54.59 & \textbf{30.00} & 55.33 & 36.21 \\
Run 2 & 15 & \textbf{42.28} & 24.55 & \textbf{58.00} & 29.02 & 57.56 & 37.16 \\
Run 3 & 15 & 41.69 & 26.12 & 53.68 & 28.91 & \textbf{58.06} & \textbf{37.42} \\
\bottomrule
\end{tabular}
}
\end{table*}






\subsection{Recipe Ranking Case Study}
\label{app:recipe_rank_case}

Table~\ref{tab:recipe_rank_case_compact} gives the sibling-candidate ranking study used as diagnostic support for the search-side ablation analysis in Section~\ref{subsec:search_ablations}.
Each row in Table~\ref{tab:recipe_rank_case_compact} is one sibling candidate from a randomly sampled 1.5B decision point.
As a compact summary of the same five audited decision points,
Table~\ref{tab:ranker_diag} reports Hit@1, Hit@2, MRR, and the average actual
rank of the Ranker's top-1 choice. The Ranker places the true best candidate at
rank 1 in 2/5 cases and within the top 2 in all 5 cases. This is a
counterfactual diagnostic rather than a large-scale ranking benchmark, but it
supports the claim that the Ranker helps allocate the full evaluation budget
toward promising candidates.

\begin{table}[h]
\centering
\caption{Compact Ranker audit metrics over five audited decision points.}
\label{tab:ranker_diag}
\small
\begin{tabular}{lr}
\toprule
Metric & Value \\
\midrule
Audited decision points & 5 \\
Hit@1 & 2/5 \\
Hit@2 & 5/5 \\
MRR & 0.70 \\
Mean actual rank of Ranker top-1 & 1.8 \\
\bottomrule
\end{tabular}
\end{table}

\begin{table}[h]
\centering
\caption{LLM Ranker audit on five randomly selected 1.5B decision points. Each decision point contains sibling candidate recipes that were retrospectively evaluated; candidates are anonymized as V1--V5 and rows are ordered by the LLM rank. Lower actual rank is better.}
\label{tab:recipe_rank_case_compact}
{
\begin{tabular}{lcccc}
\toprule
Decision & Variant & LLM rank & Actual rank & Avg score \\
\midrule
Decision 1 & V3 & 1 & 2 & 39.95 \\
Decision 1 & V1 & 2 & 1 & 41.96 \\
Decision 1 & V2 & 3 & 4 & 36.84 \\
Decision 1 & V4 & 4 & 3 & 37.80 \\
Decision 1 & V5 & 5 & 5 & 32.19 \\
\midrule
Decision 2 & V1 & 1 & 1 & 40.62 \\
Decision 2 & V4 & 2 & 3 & 39.47 \\
Decision 2 & V3 & 3 & 2 & 39.90 \\
Decision 2 & V5 & 4 & 5 & 38.65 \\
Decision 2 & V2 & 5 & 4 & 39.17 \\
\midrule
Decision 3 & V3 & 1 & 2 & 41.07 \\
Decision 3 & V2 & 2 & 1 & 41.43 \\
Decision 3 & V4 & 3 & 5 & 37.04 \\
Decision 3 & V1 & 4 & 3 & 40.33 \\
Decision 3 & V5 & 5 & 4 & 38.57 \\
\midrule
Decision 4 & V2 & 1 & 3 & 38.72 \\
Decision 4 & V1 & 2 & 1 & 40.79 \\
Decision 4 & V5 & 3 & 4 & 35.42 \\
Decision 4 & V3 & 4 & 5 & 30.67 \\
Decision 4 & V4 & 5 & 2 & 38.78 \\
\midrule
Decision 5 & V1 & 1 & 1 & 42.20 \\

Decision 5 & V5 & 2 & 2 & 41.69 \\

Decision 5 & V3 & 3 & 3 & 39.33 \\

Decision 5 & V2 & 4 & 5 & 33.16 \\

Decision 5 & V4 & 5 & 4 & 37.22 \\
\bottomrule
\end{tabular}
}
\end{table}

\newpage

\section{Additional Empirical Analyses and Baseline Boundaries}
\label{app:additional_analyses}

\subsection{Comparison Boundary for LLM-Driven Data-Processing Agents}
\label{app:llm_autodp}

We do not include LLM-AutoDP in the main quantitative table because it does not match the controlled fixed-pool data recipe search protocol used in this paper.
LLM-AutoDP relies on external LLM instructions embedded in data-processing modules and was designed around strong domain-specific assumptions.
We nevertheless ran an early Qwen2.5-1.5B pilot to understand whether it could serve as a neighboring LLM-driven baseline in our fixed-pool setting.
The evaluated pilot output is reported in Table~\ref{tab:llm_autodp_pilot}.

\begin{table}[h]
\centering
\caption{Early LLM-AutoDP pilot result on the Qwen2.5-1.5B setting. Scores are reported on the same in-distribution validation suite used for the main search objective.}
\label{tab:llm_autodp_pilot}
{\small
\setlength{\tabcolsep}{4.2pt}
\renewcommand{\arraystretch}{1.08}
\begin{tabular}{lccccc}
\toprule
Method & GPQA & GSM8K & BBH & MMLU & \textbf{Avg} \\
\midrule
LLM-AutoDP pilot & 16.74 & 16.07 & 22.17 & 48.94 & 25.98 \\
\bottomrule
\end{tabular}
}
\end{table}

We stopped this pilot before running a complete repeated baseline for two practical reasons.
First, applying LLM-AutoDP API calls over a large instruction pool introduced prohibitive wall-clock latency and API overhead compared with the fixed-pool operators used in AutoSelection.
Second, the available implementation was not domain-neutral: several LLM-facing modules were written for medical-domain data selection.
For example, the LLM optimizer and filtering prompts contain instructions that favor medically relevant samples, and other prompt-based filters ask the LLM to judge data quality through that domain-specific lens.
When moved to our mixed instruction pool, these built-in priors can select for the wrong notion of relevance rather than for GPQA/GSM8K/BBH/MMLU performance.

This pilot therefore serves mainly as a boundary case for the comparison and clarifies why the fixed-pool boundary in the main text is useful.
Once an LLM is placed directly inside the data-selection stage, the observed result can depend on generator quality, prompt design, domain-specific judging criteria, and API behavior in addition to the underlying data recipe.
Highly customized prompts can help in their intended domain, but they also introduce brittle priors when transferred to a different pool, as seen in the medical-relevance prompts above.
AutoSelection instead keeps LLMs on the search side, where they summarize evaluated histories and rank grounded recipe edits without rewriting, augmenting, or individually judging every training sample.
This design isolates the measured gains as much as possible to choices over grounded operators on a fixed raw pool, rather than to prompt engineering or newly introduced samples.

\subsection{Cross-Scale Transfer Checks}
\label{app:transferability}

The 7B transfer check is used as a supporting analysis, not as the main evidence for AutoSelection.
We take several recipes discovered in the 1.5B search, including recipes that are slightly weaker under the 1.5B objective, and evaluate the corresponding selected subsets on a larger 7B model.
As shown in Table~\ref{tab:transfer_7b}, the observed behavior is trend-level rather than exact: recipes that are strong at 1.5B tend to remain competitive, but the ordering is not perfectly preserved.
This supports a cautious interpretation of transferability: small-model recipe search can reveal useful data-construction motifs, but larger-model validation remains necessary before making final claims.

\begin{table*}[h]
\centering
\caption{Transfer from 1.5B recipe search to 7B evaluation after excluding the full-data baseline. Ranks are recomputed over the seven transferred recipes.}
\label{tab:transfer_7b}
{\small
\setlength{\tabcolsep}{3.0pt}
\begin{tabular*}{\textwidth}{@{}l@{\extracolsep{\fill}}ccccccccc@{}}
\toprule
\multirow{2}{*}{Recipe}
& \multicolumn{5}{c}{Cross-scale ranking}
& \multicolumn{4}{c}{7B validation scores} \\
\cmidrule(lr){2-6}
\cmidrule(l){7-10}
& 1.5B score & 1.5B rank & 7B score & 7B rank & Rank change
& GPQA & GSM8K & BBH & MMLU \\
\midrule
Recipe 1 & \underline{41.96} & \underline{2} & \textbf{51.92} & \textbf{1} & +1
& 19.64 & \textbf{76.19} & \underline{41.63} & \textbf{70.22} \\
Recipe 2 & \textbf{42.23} & \textbf{1} & \underline{51.90} & \underline{2} & -1
& 20.09 & \underline{74.91} & \textbf{43.37} & \underline{69.22} \\
Recipe 3 & 36.29 & 4 & 51.76 & 3 & +1
& \textbf{25.00} & 71.95 & \underline{41.63} & 68.44 \\
Recipe 4 & 34.79 & 5 & 50.33 & 4 & +1
& \underline{20.98} & 73.31 & 38.48 & 68.56 \\
Recipe 5 & 39.06 & 3 & 49.22 & 5 & -2
& 18.53 & 73.09 & 37.50 & 67.78 \\
Recipe 6 & 25.74 & 7 & 44.54 & 6 & +1
& 14.96 & 65.96 & 33.48 & 63.78 \\
Recipe 7 & 25.90 & 6 & 39.47 & 7 & -1
& 13.84 & 56.48 & 29.46 & 58.11 \\
\bottomrule
\end{tabular*}
}
\end{table*}
\newpage
\subsection{Metric-Distribution Comparison}
\label{app:metric_distributions}

\begin{figure*}[t]
\centering
\scriptsize
\setlength{\tabcolsep}{1.2pt}
\renewcommand{\arraystretch}{0.86}
\begin{tabular}{@{}lcccccc@{}}
 & MONA-GPQA & MONA-GSM8K & MONA-BBH & MONA-MMLU & IFD & Entropy \\
Recipe 1 &
\includegraphics[width=0.145\textwidth]{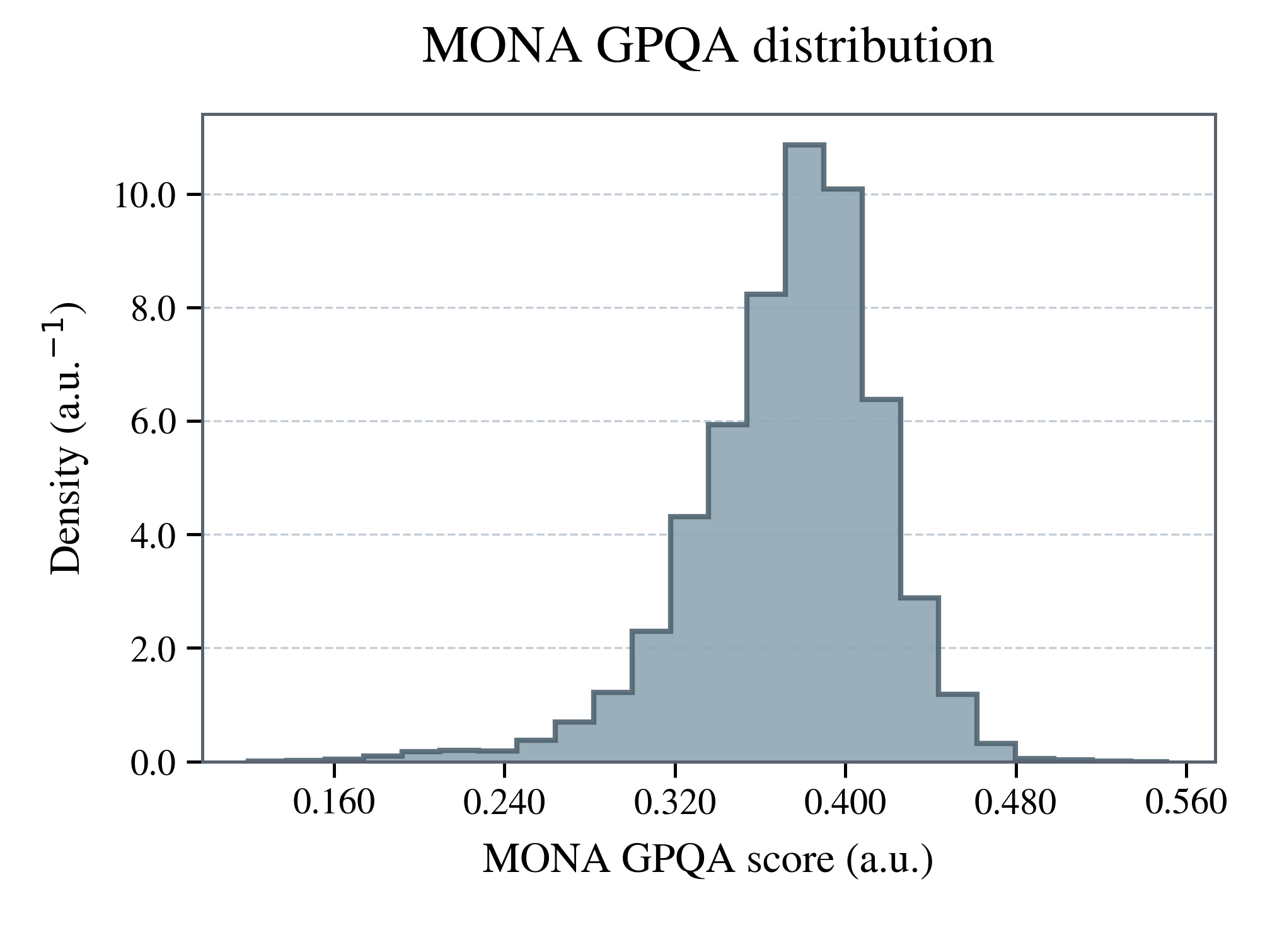} &
\includegraphics[width=0.145\textwidth]{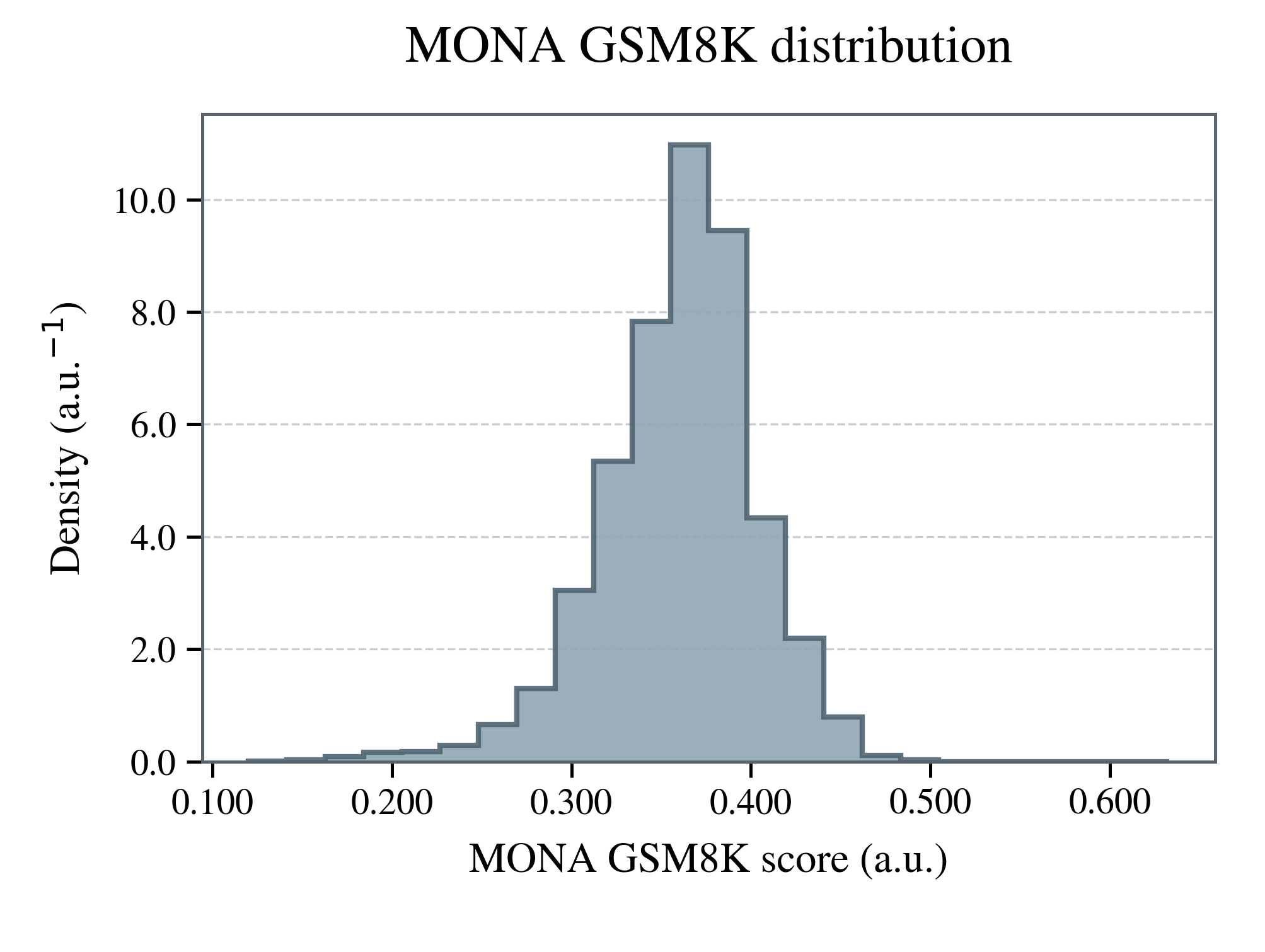} &
\includegraphics[width=0.145\textwidth]{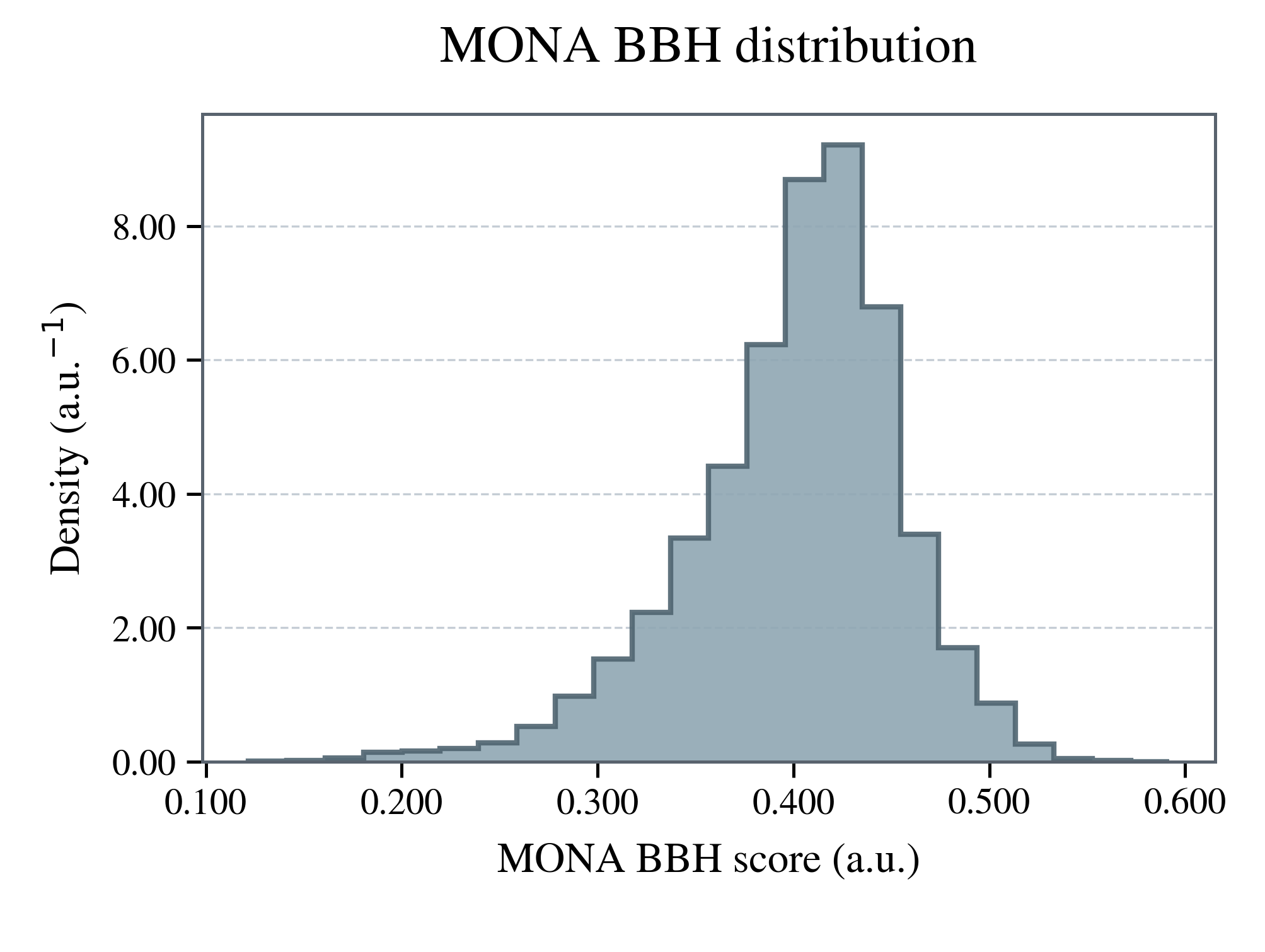} &
\includegraphics[width=0.145\textwidth]{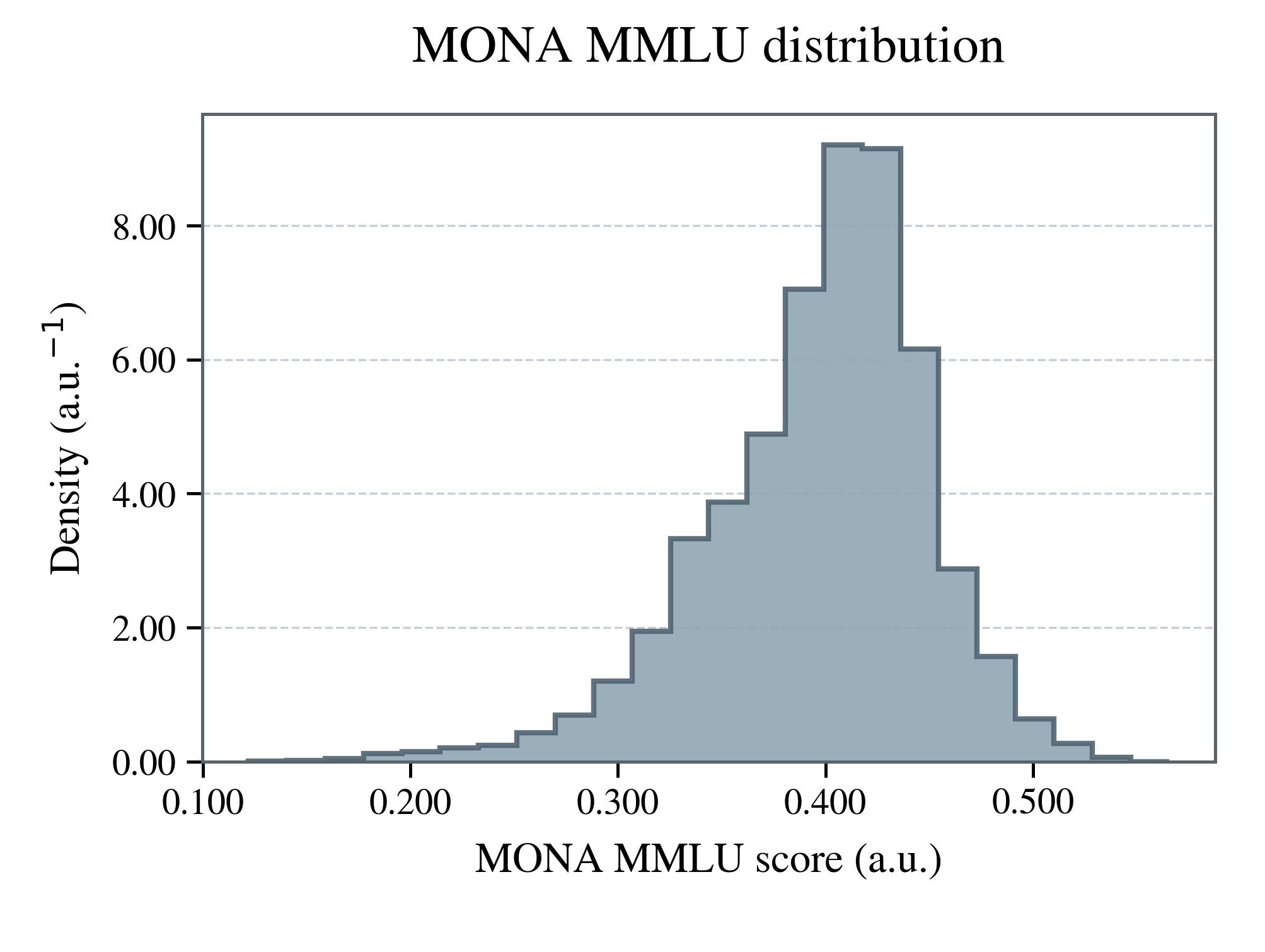} &
\includegraphics[width=0.145\textwidth]{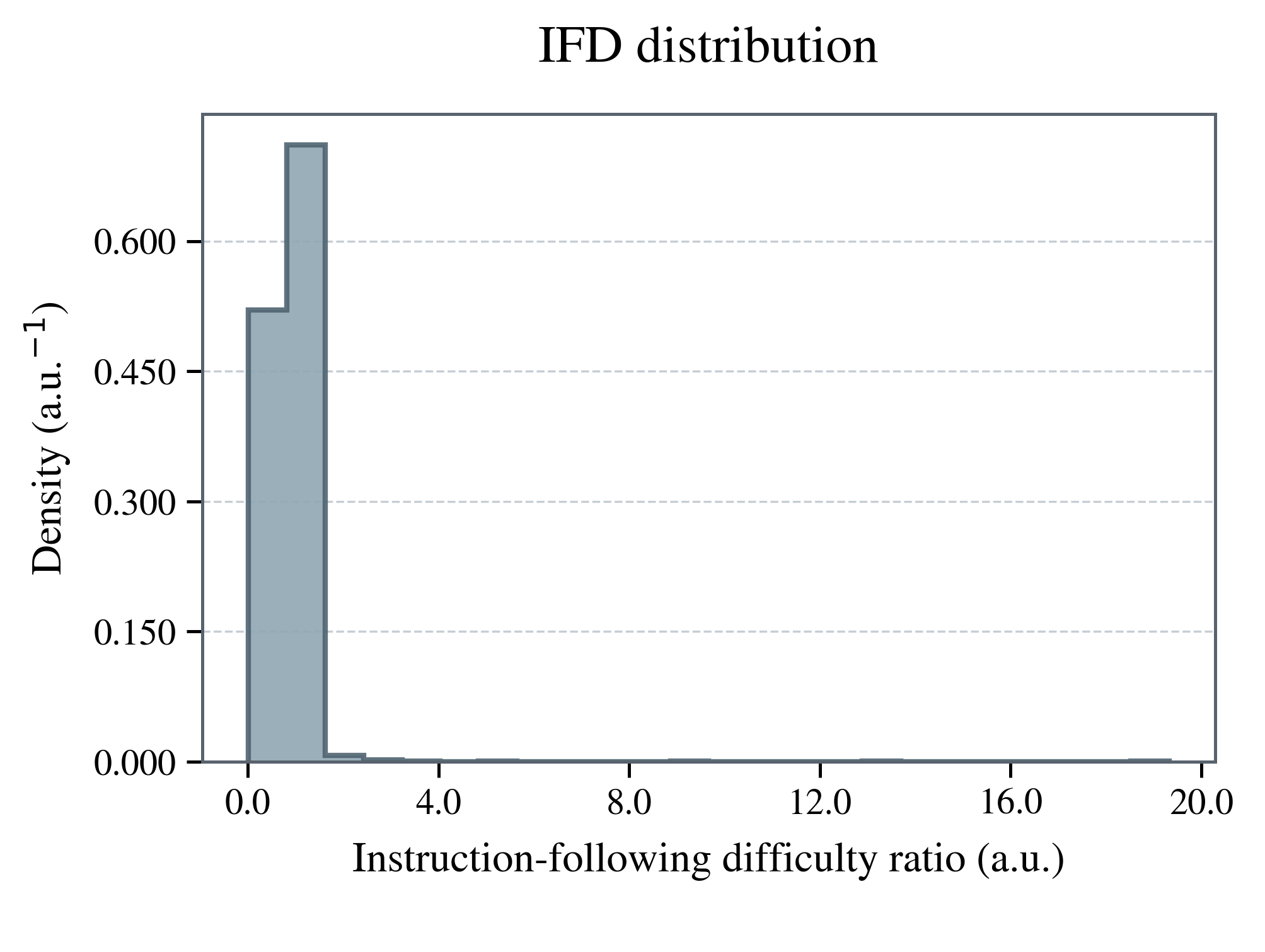} &
\includegraphics[width=0.145\textwidth]{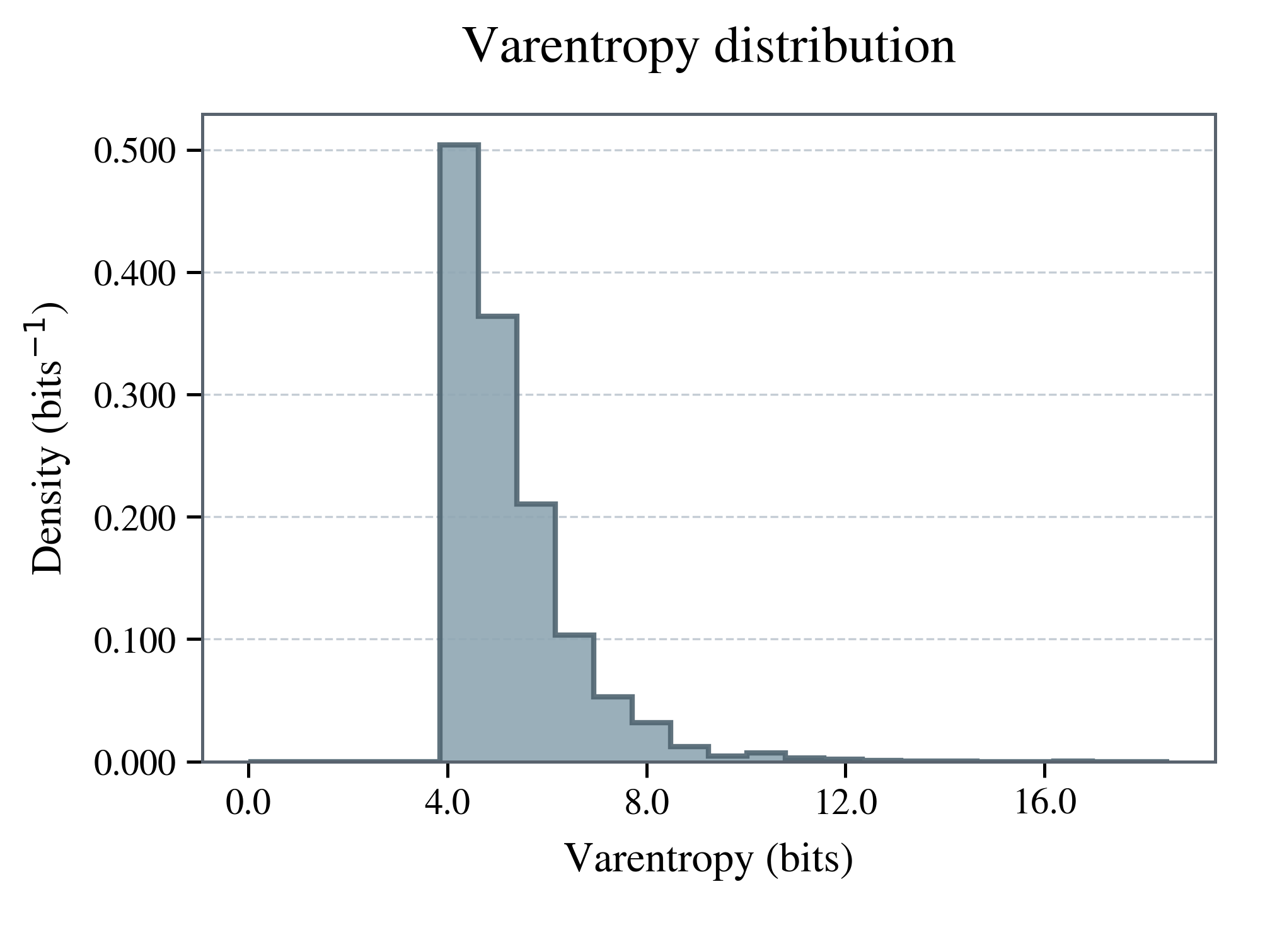} \\
Recipe 2 &
\includegraphics[width=0.145\textwidth]{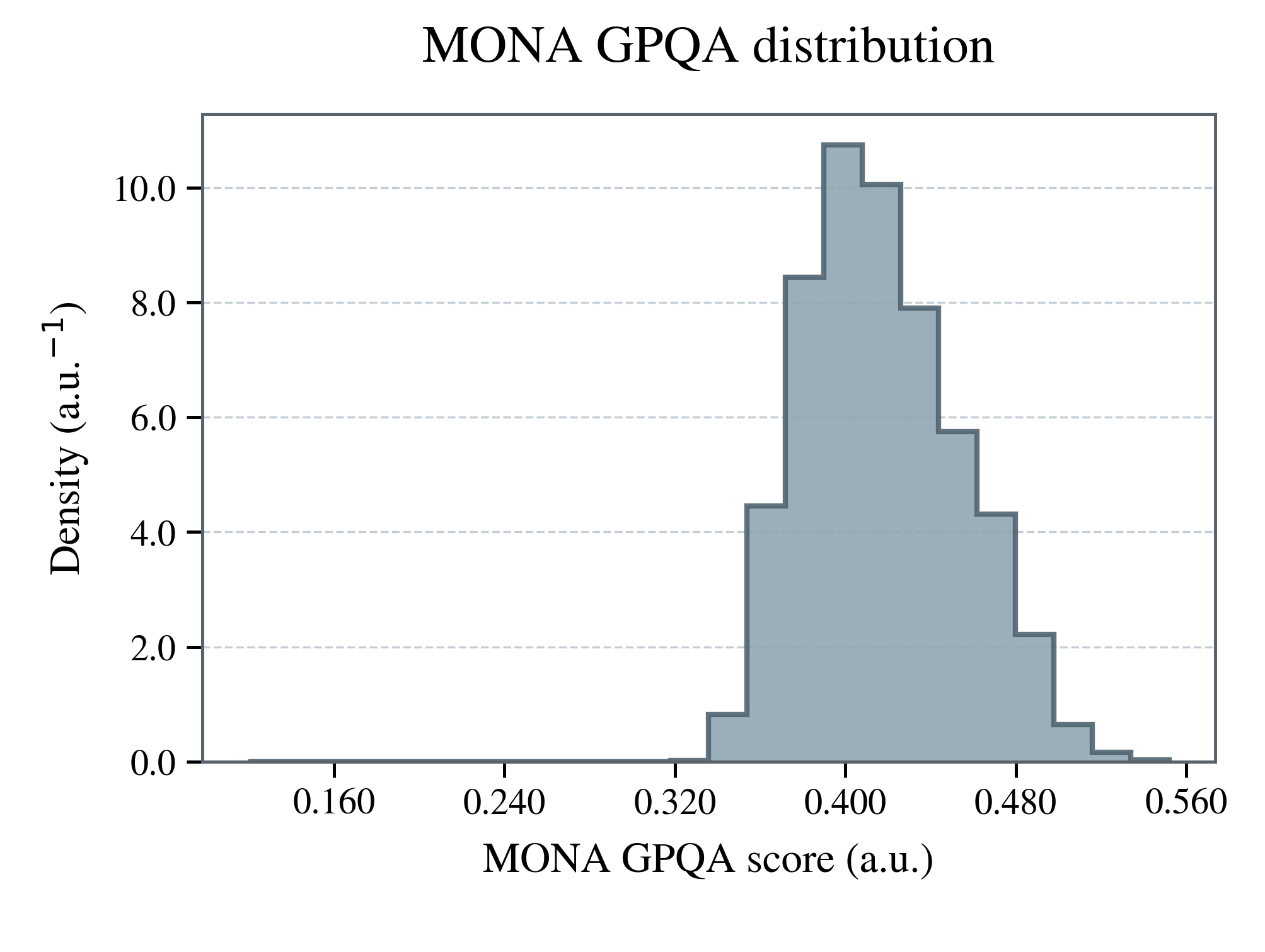} &
\includegraphics[width=0.145\textwidth]{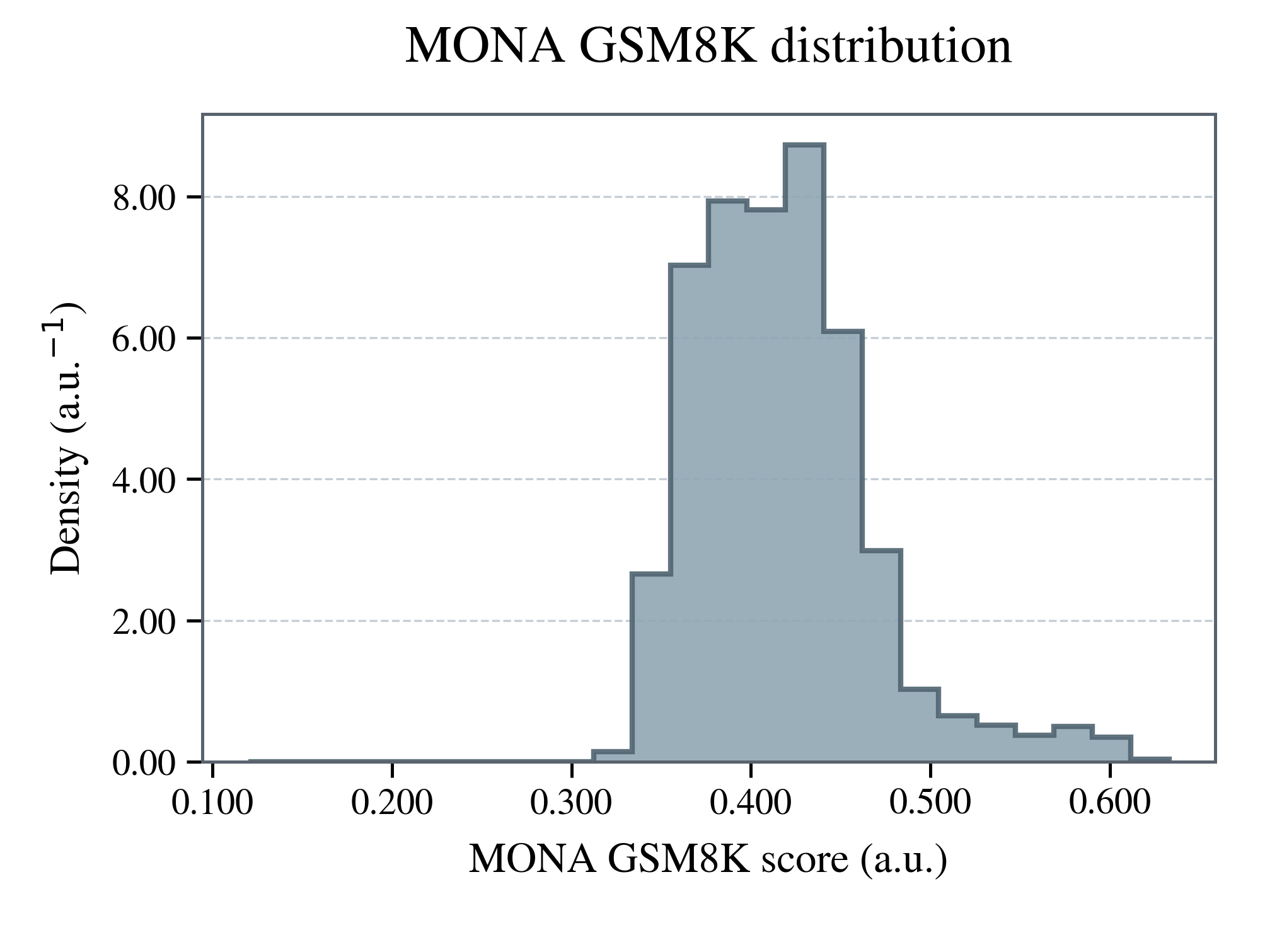} &
\includegraphics[width=0.145\textwidth]{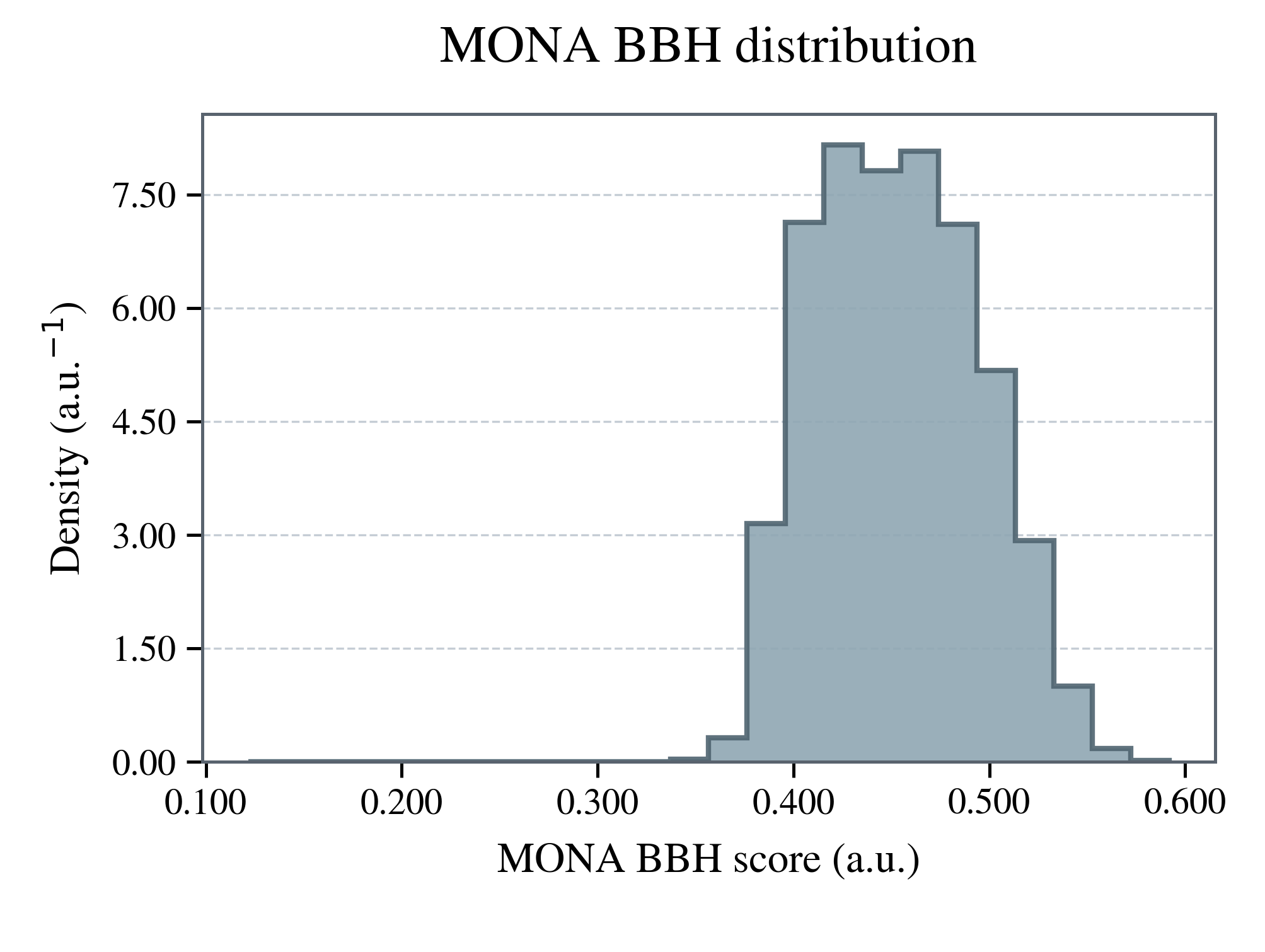} &
\includegraphics[width=0.145\textwidth]{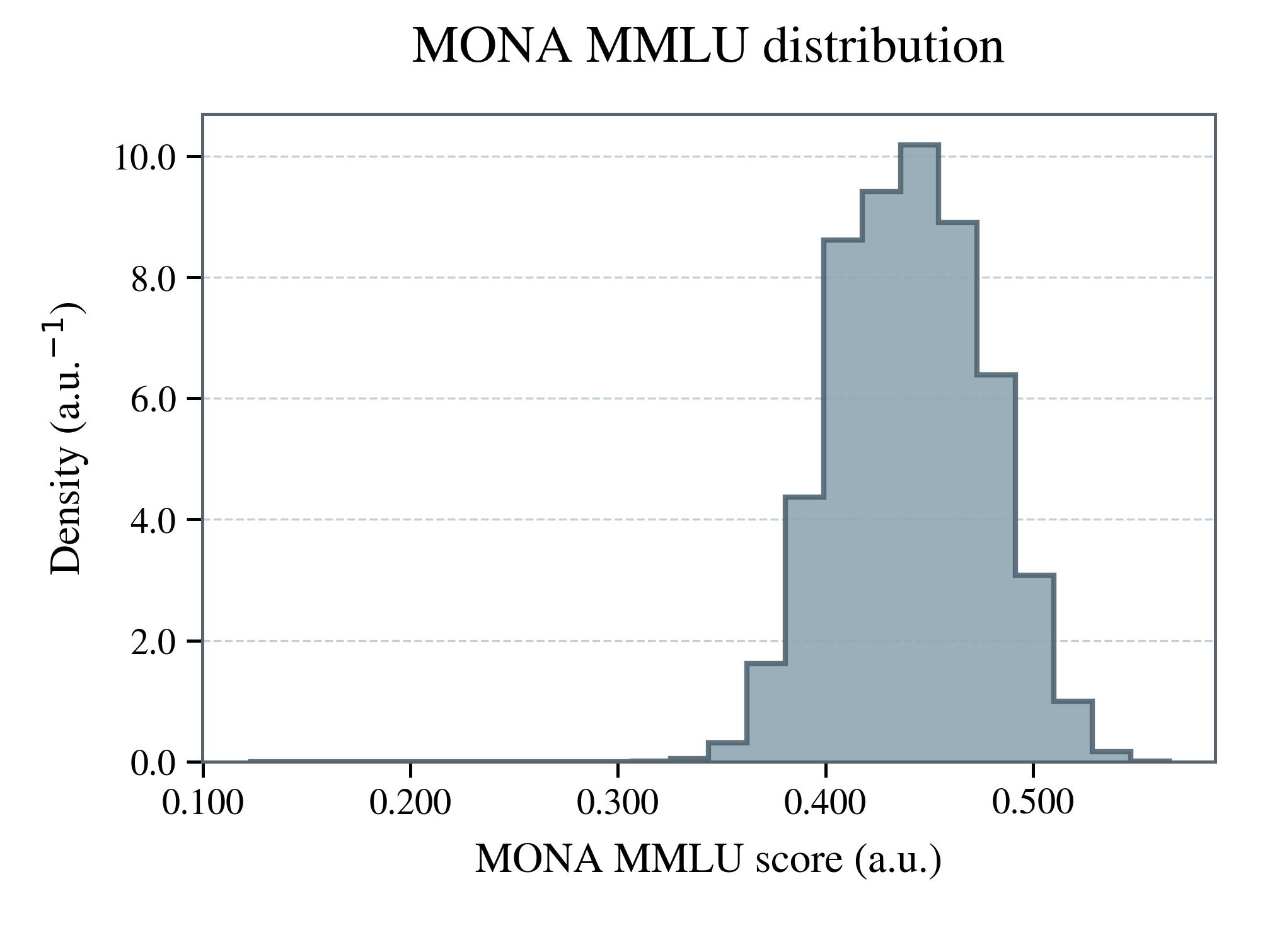} &
\includegraphics[width=0.145\textwidth]{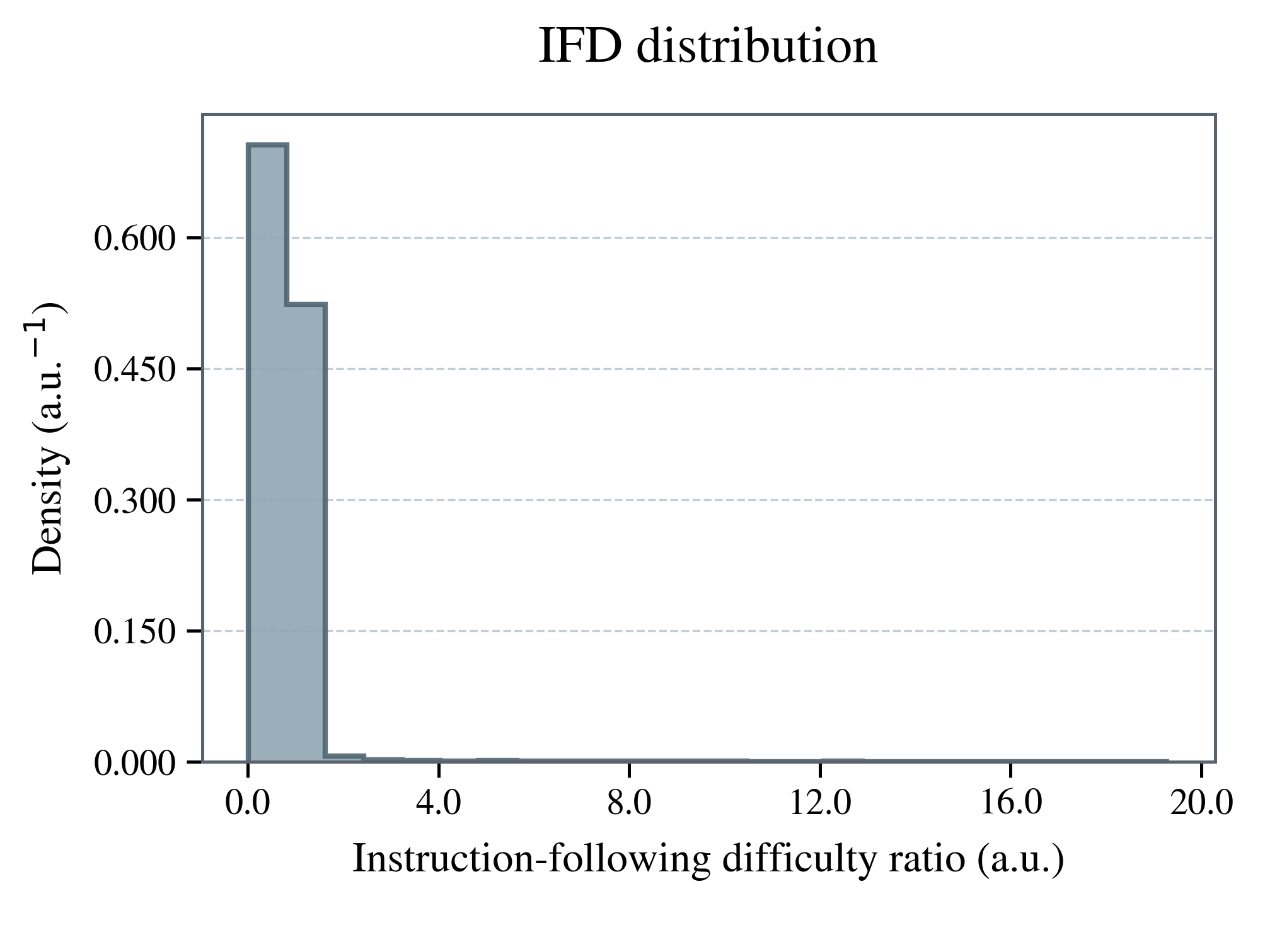} &
\includegraphics[width=0.145\textwidth]{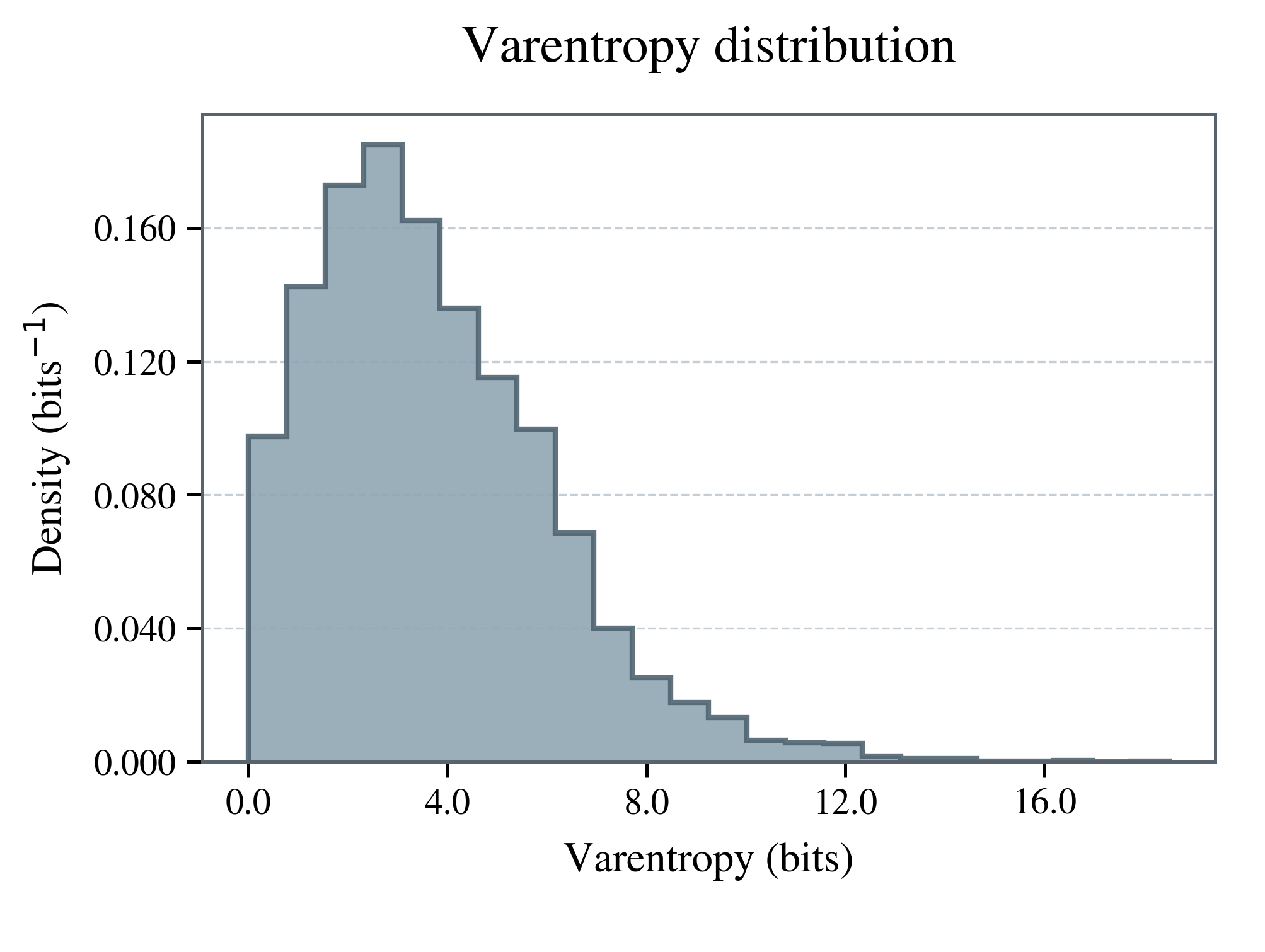} \\
Recipe 3 &
\includegraphics[width=0.145\textwidth]{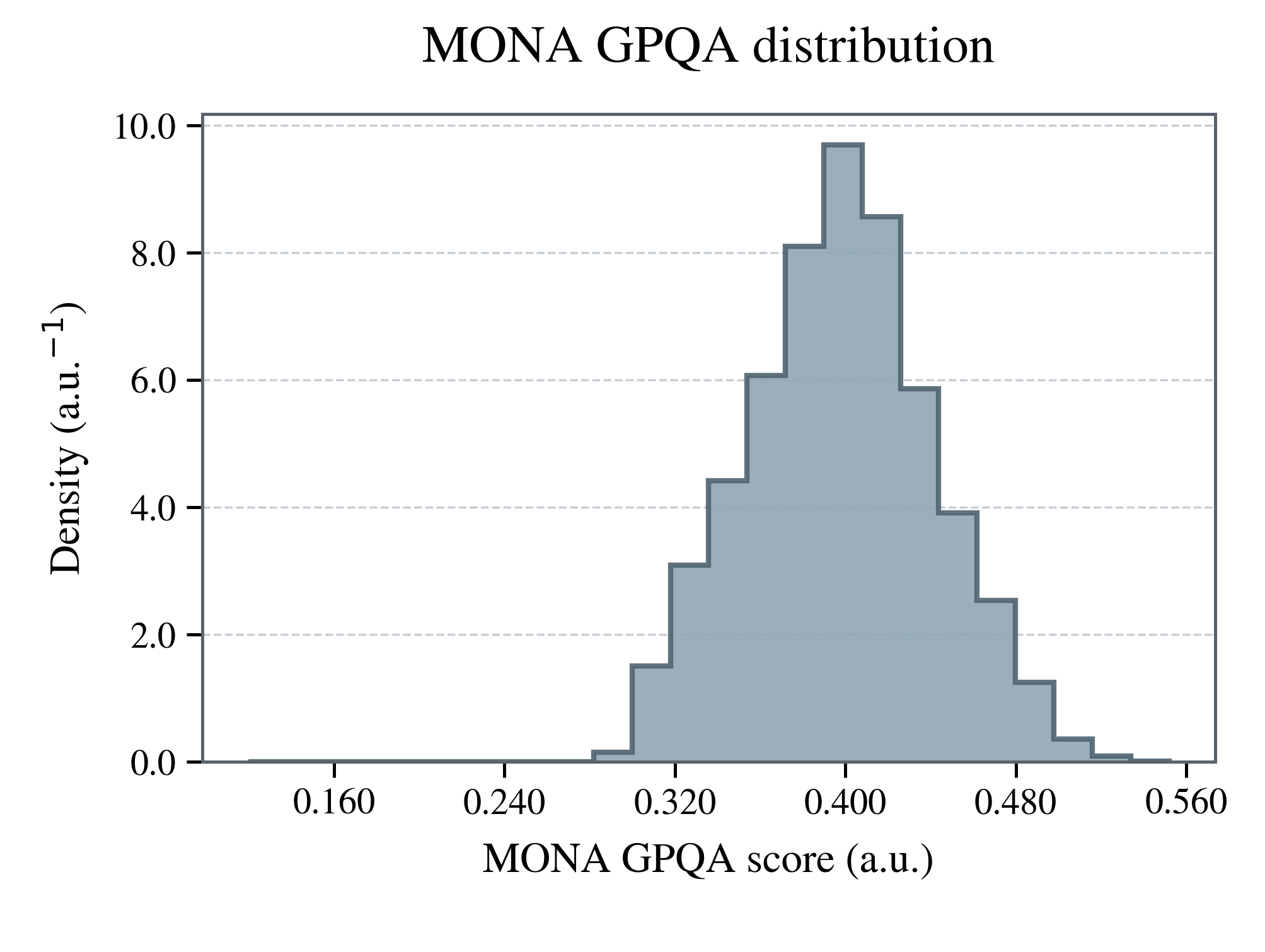} &
\includegraphics[width=0.145\textwidth]{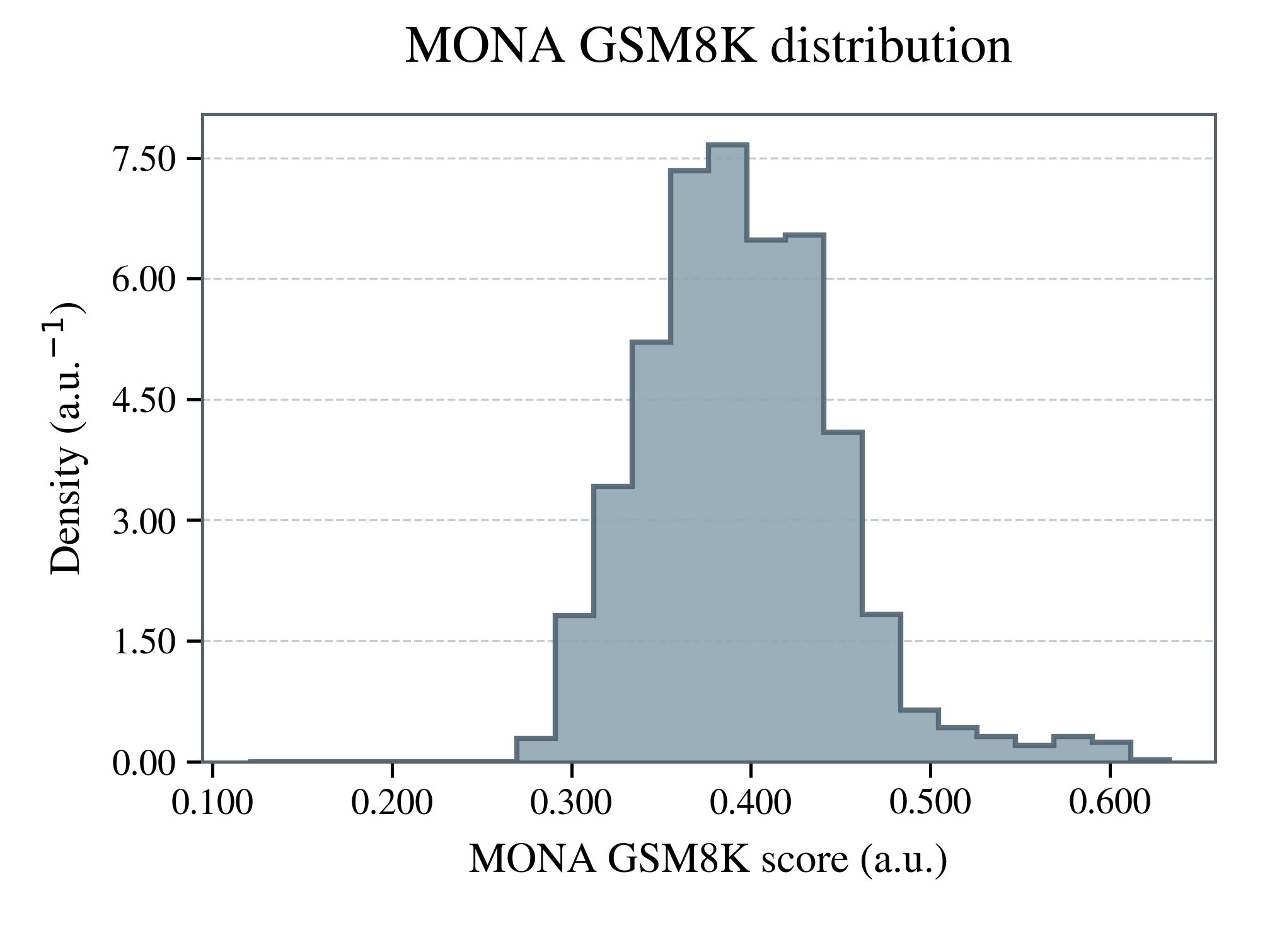} &
\includegraphics[width=0.145\textwidth]{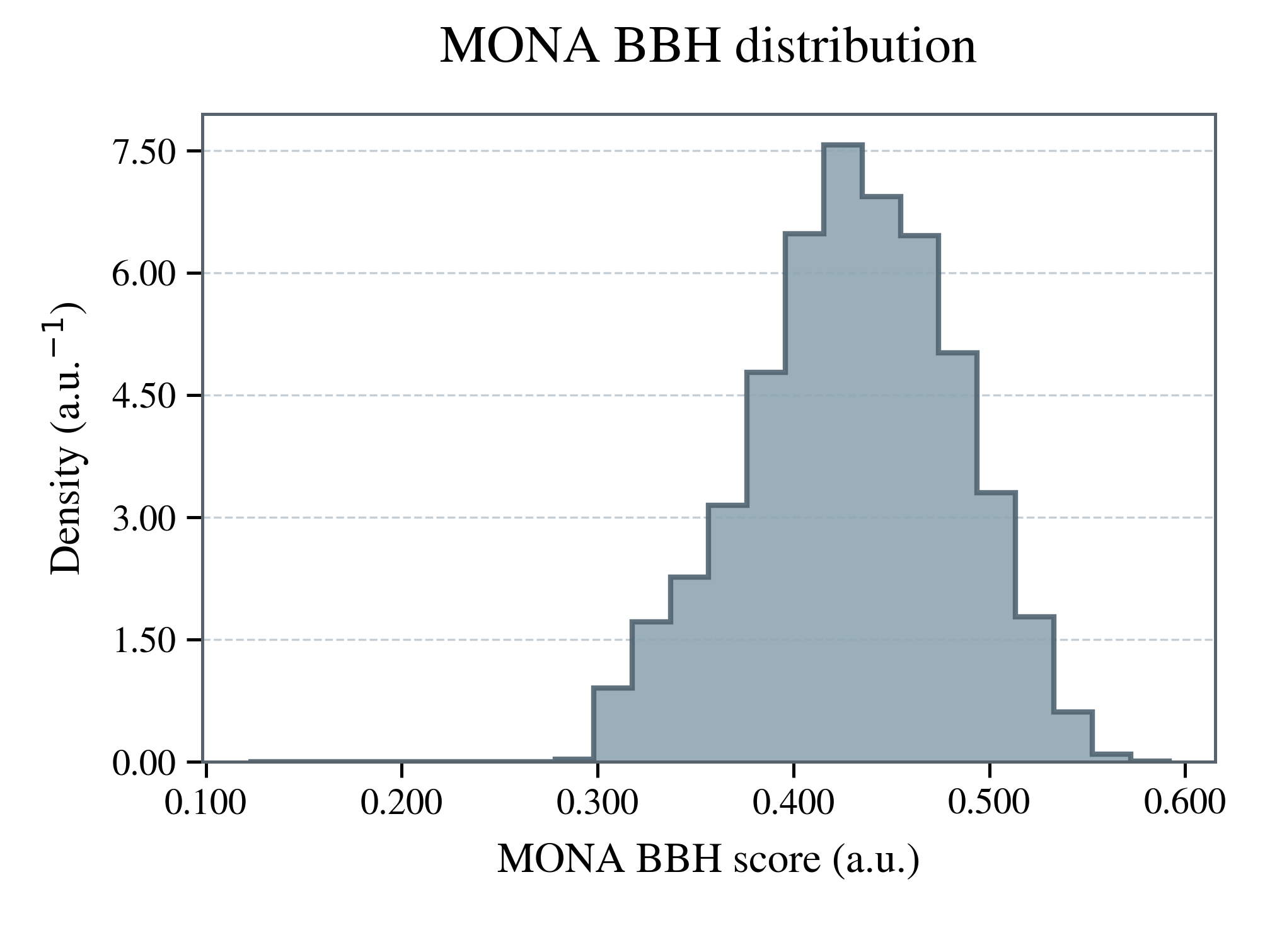} &
\includegraphics[width=0.145\textwidth]{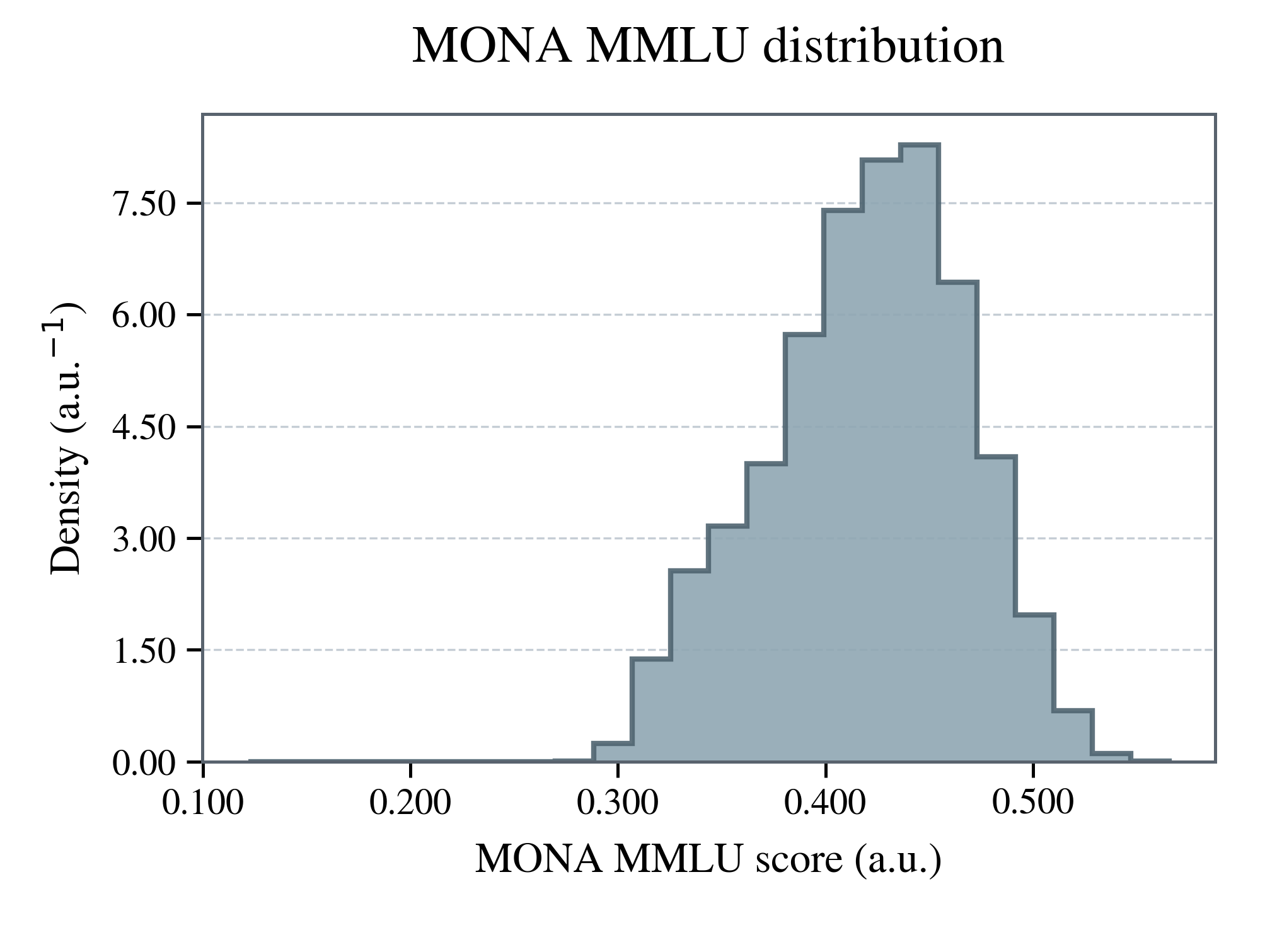} &
\includegraphics[width=0.145\textwidth]{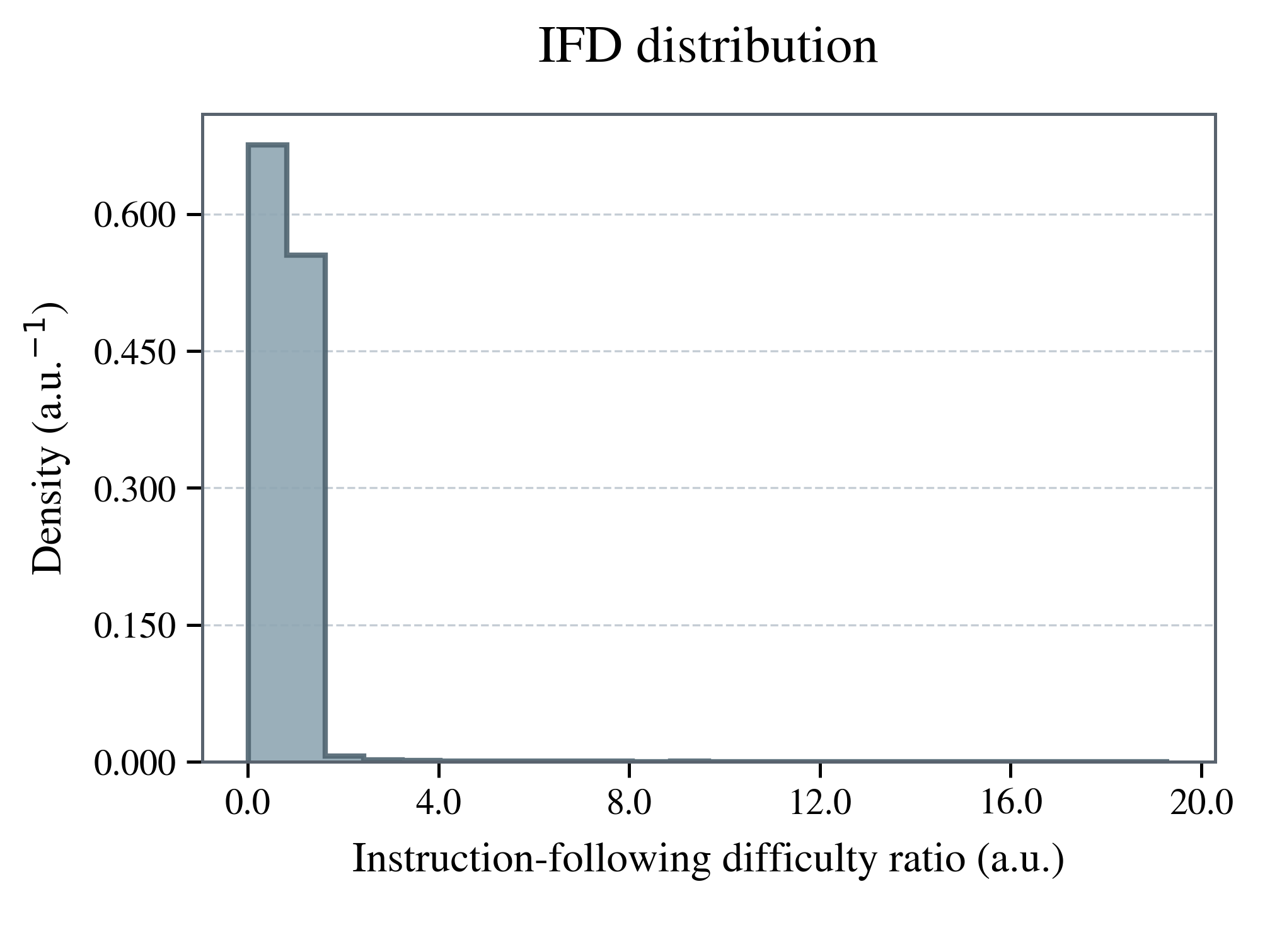} &
\includegraphics[width=0.145\textwidth]{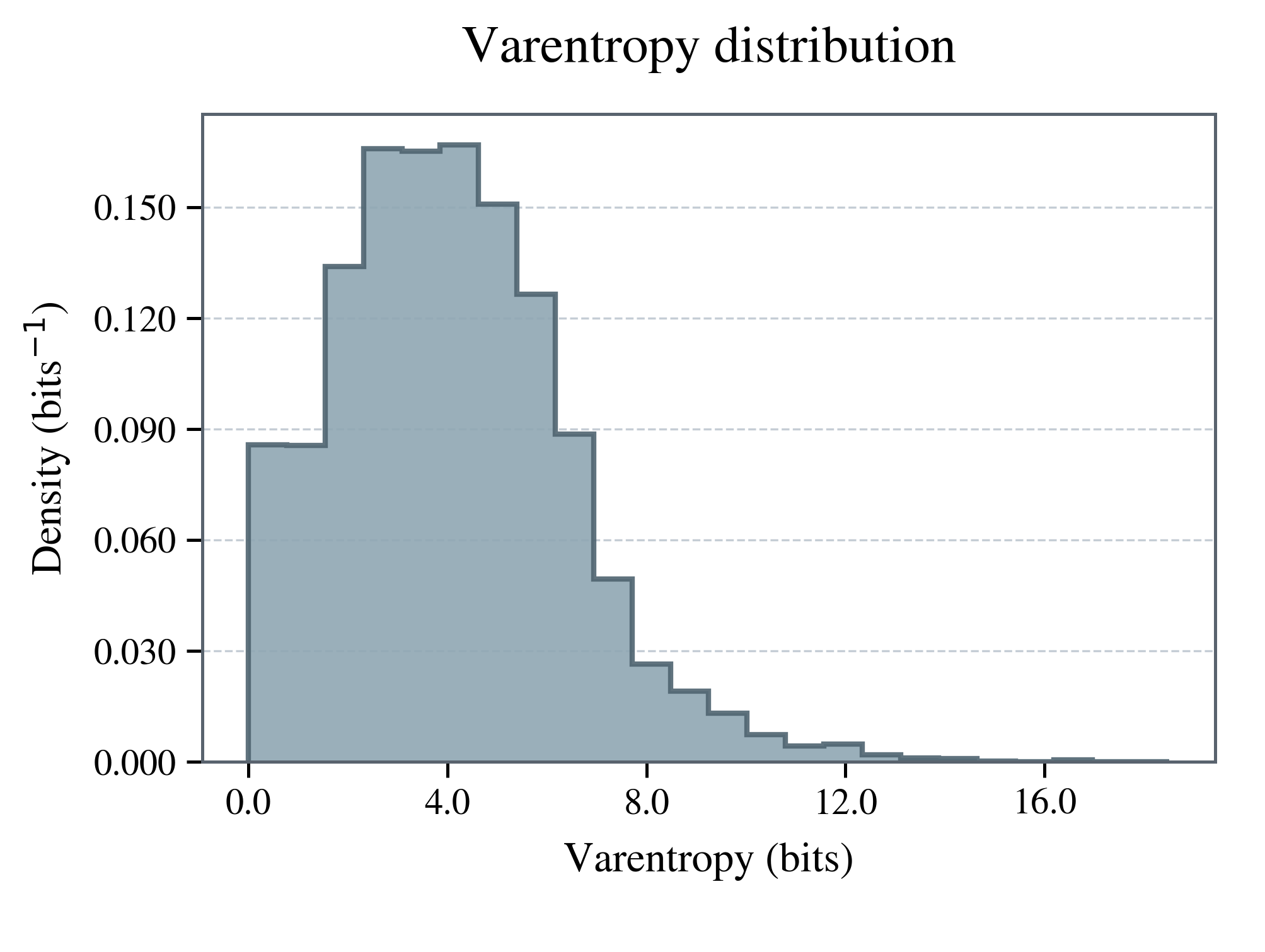} \\
Recipe 4 &
\includegraphics[width=0.145\textwidth]{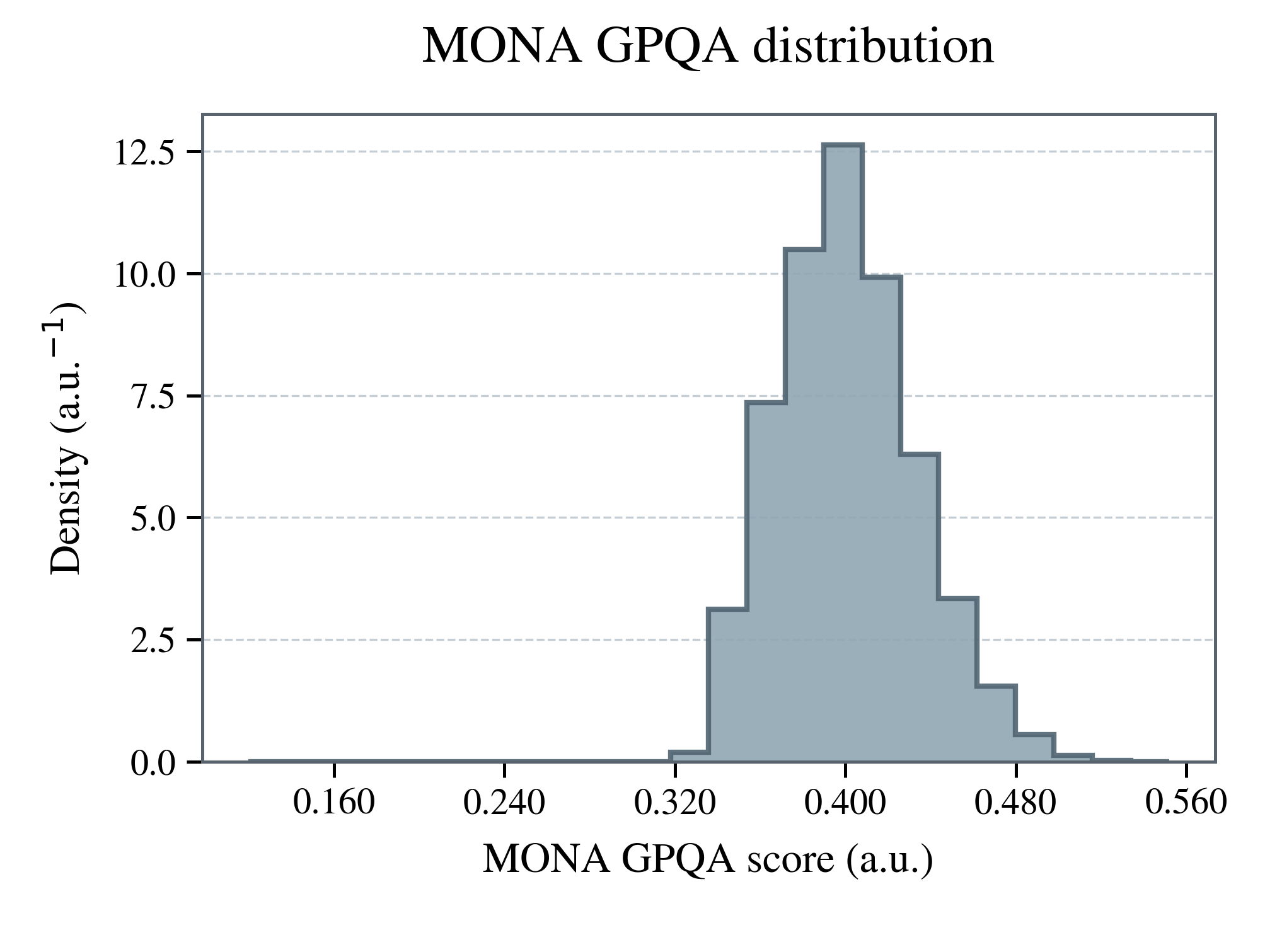} &
\includegraphics[width=0.145\textwidth]{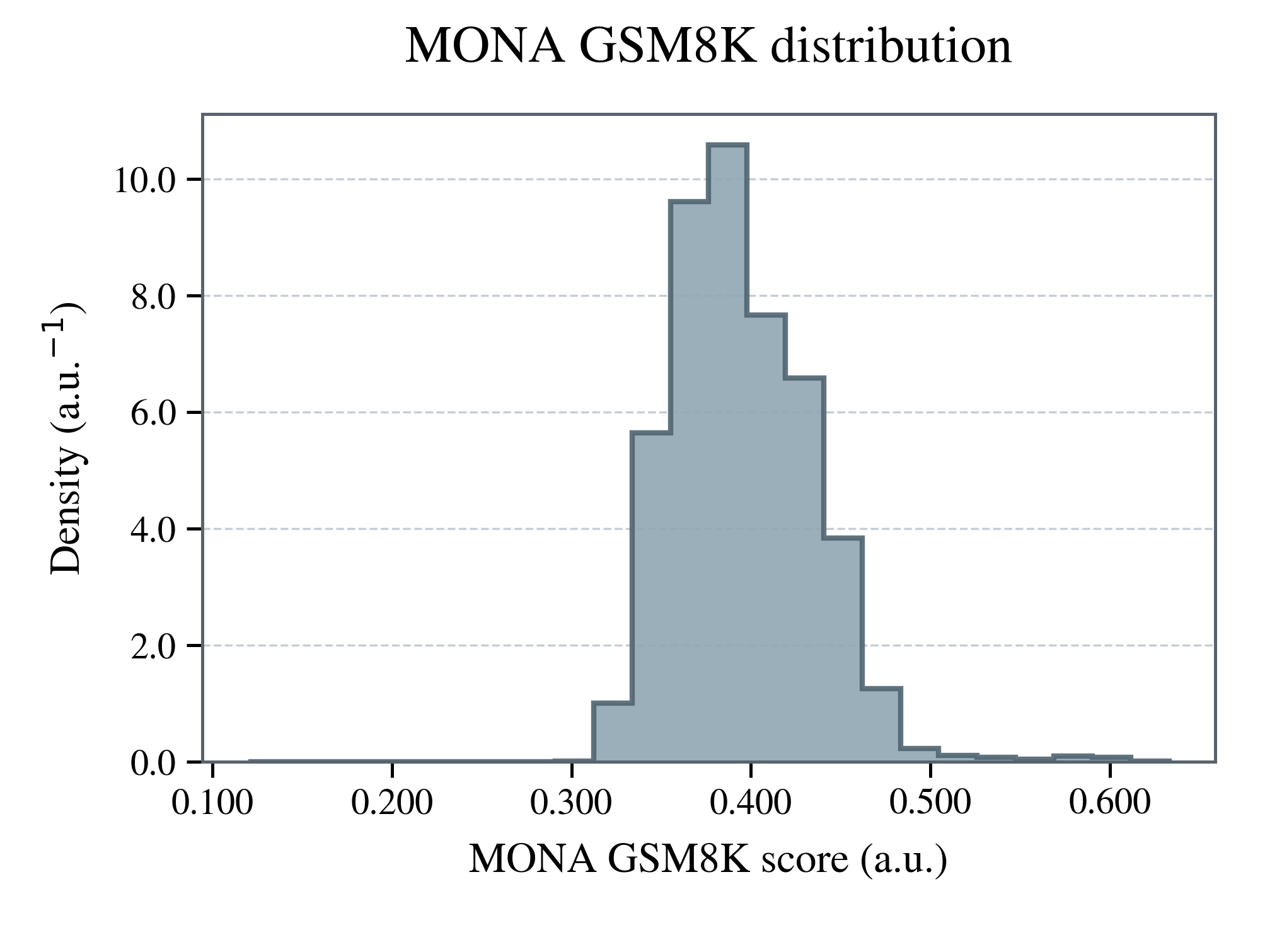} &
\includegraphics[width=0.145\textwidth]{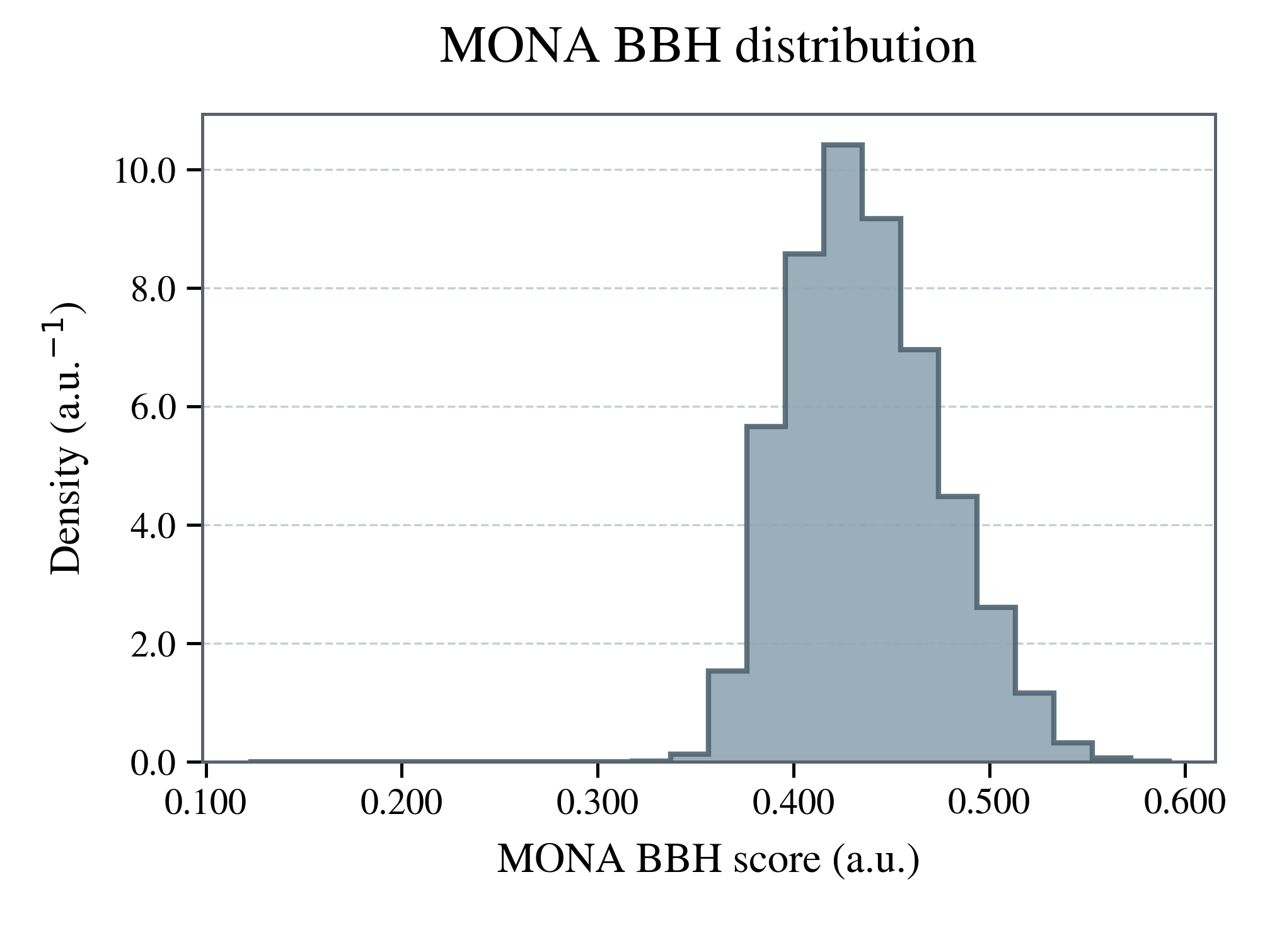} &
\includegraphics[width=0.145\textwidth]{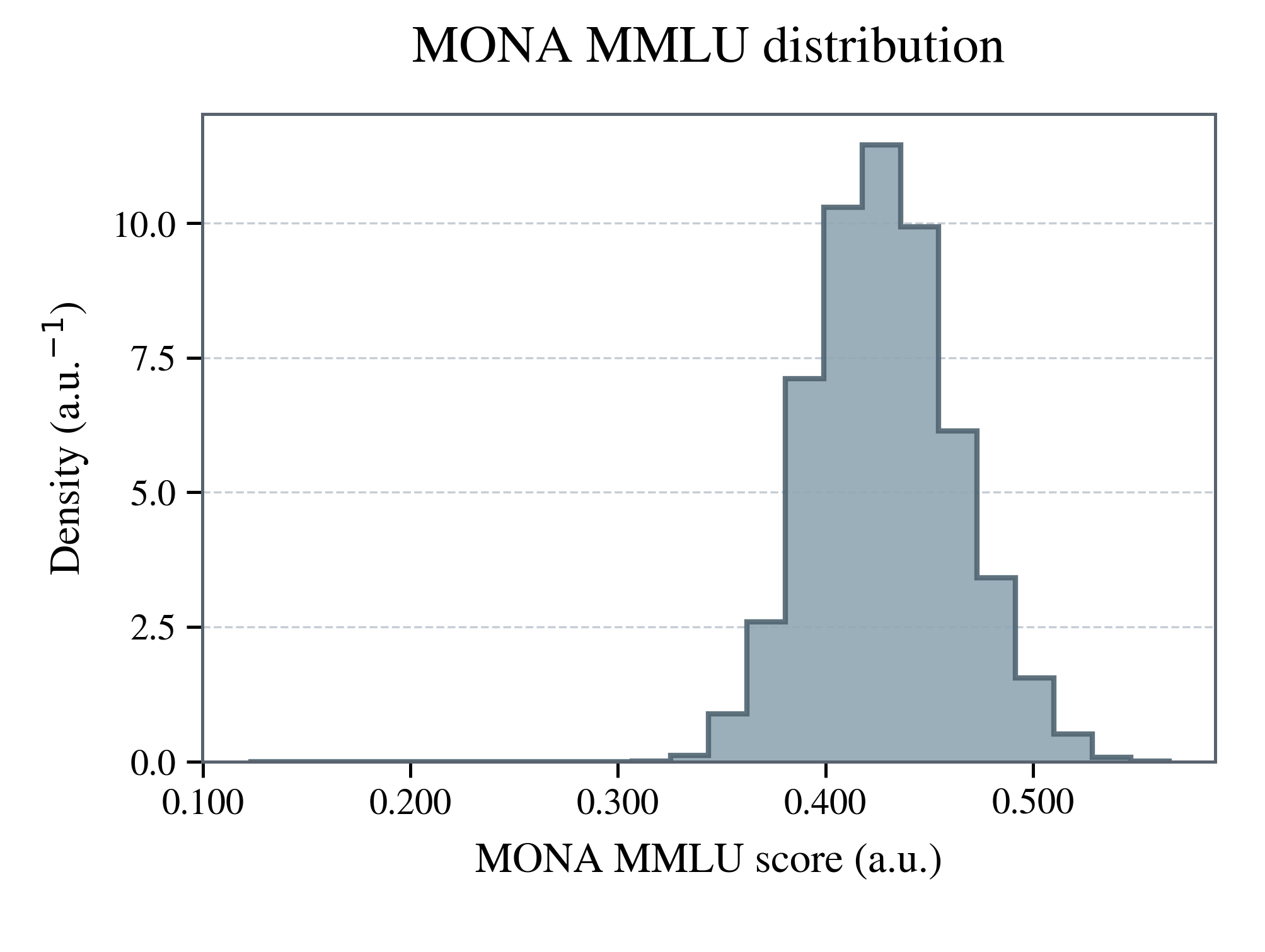} &
\includegraphics[width=0.145\textwidth]{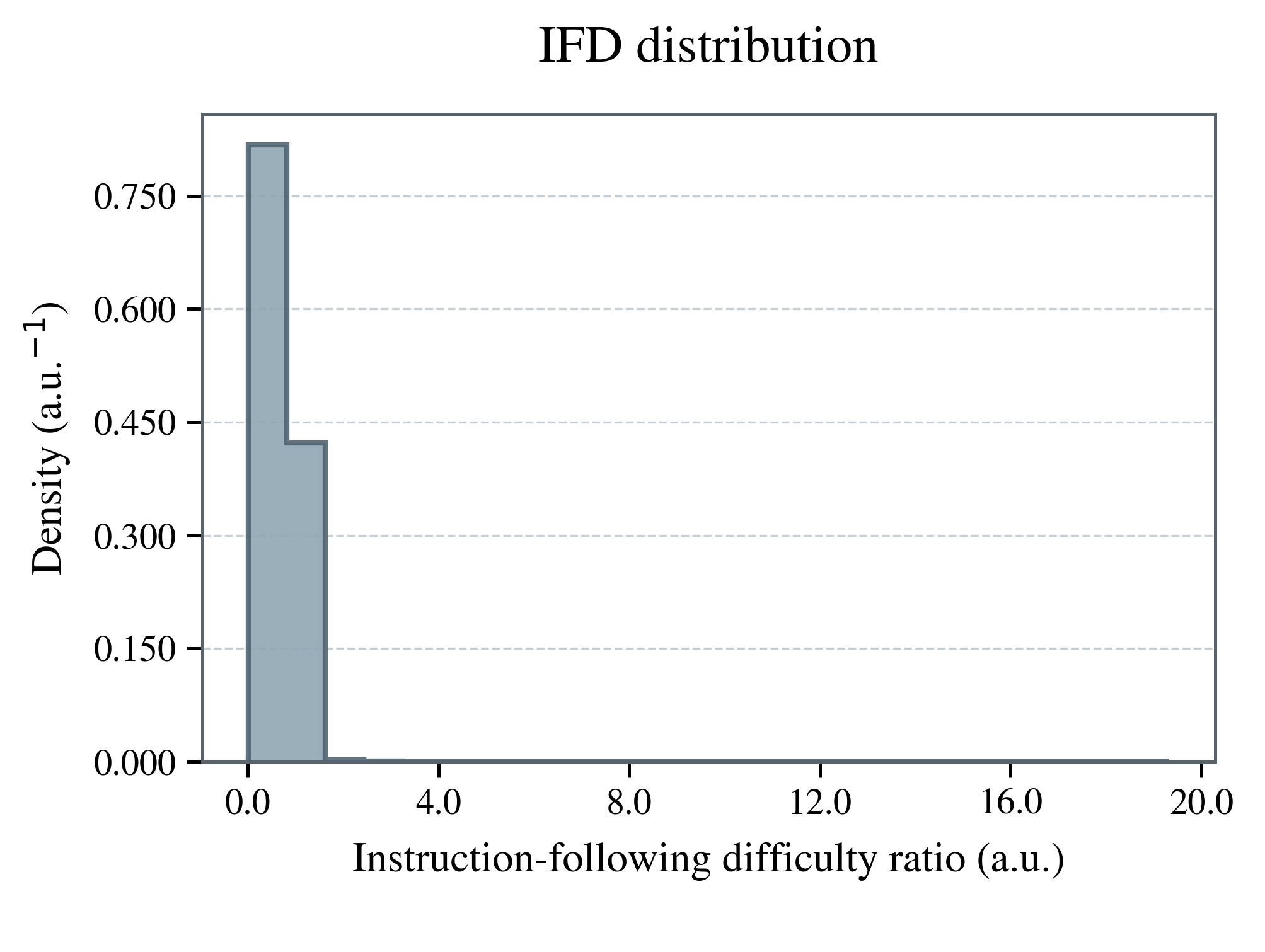} &
\includegraphics[width=0.145\textwidth]{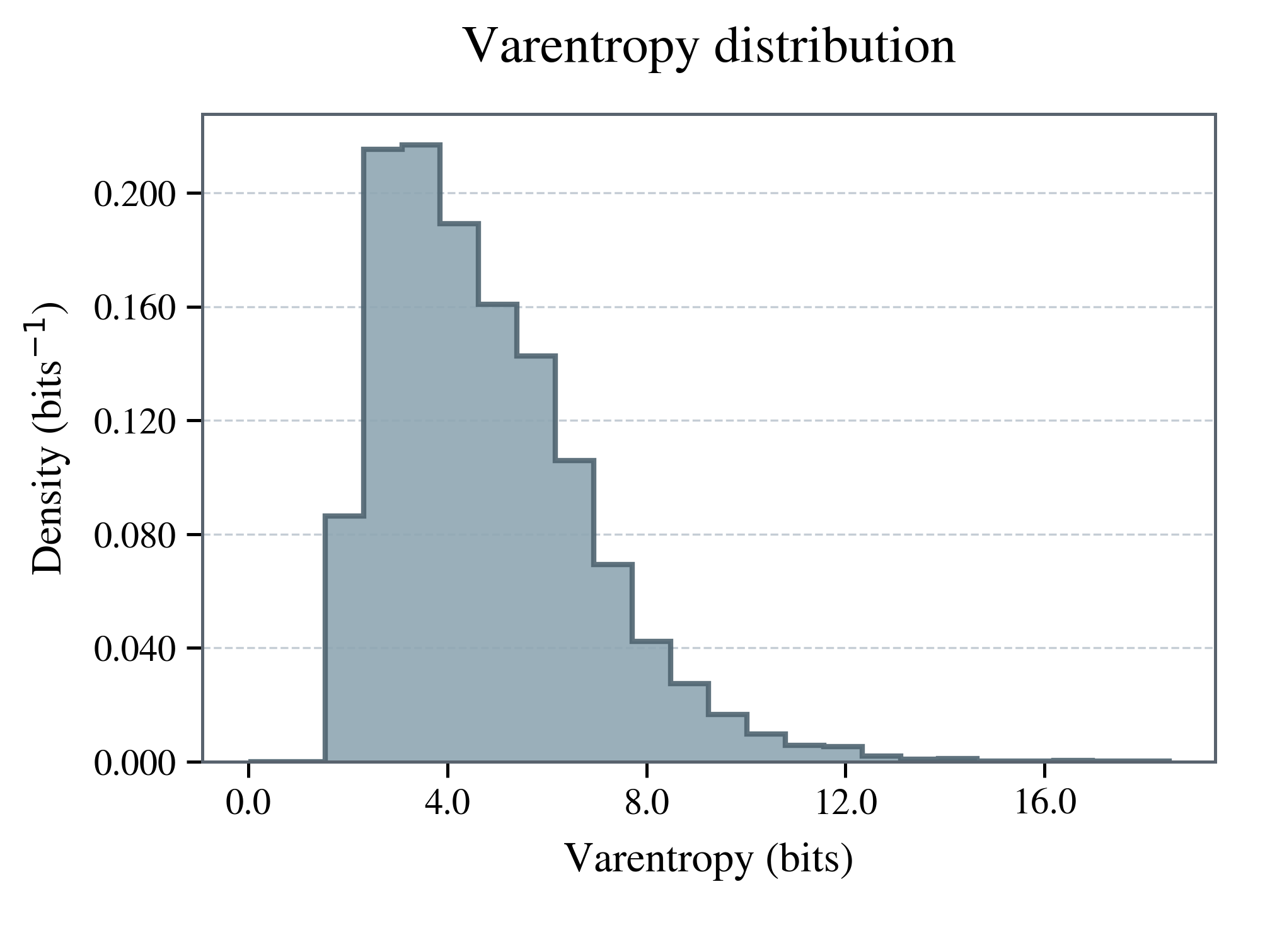} \\
Recipe 5 &
\includegraphics[width=0.145\textwidth]{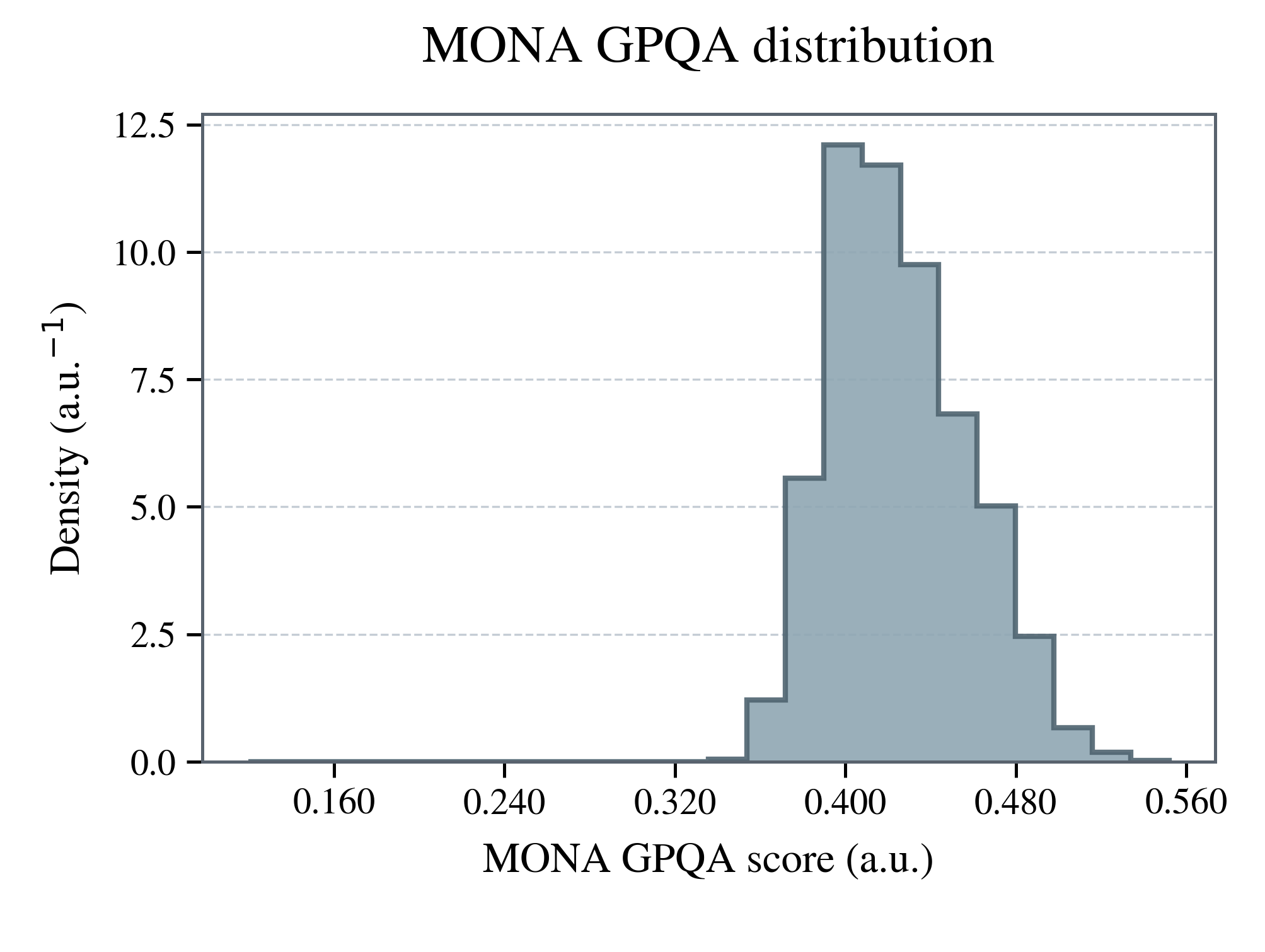} &
\includegraphics[width=0.145\textwidth]{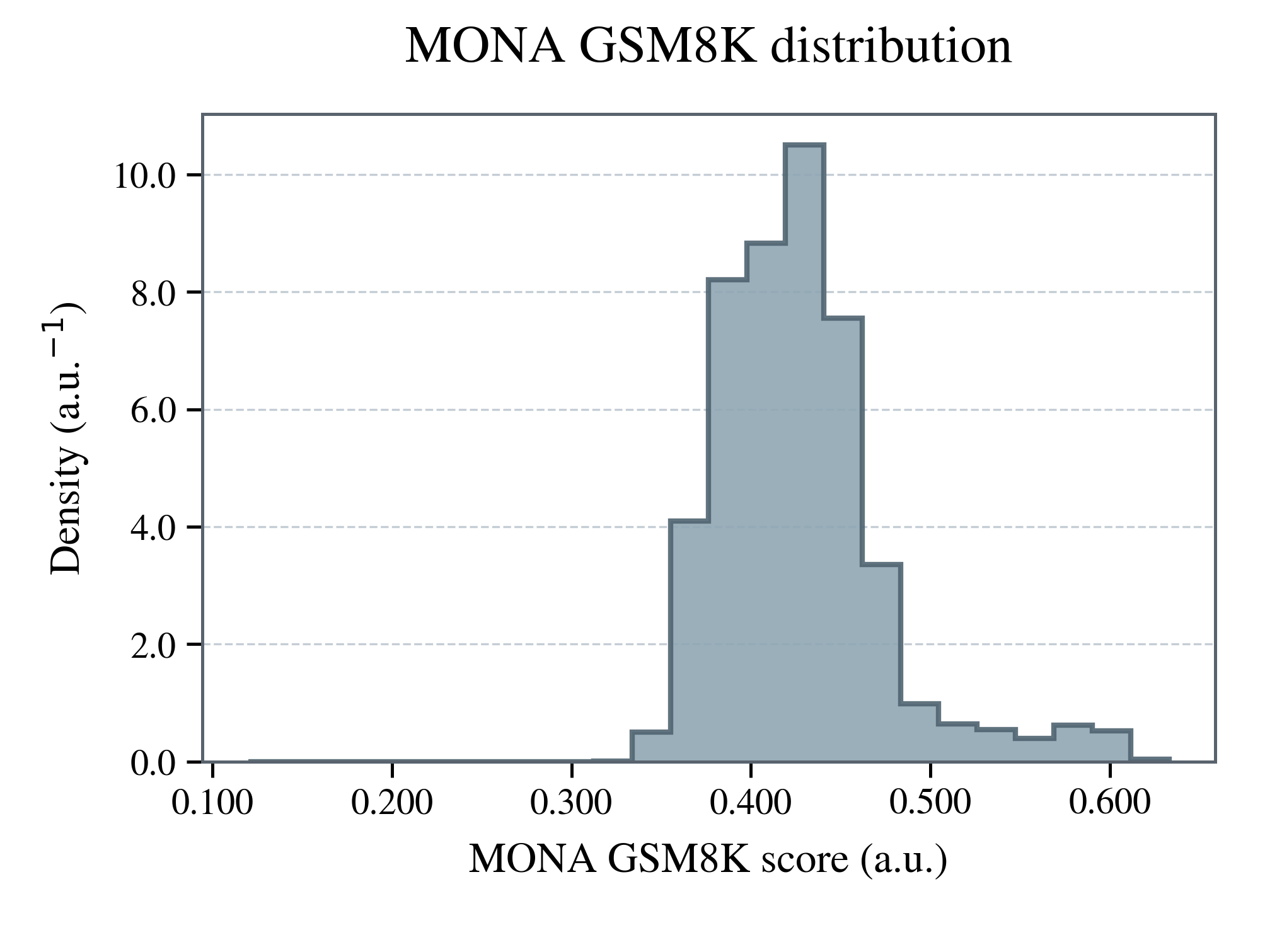} &
\includegraphics[width=0.145\textwidth]{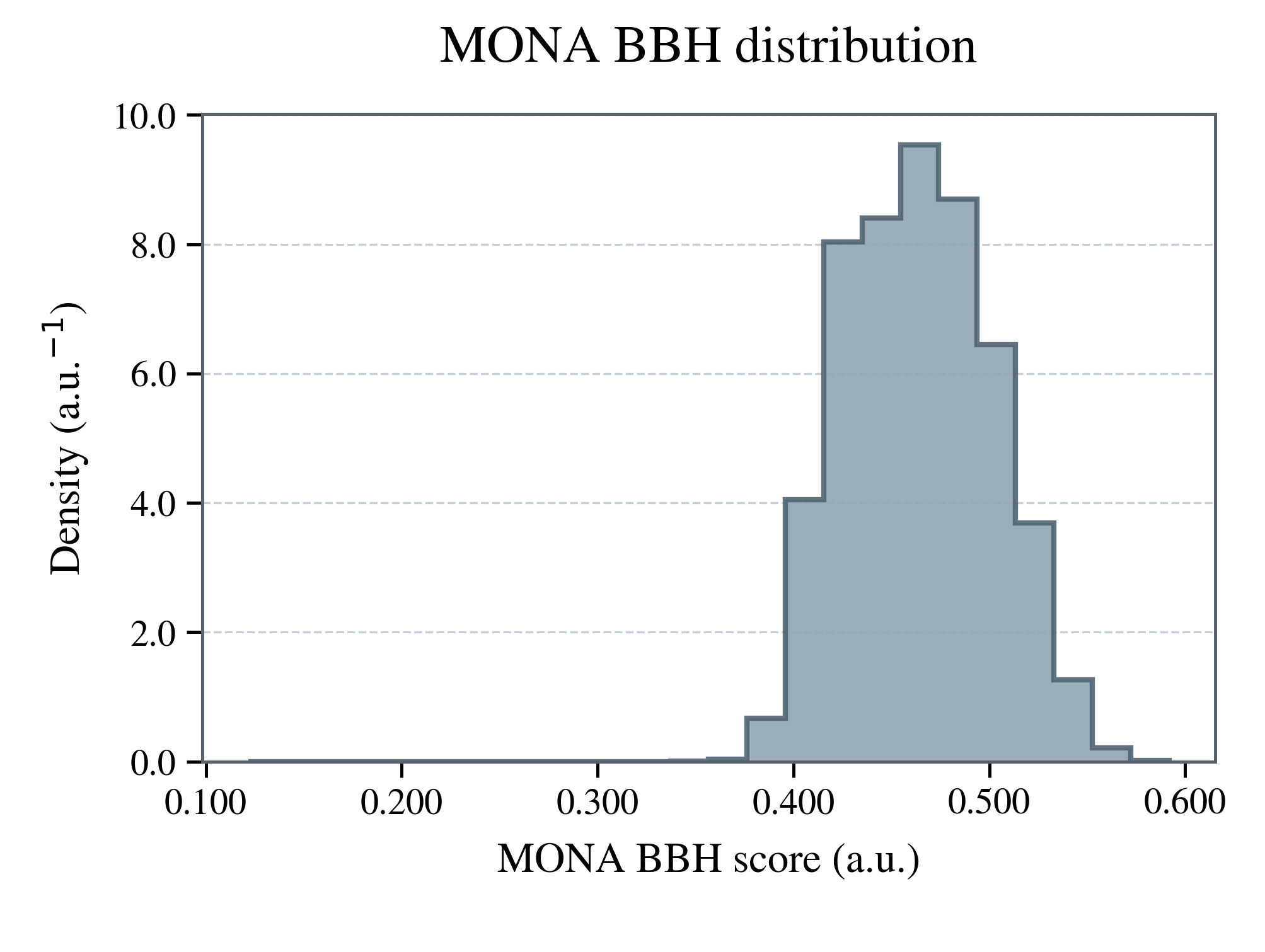} &
\includegraphics[width=0.145\textwidth]{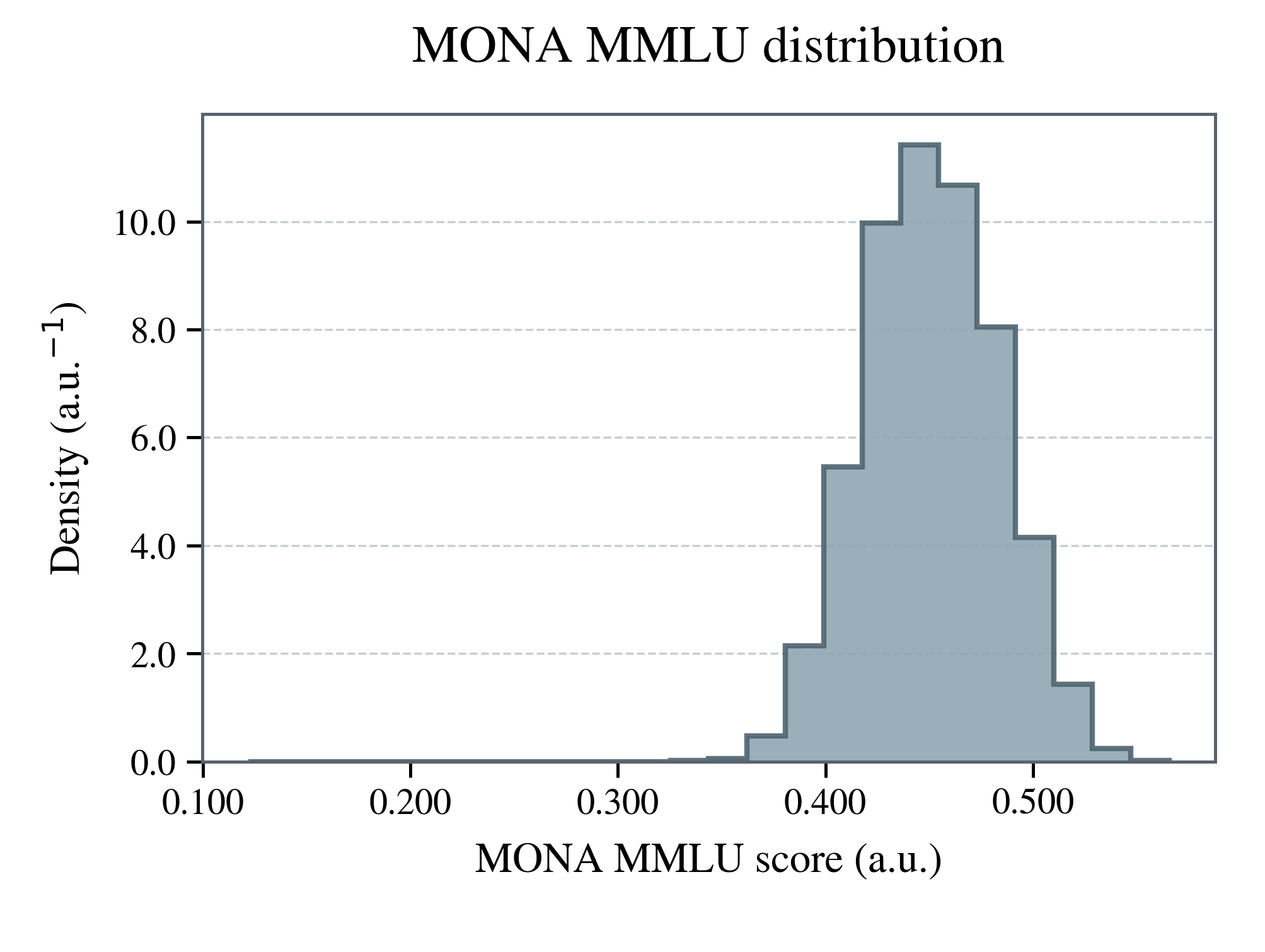} &
\includegraphics[width=0.145\textwidth]{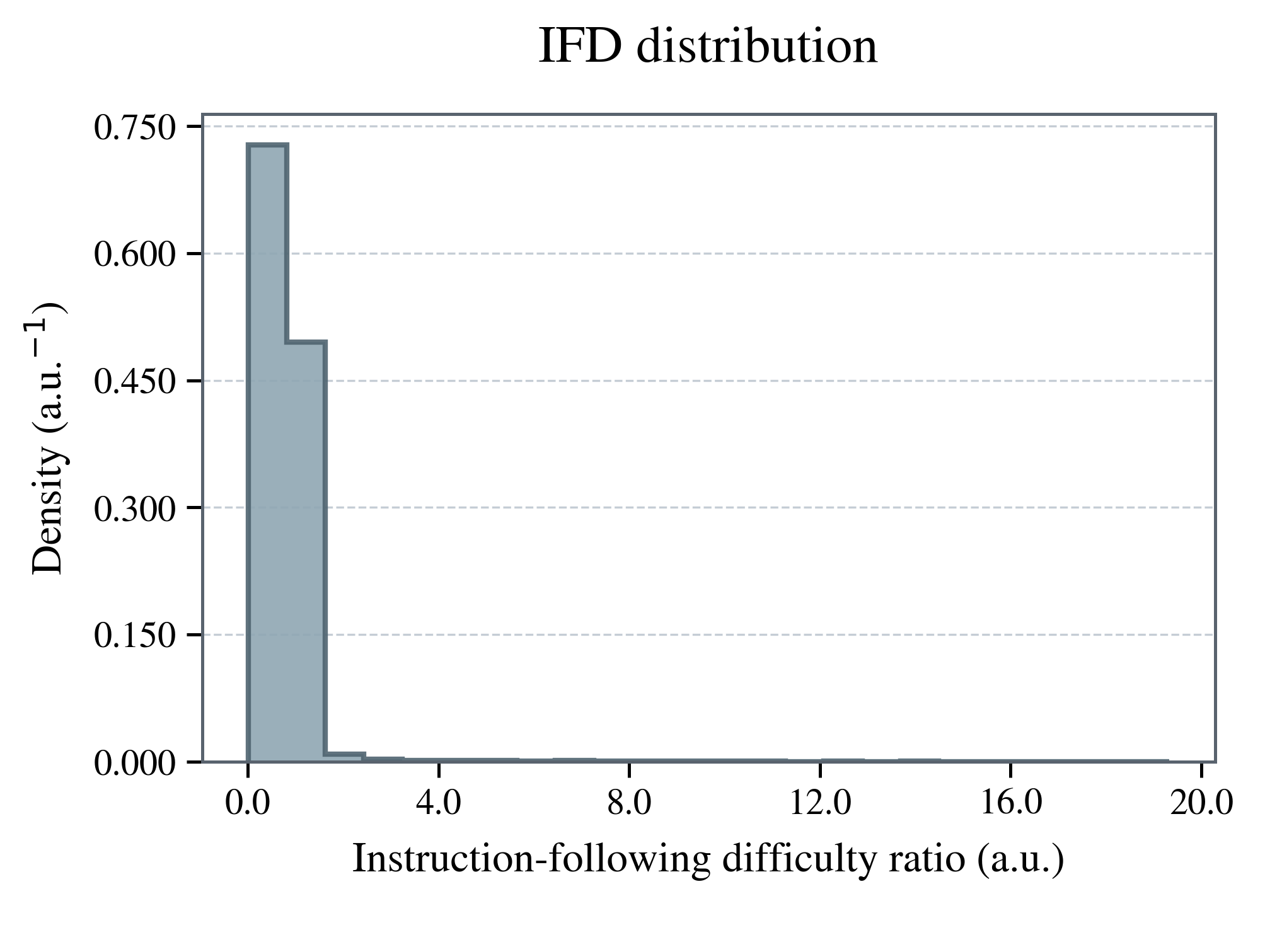} &
\includegraphics[width=0.145\textwidth]{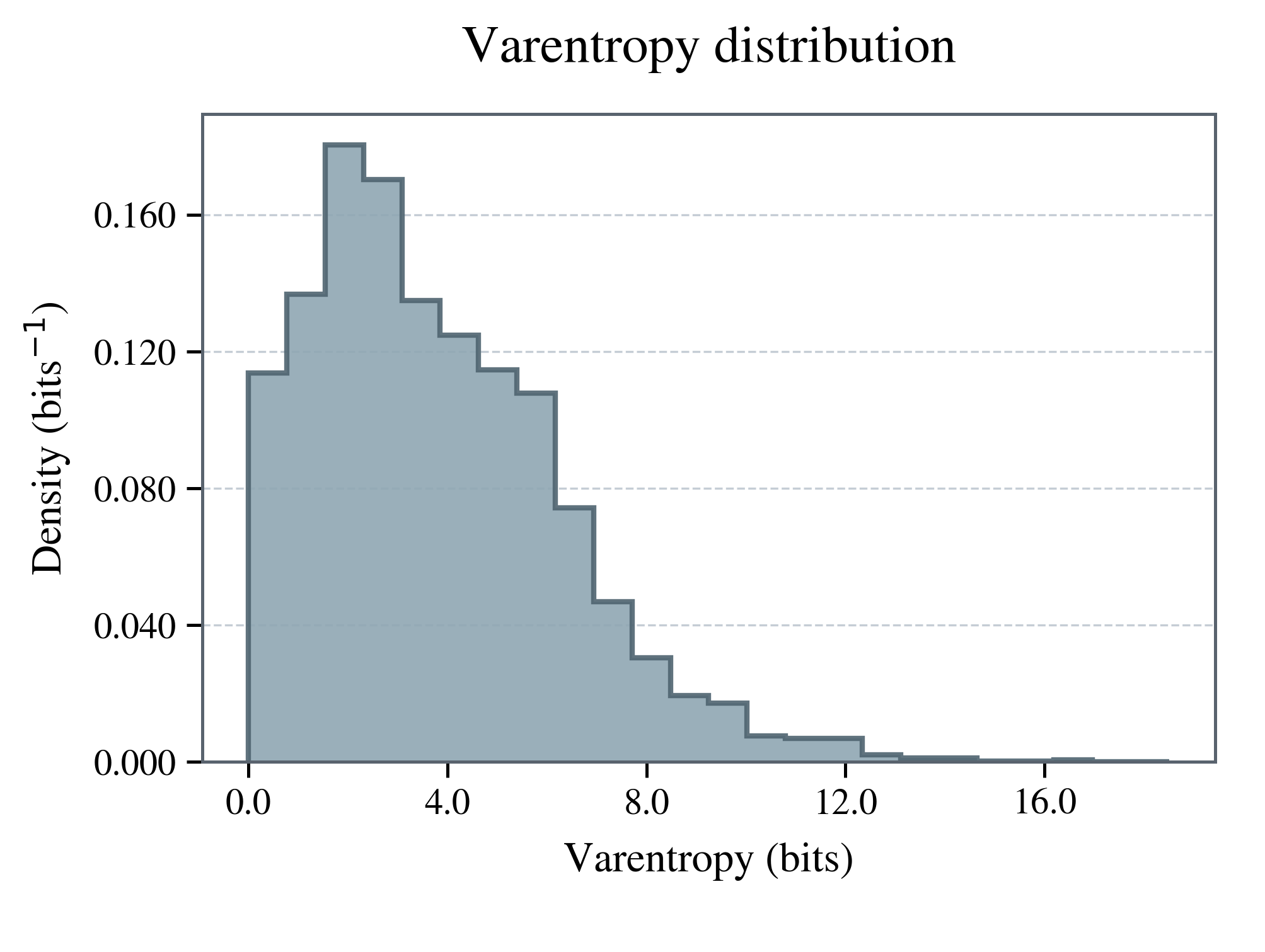} \\
Recipe 6 &
\includegraphics[width=0.145\textwidth]{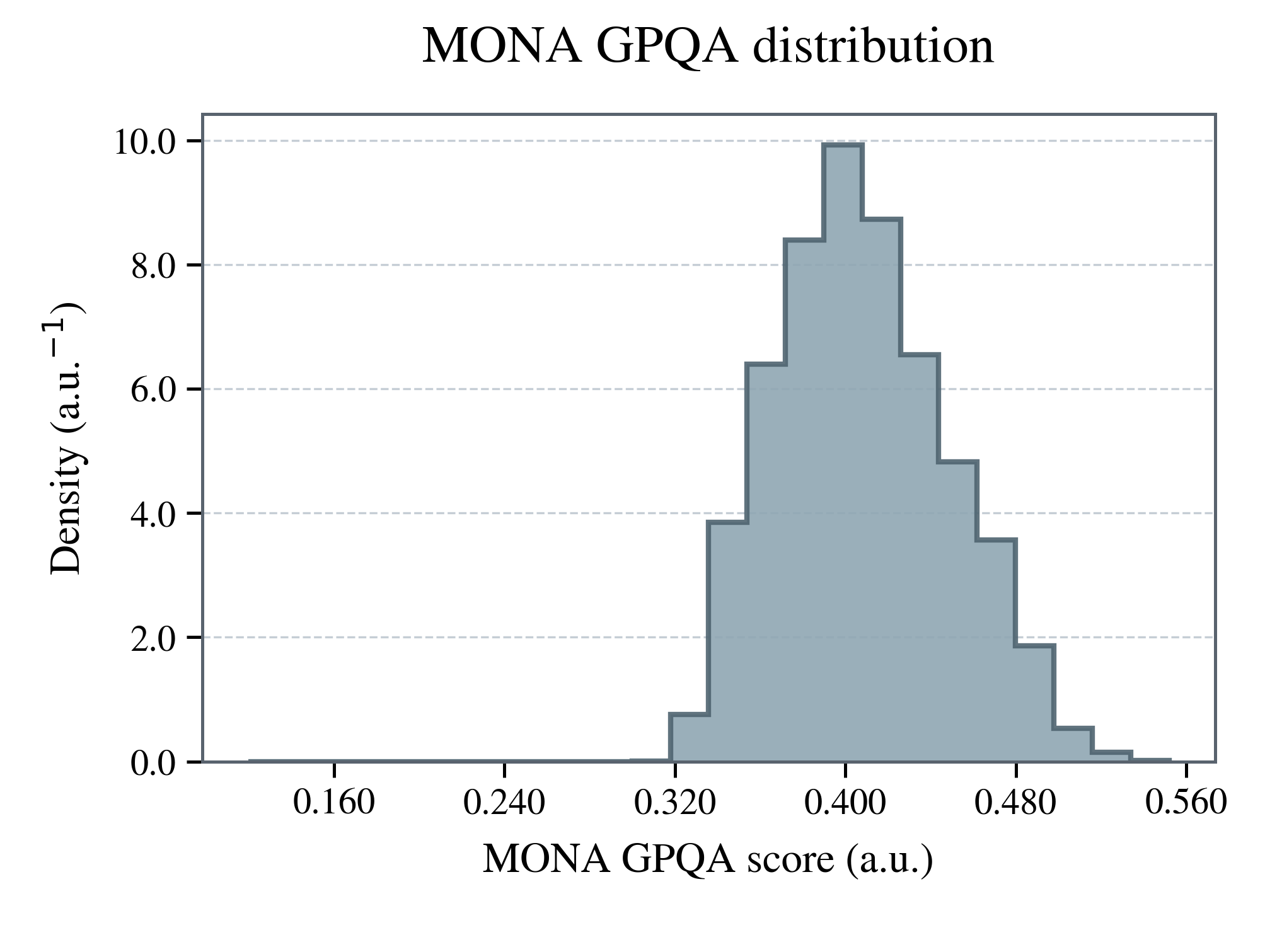} &
\includegraphics[width=0.145\textwidth]{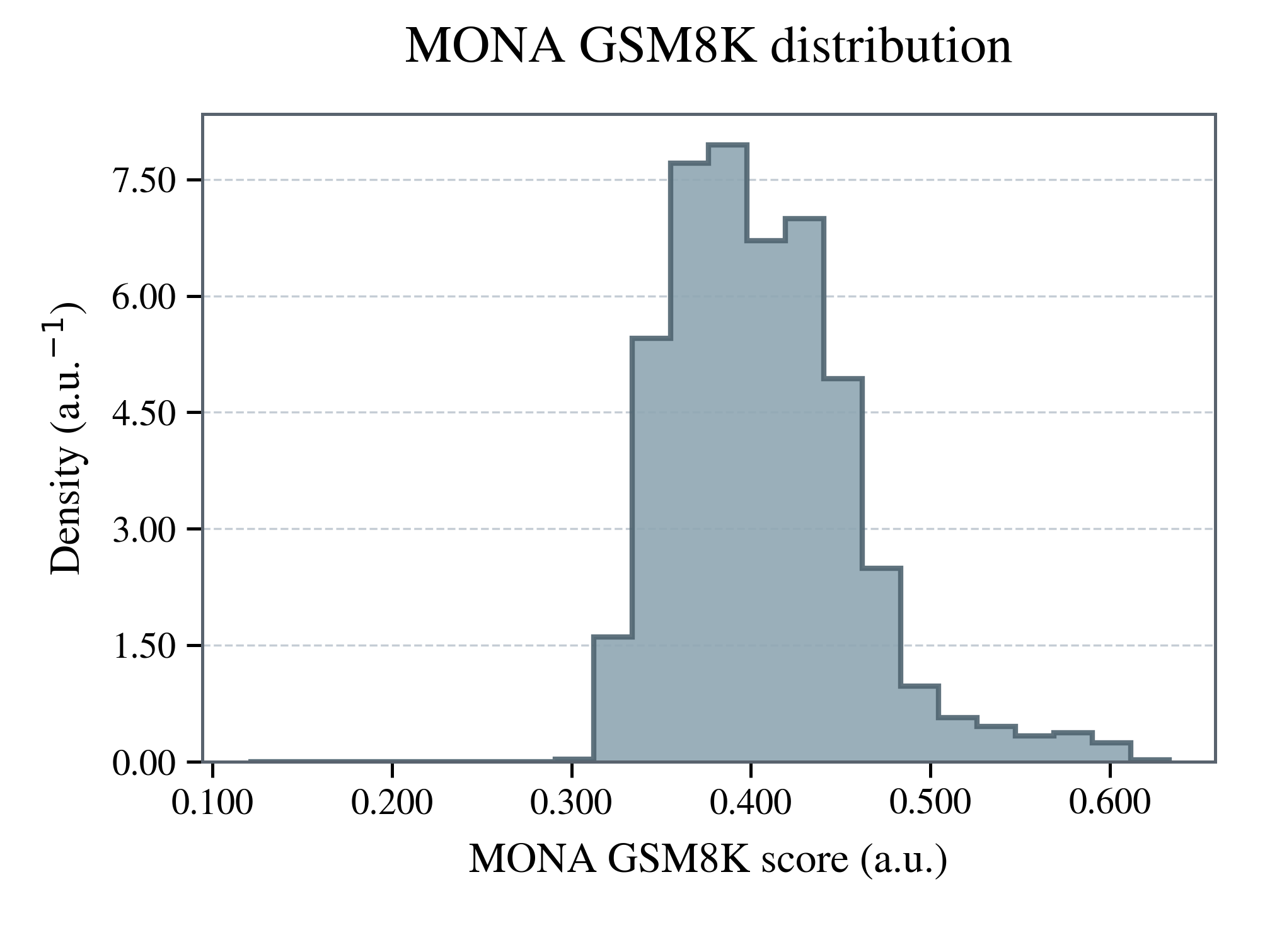} &
\includegraphics[width=0.145\textwidth]{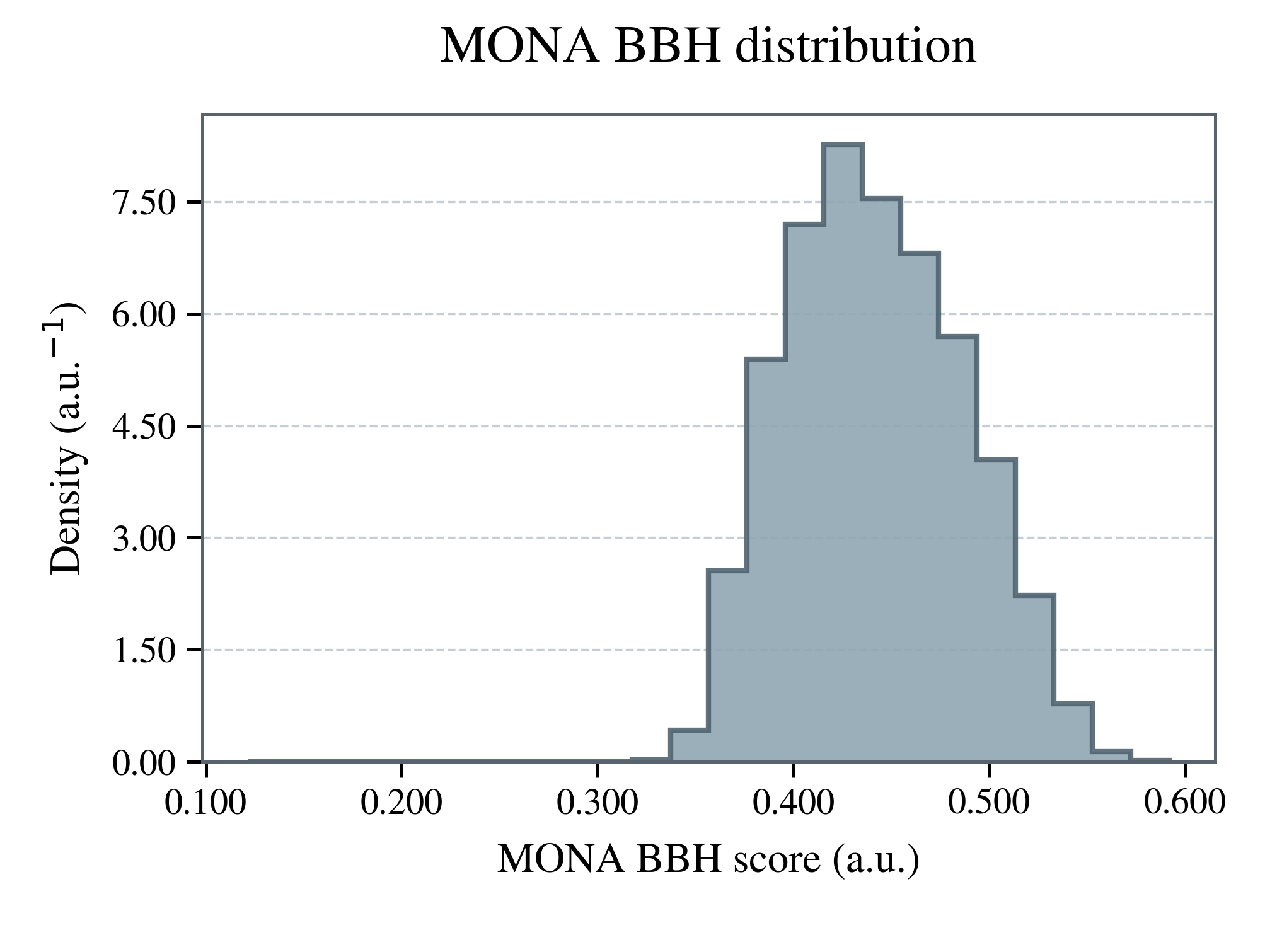} &
\includegraphics[width=0.145\textwidth]{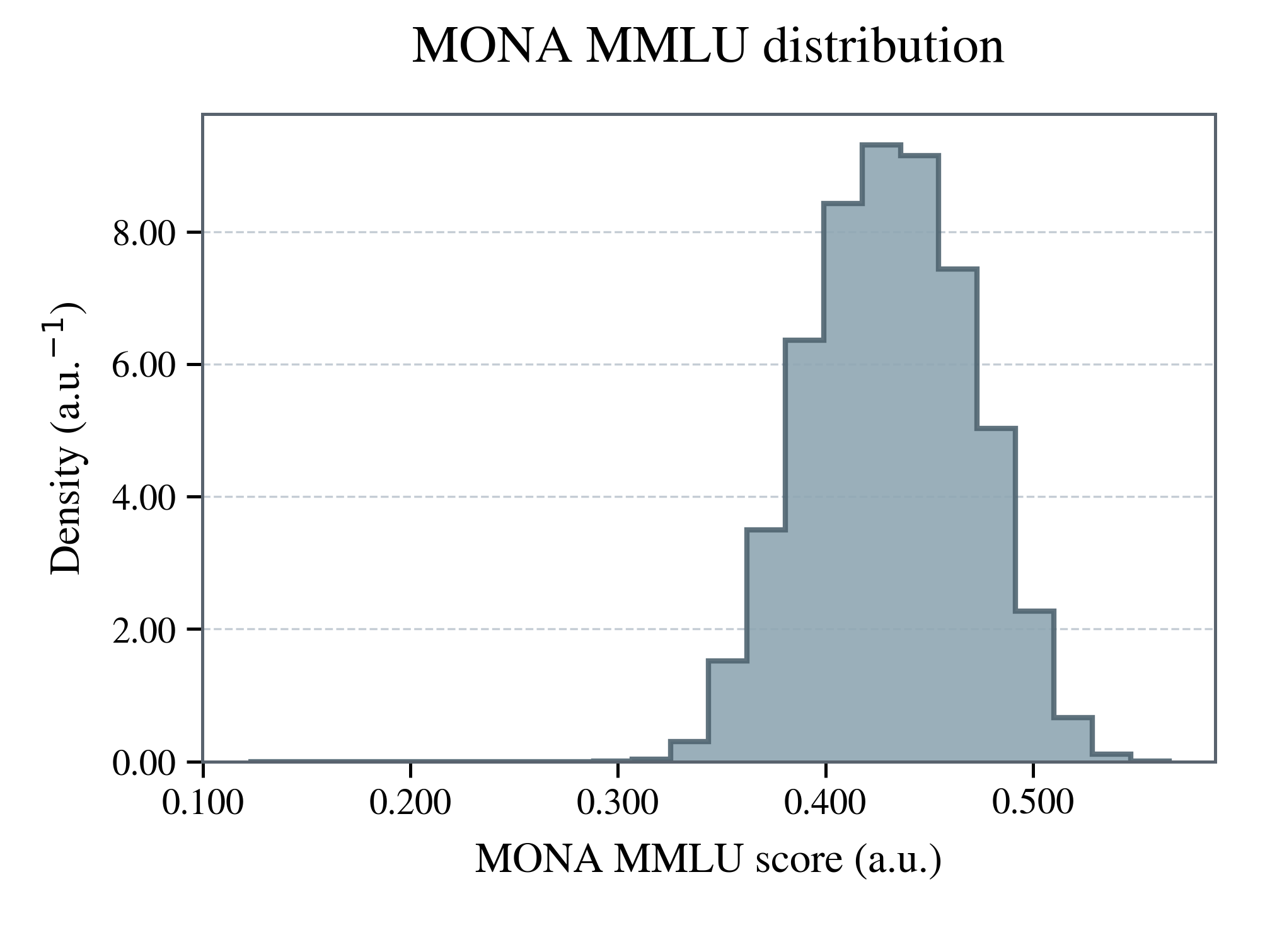} &
\includegraphics[width=0.145\textwidth]{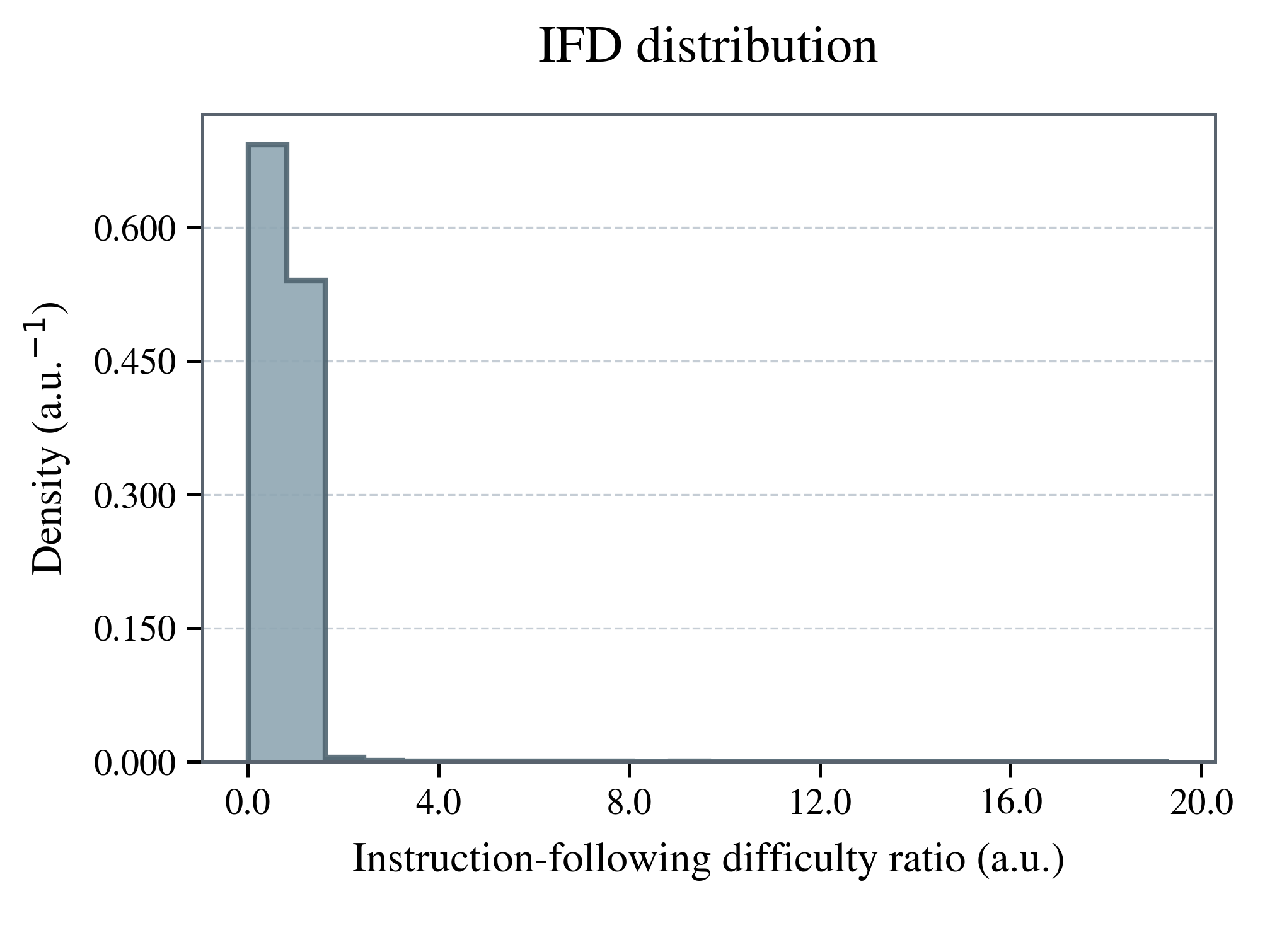} &
\includegraphics[width=0.145\textwidth]{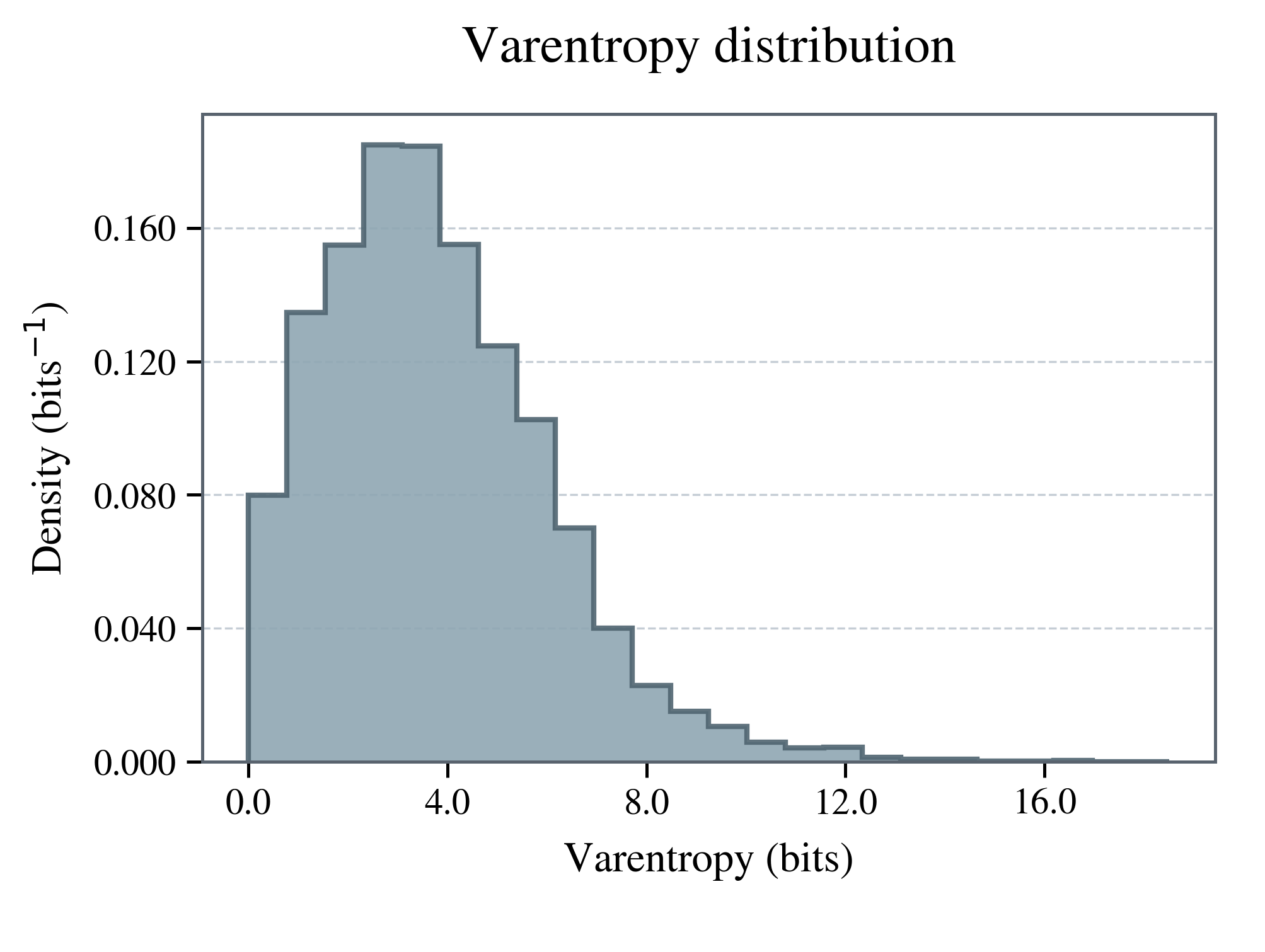} \\
Recipe 7 &
\includegraphics[width=0.145\textwidth]{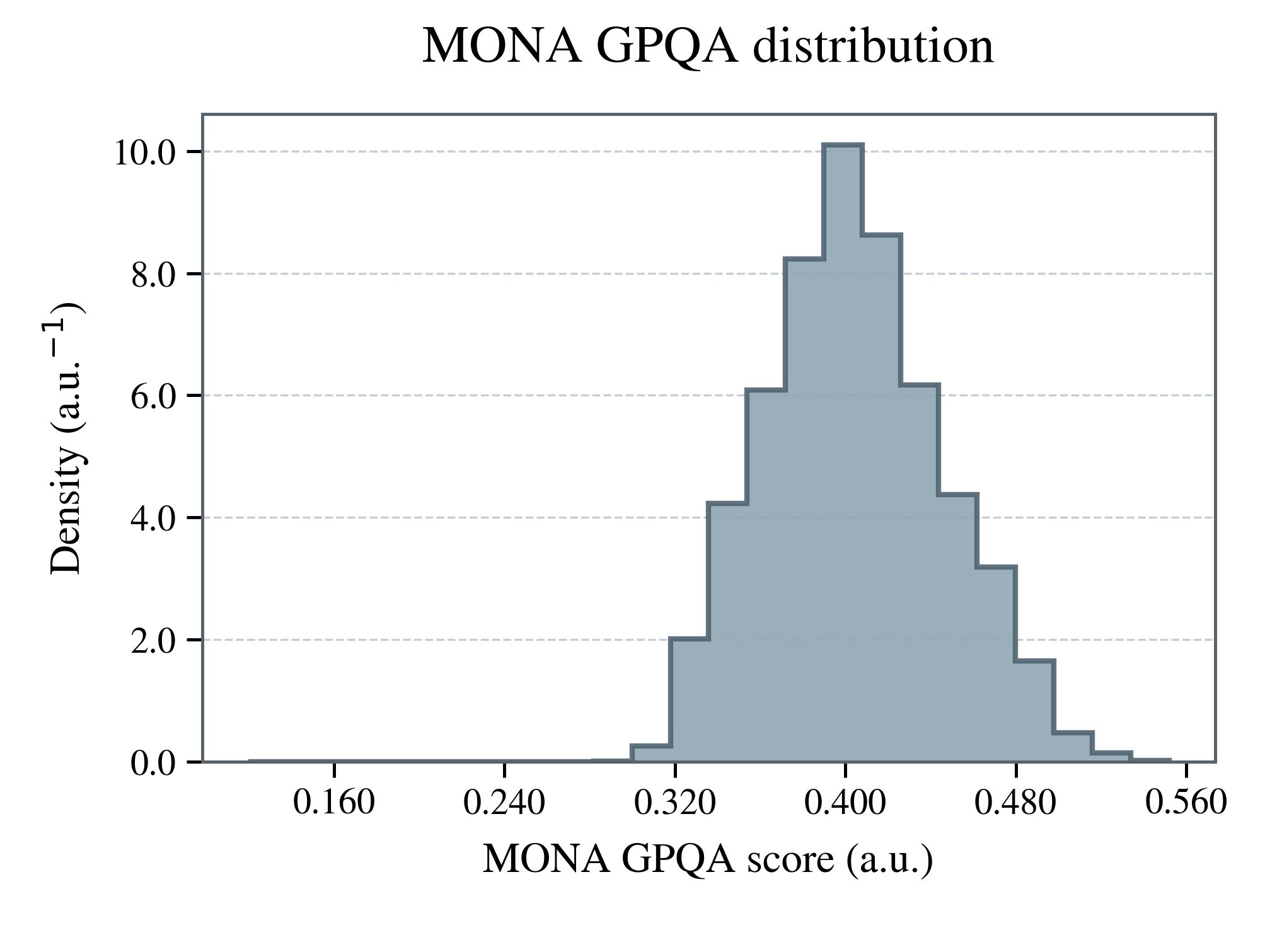} &
\includegraphics[width=0.145\textwidth]{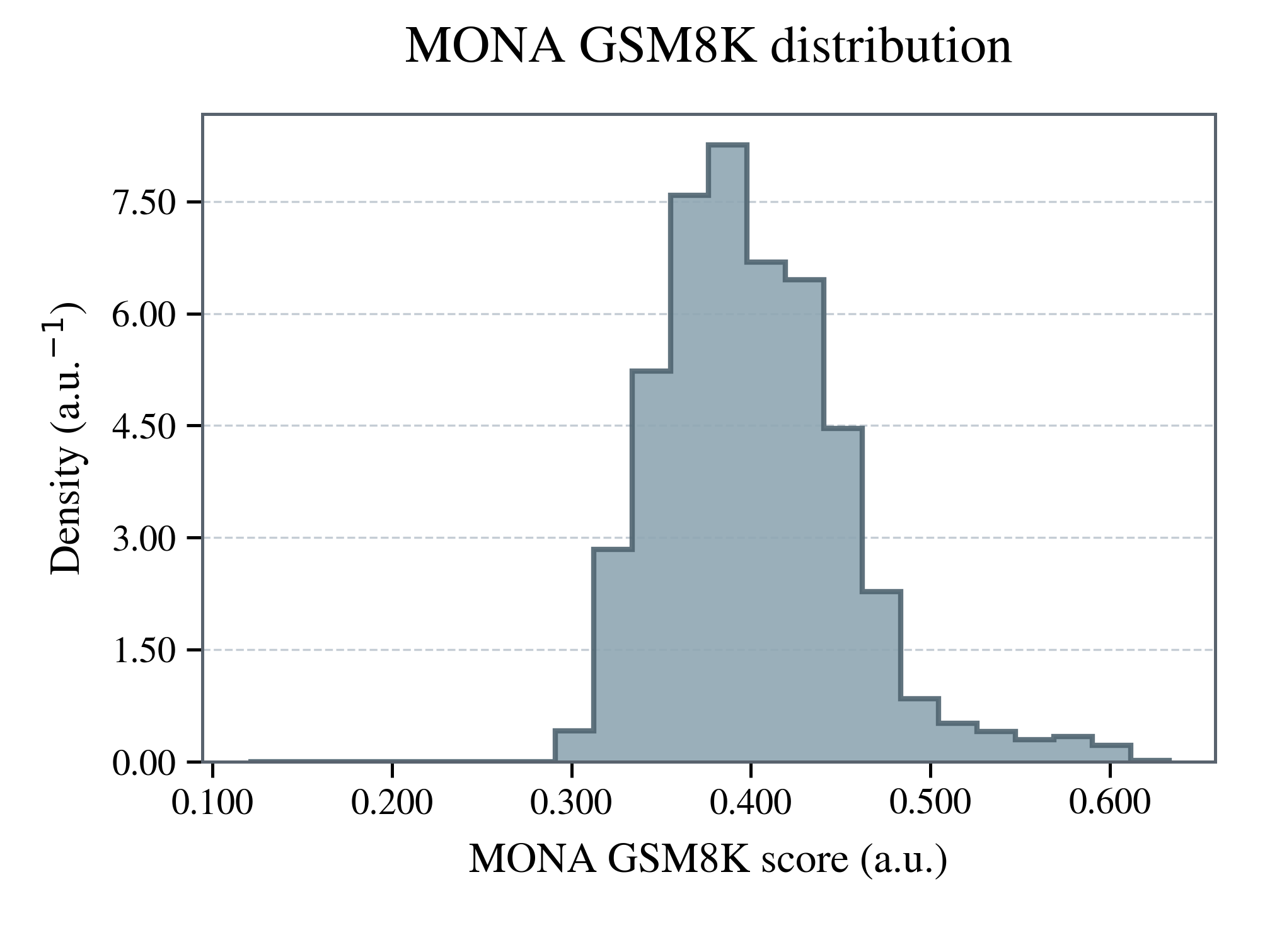} &
\includegraphics[width=0.145\textwidth]{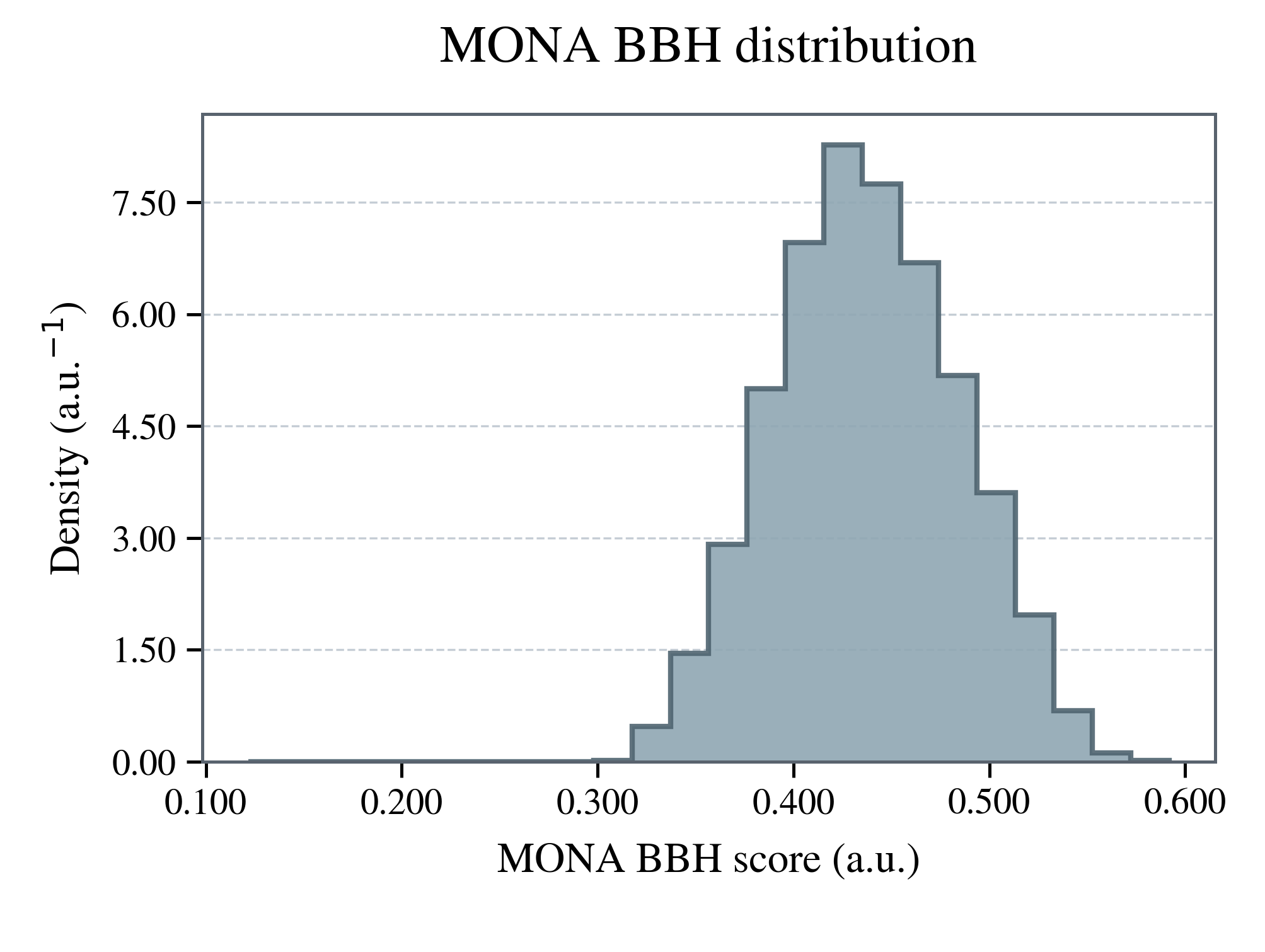} &
\includegraphics[width=0.145\textwidth]{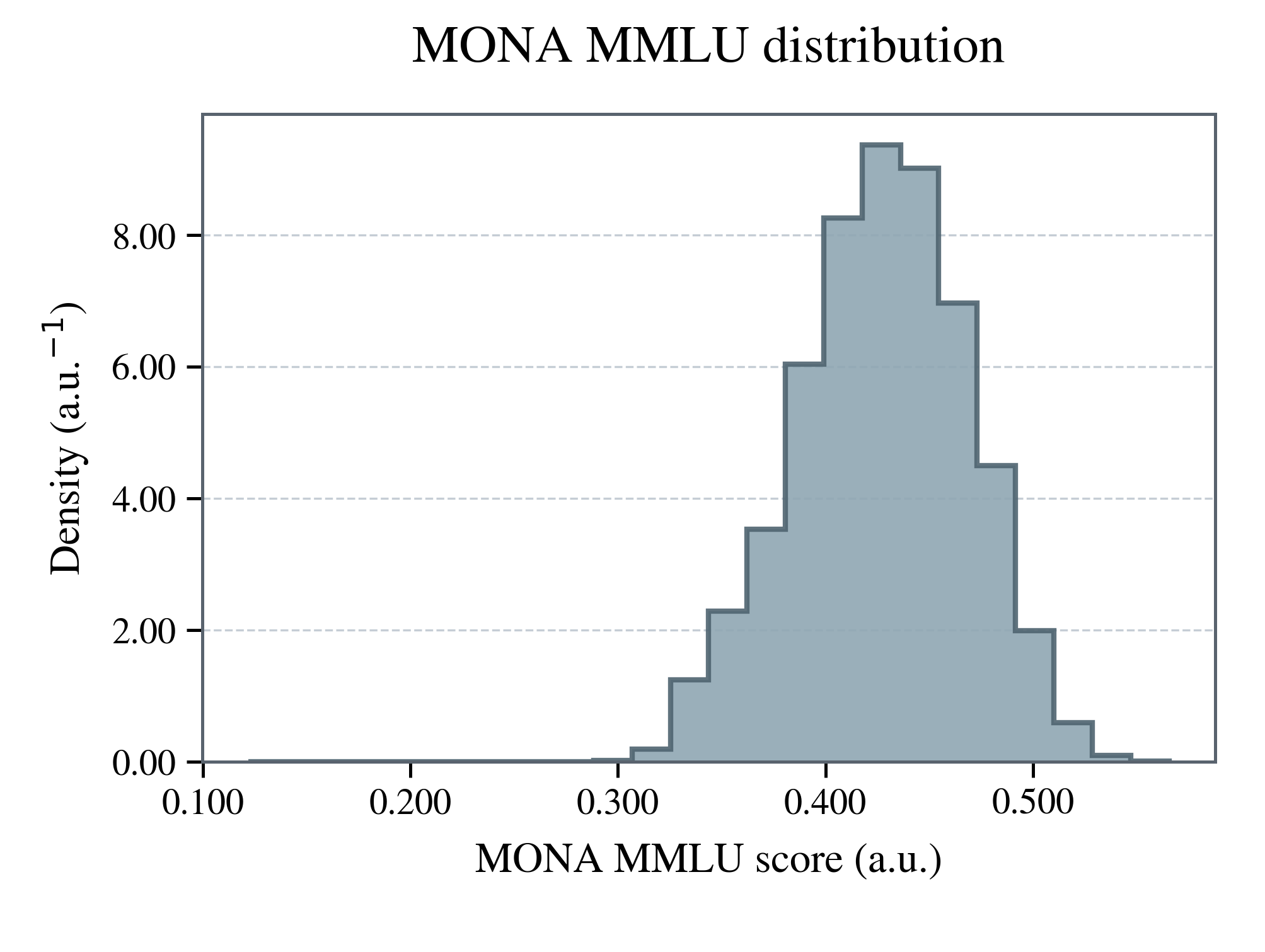} &
\includegraphics[width=0.145\textwidth]{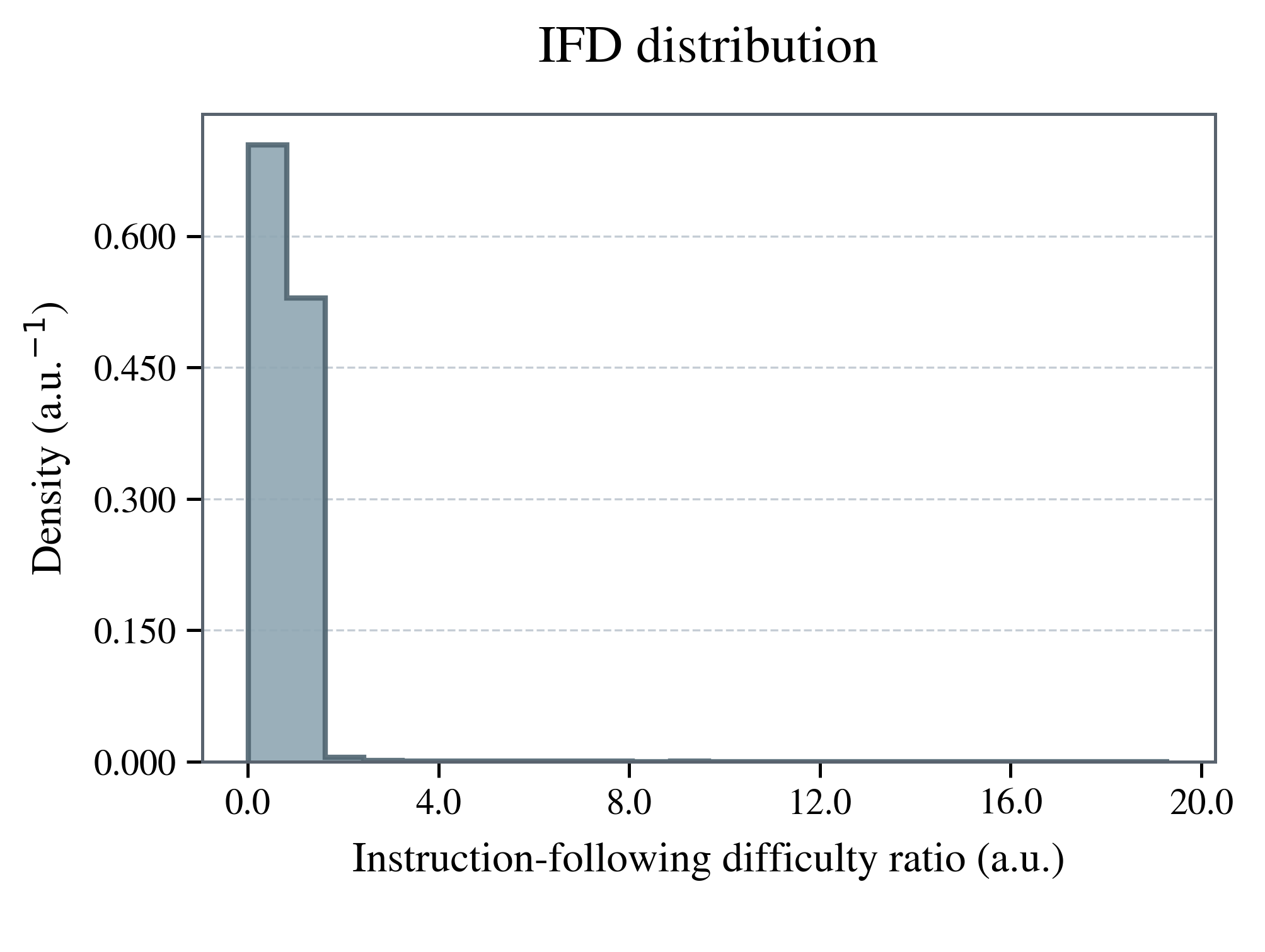} &
\includegraphics[width=0.145\textwidth]{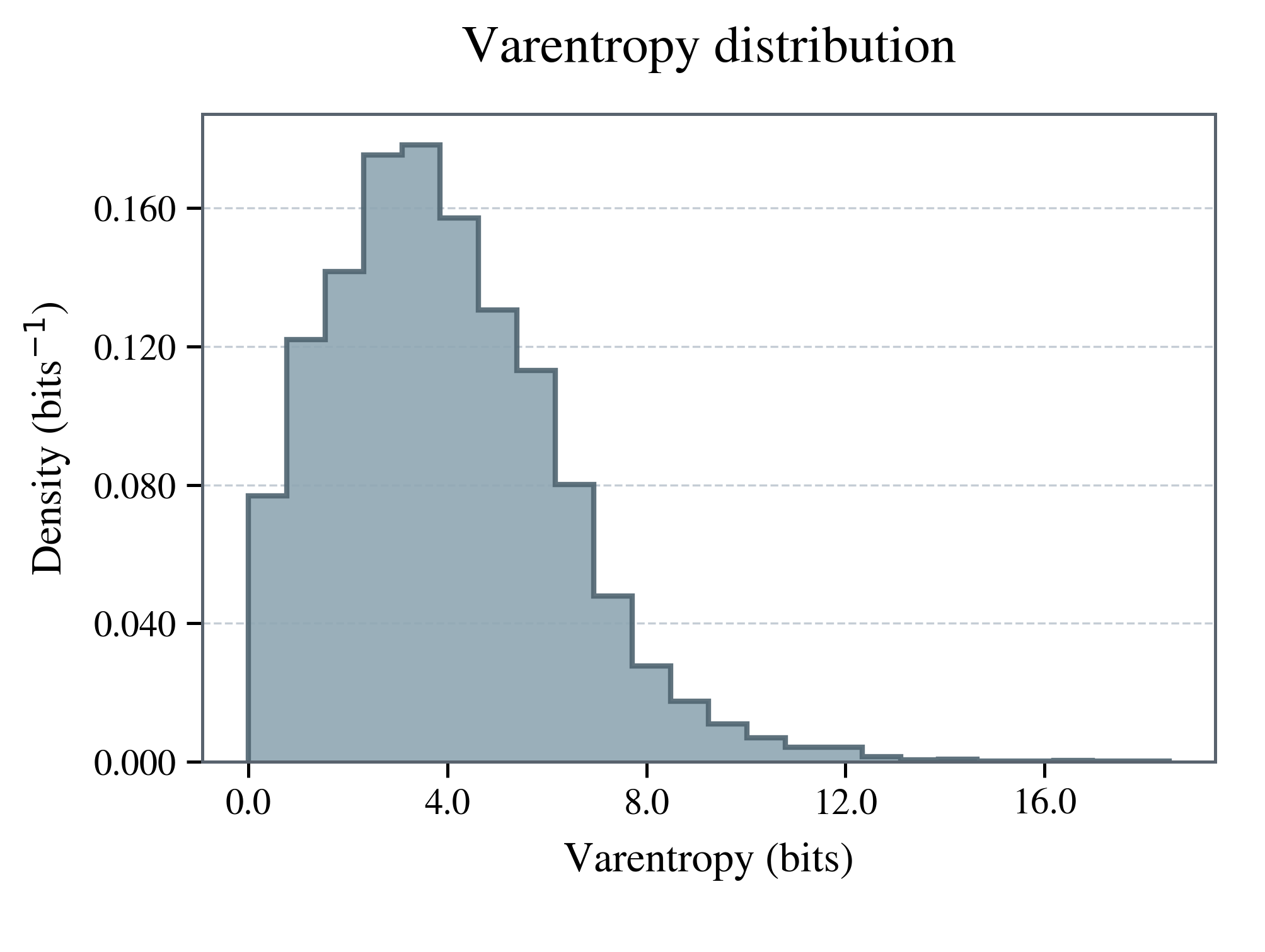} \\
Recipe 8 &
\includegraphics[width=0.145\textwidth]{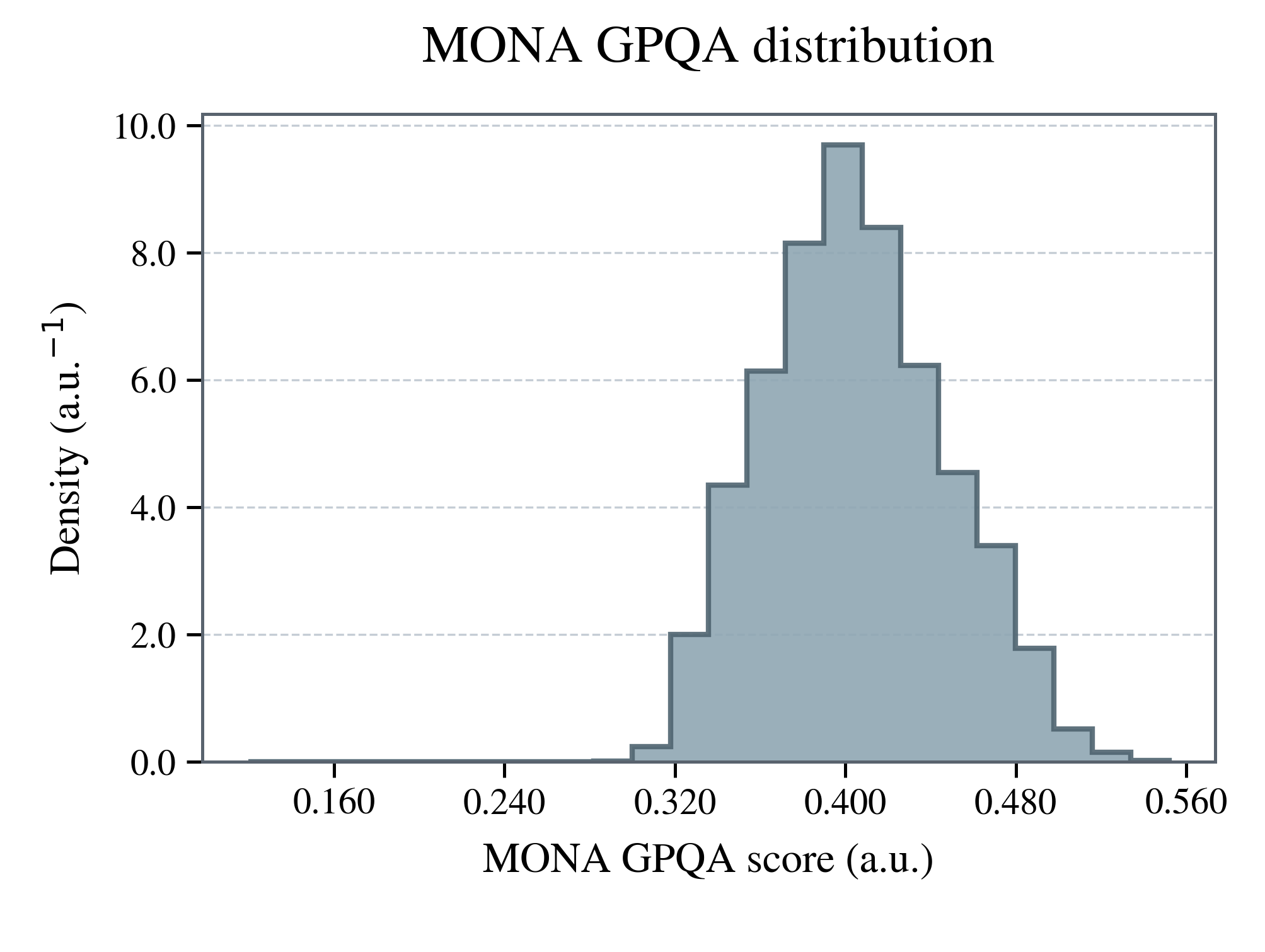} &
\includegraphics[width=0.145\textwidth]{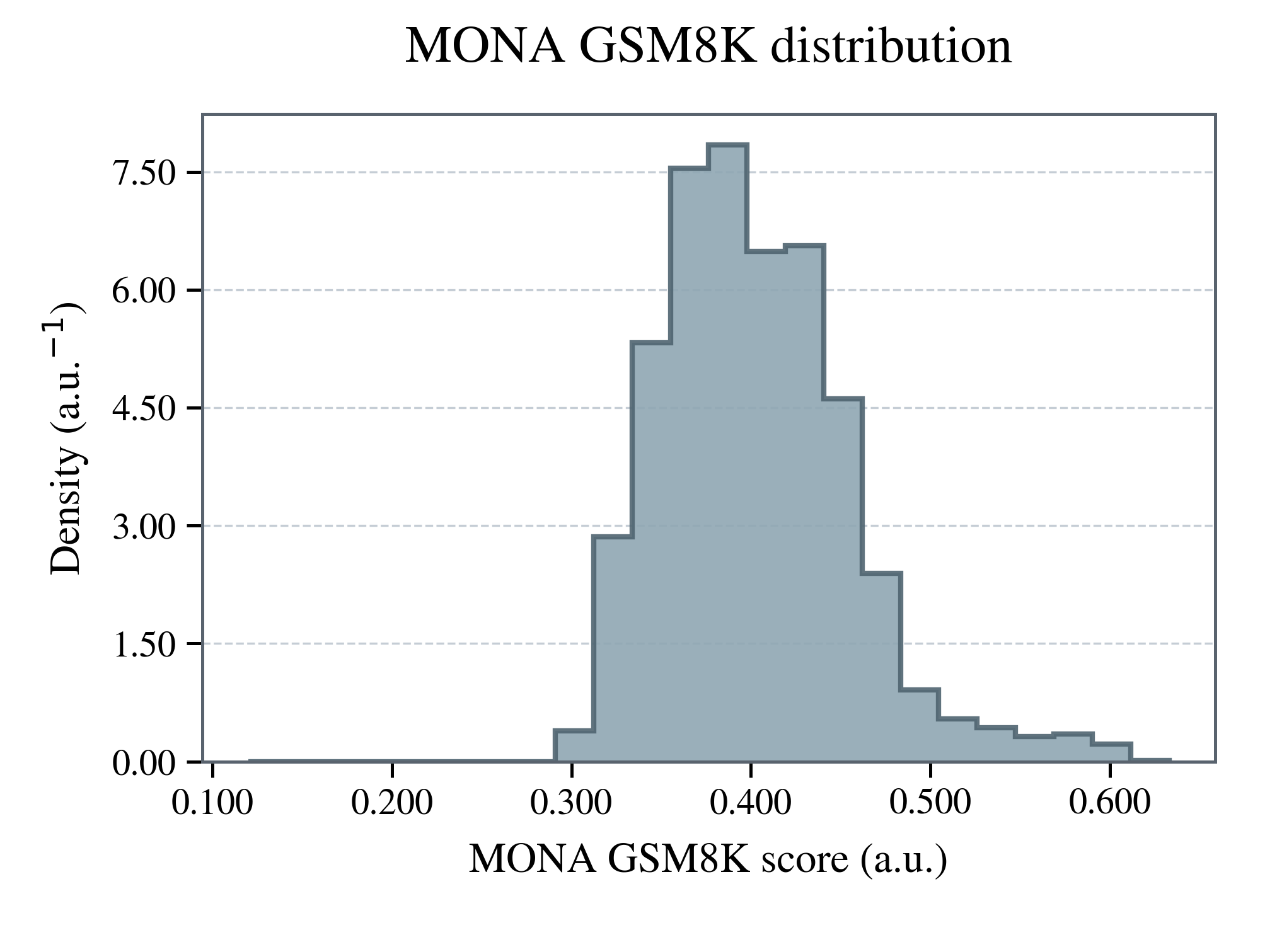} &
\includegraphics[width=0.145\textwidth]{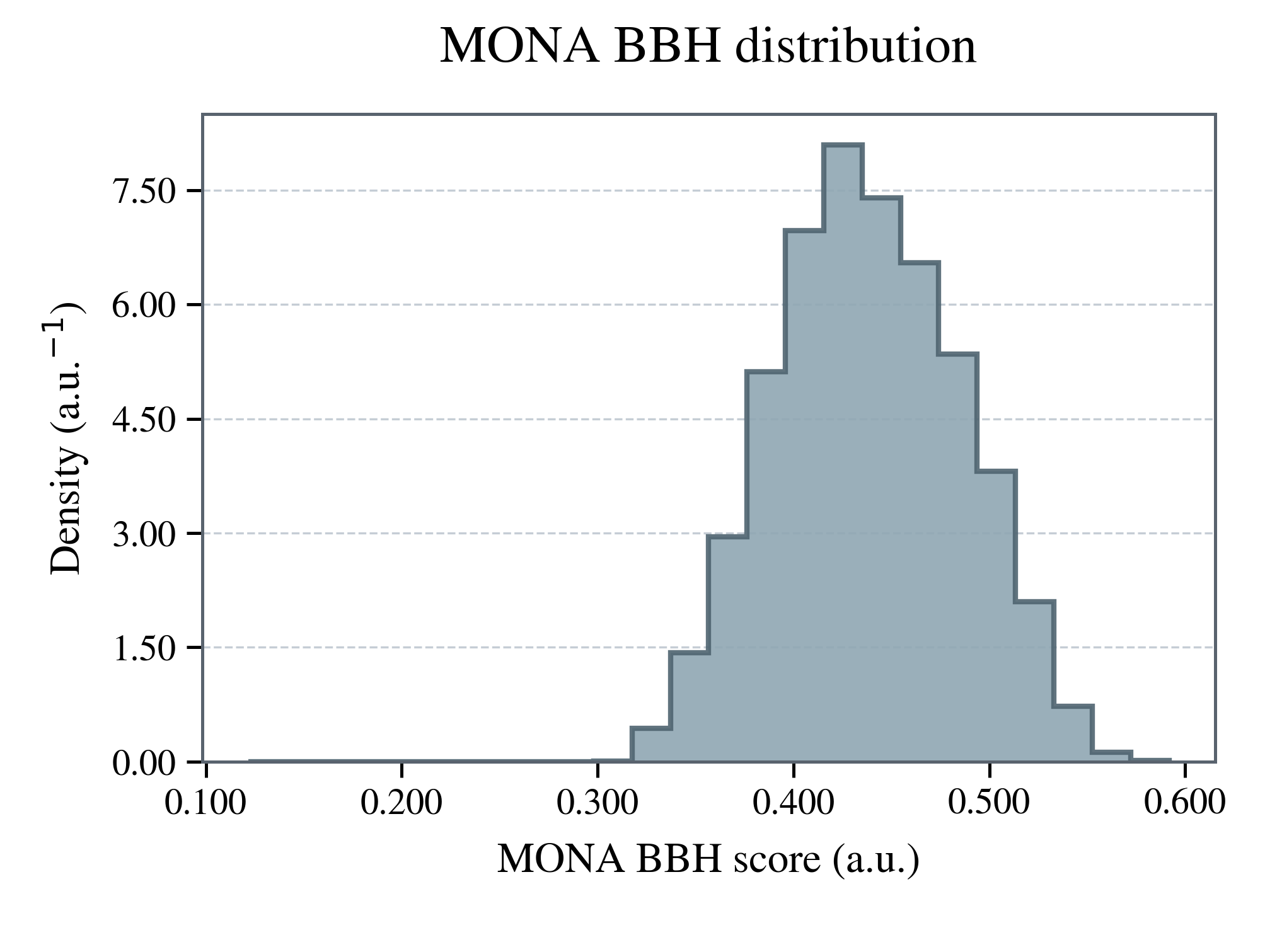} &
\includegraphics[width=0.145\textwidth]{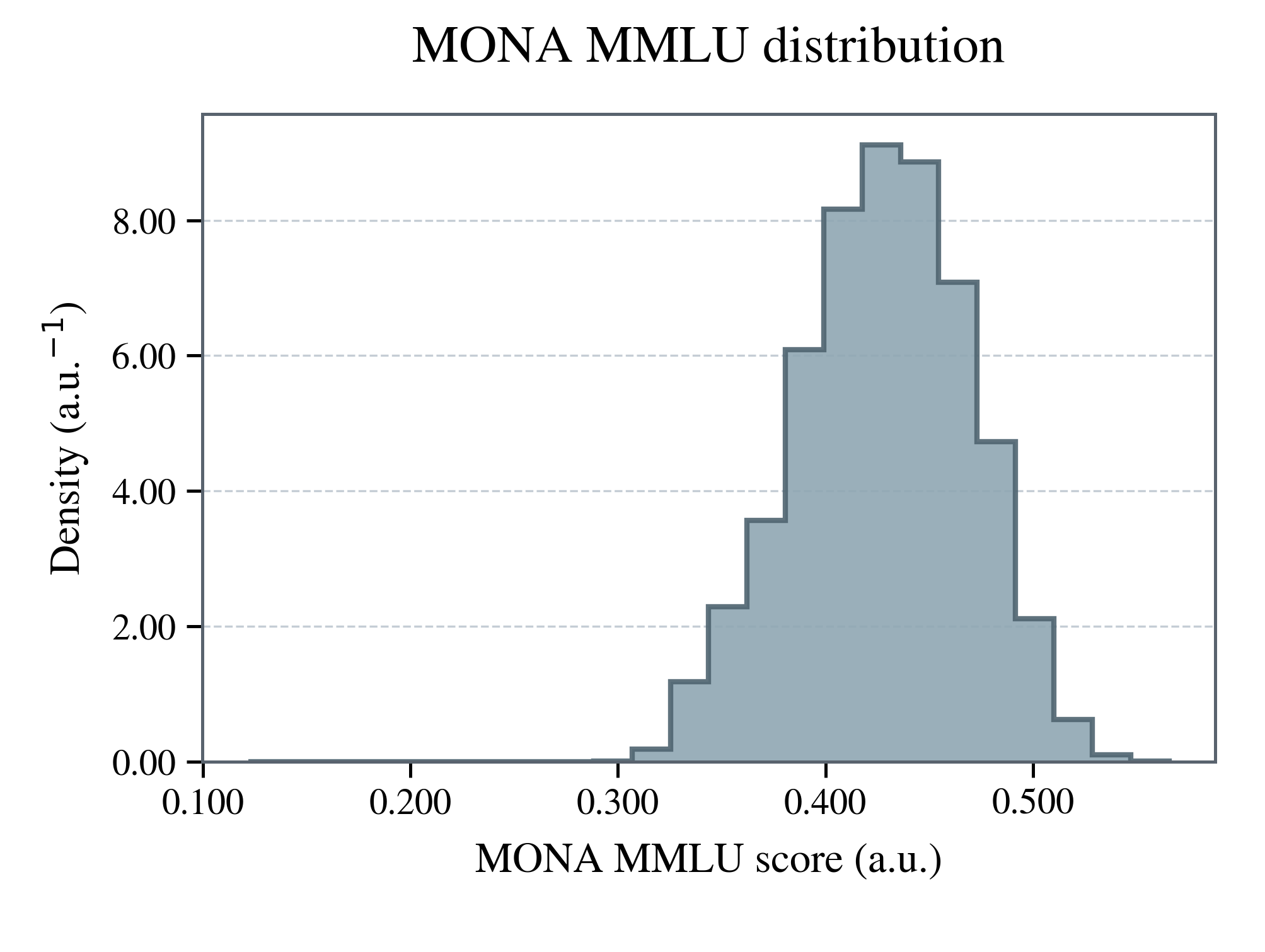} &
\includegraphics[width=0.145\textwidth]{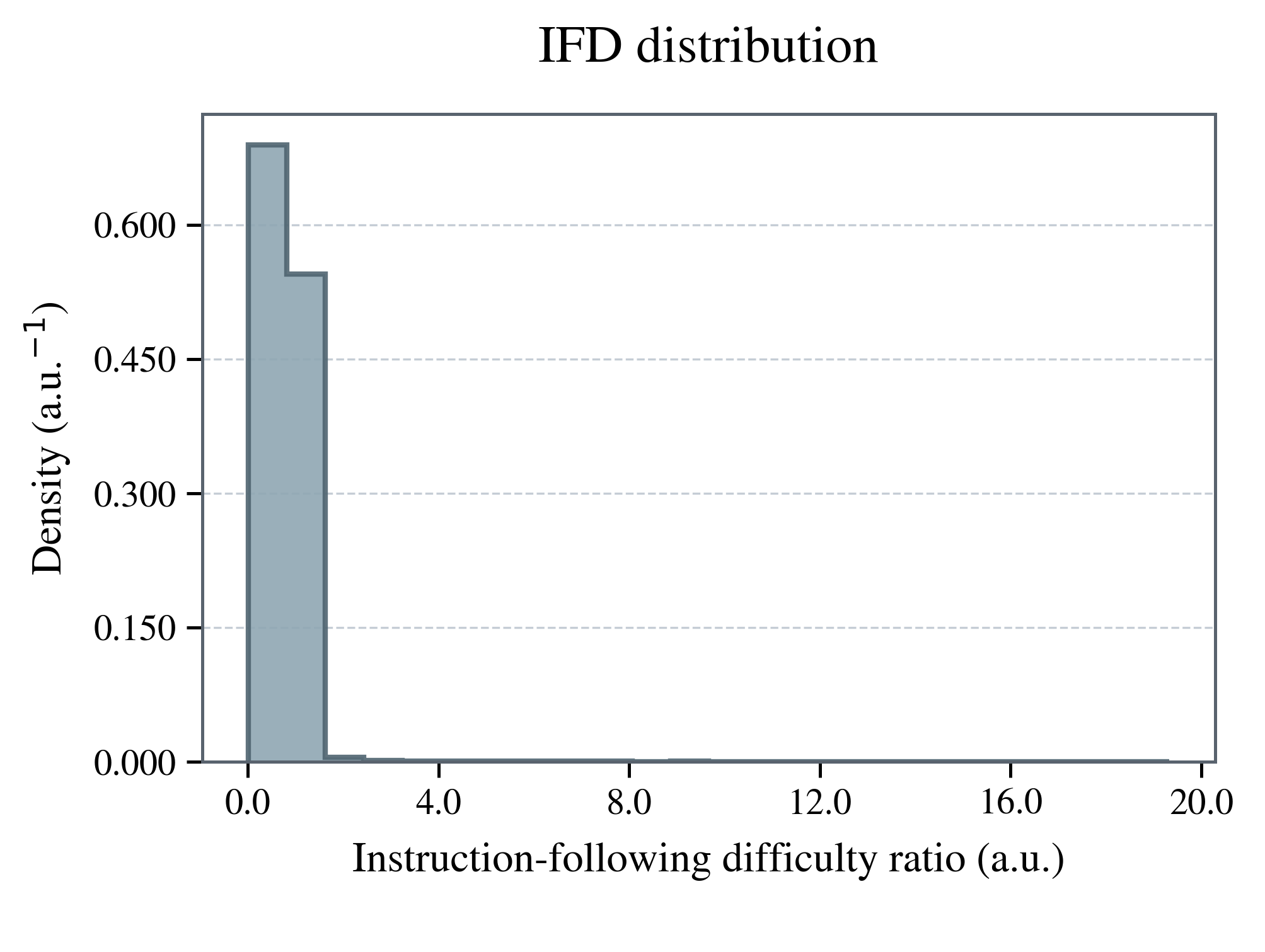} &
\includegraphics[width=0.145\textwidth]{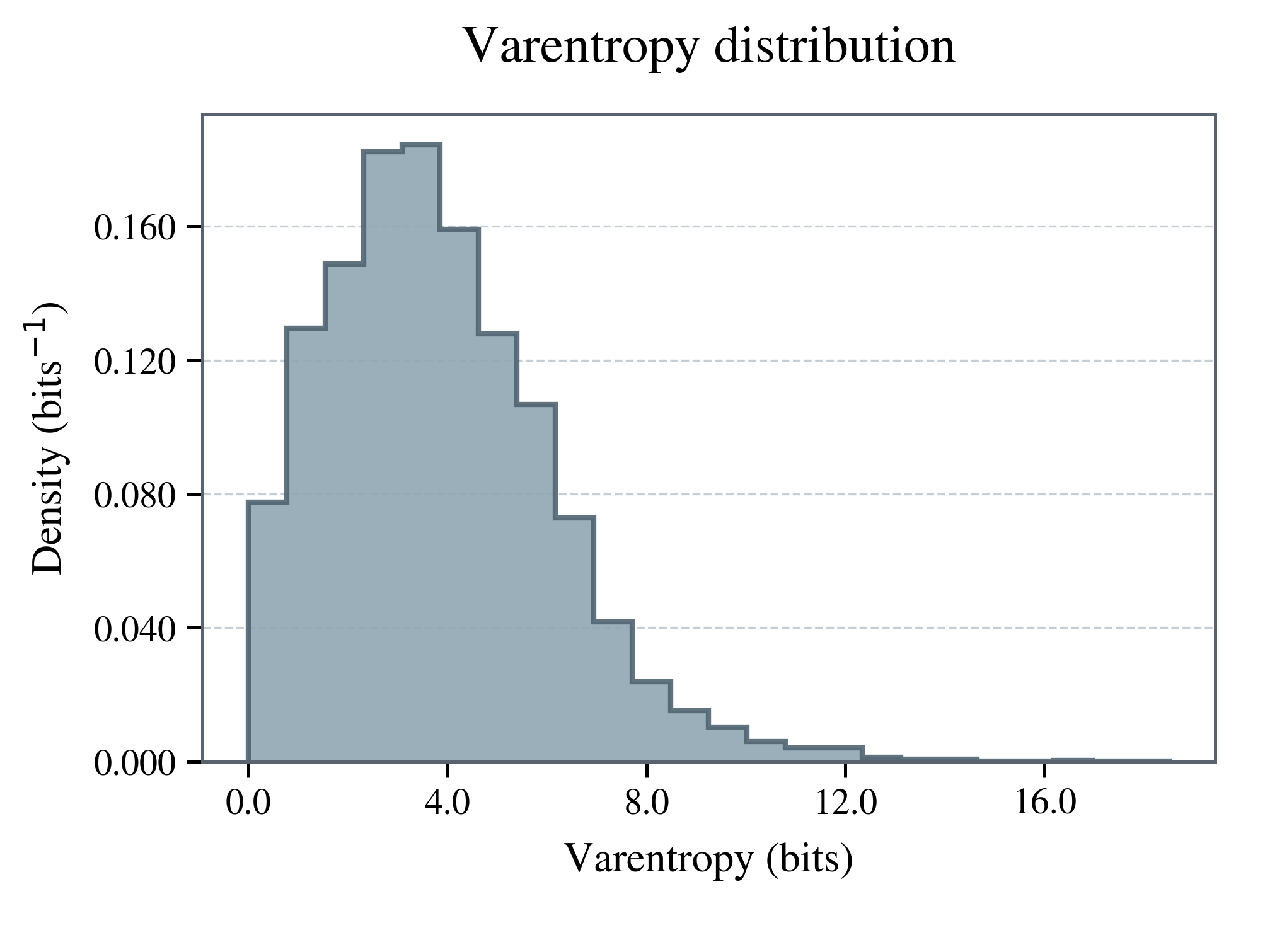} \\
Recipe 9 &
\includegraphics[width=0.145\textwidth]{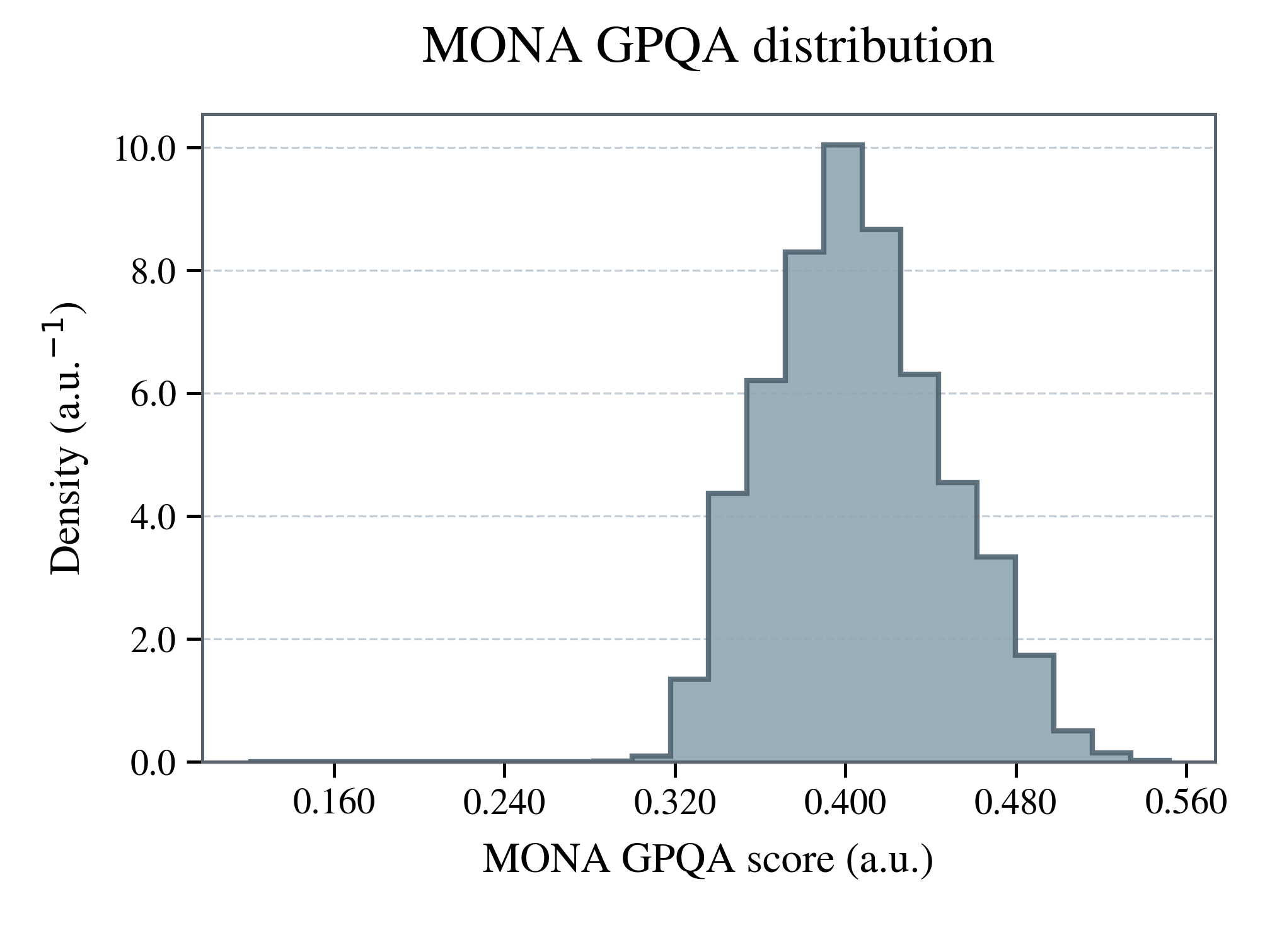} &
\includegraphics[width=0.145\textwidth]{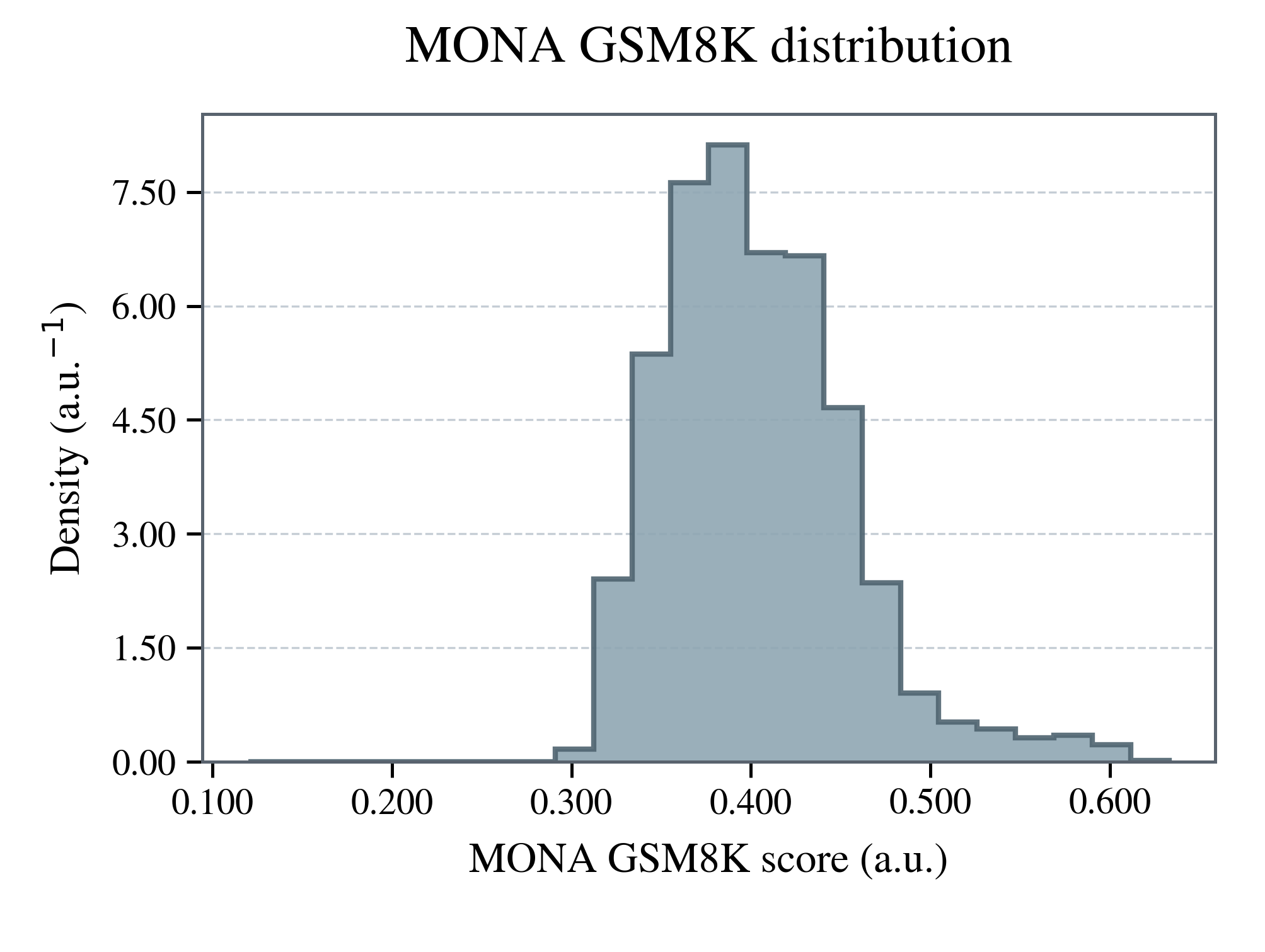} &
\includegraphics[width=0.145\textwidth]{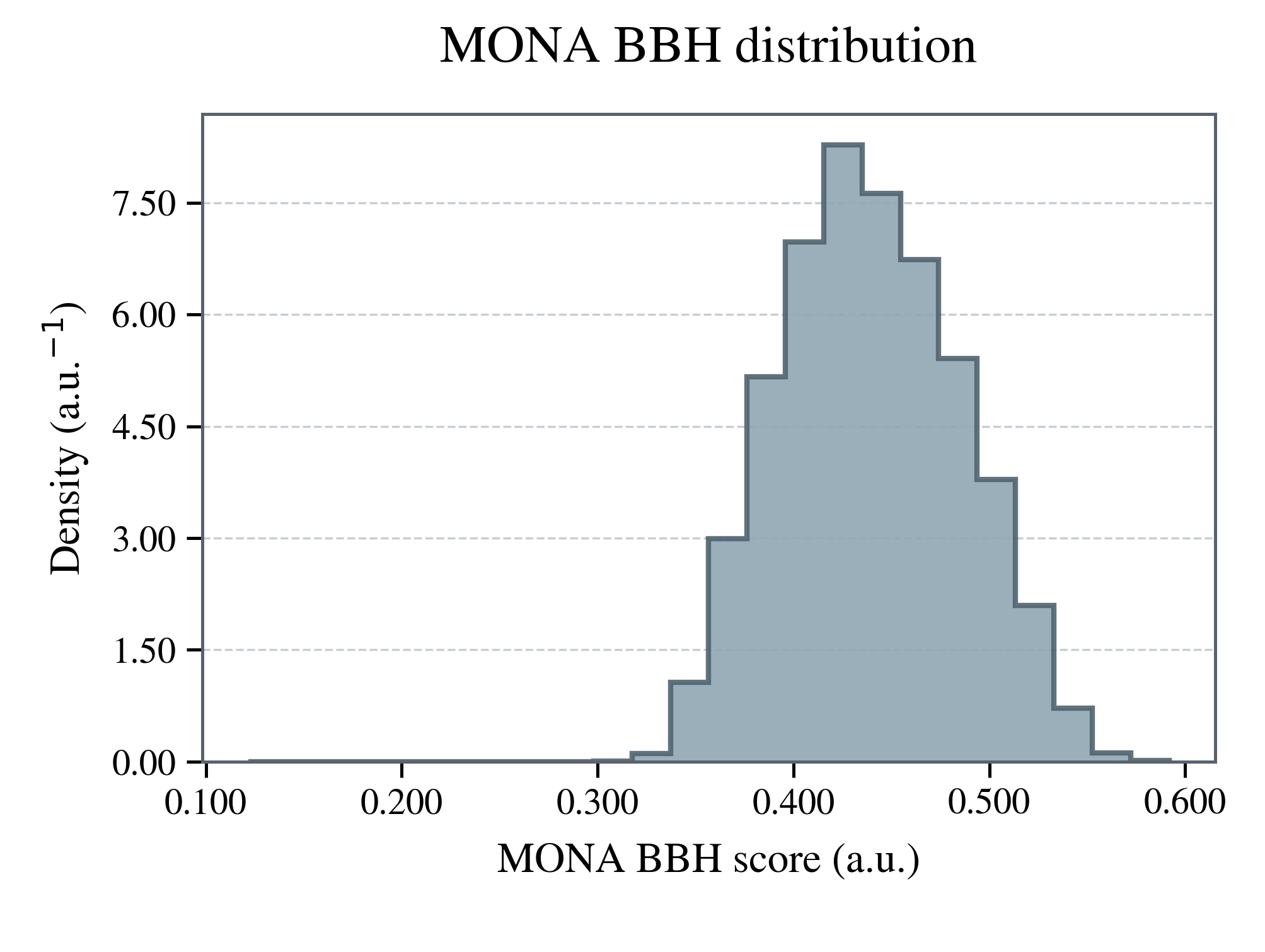} &
\includegraphics[width=0.145\textwidth]{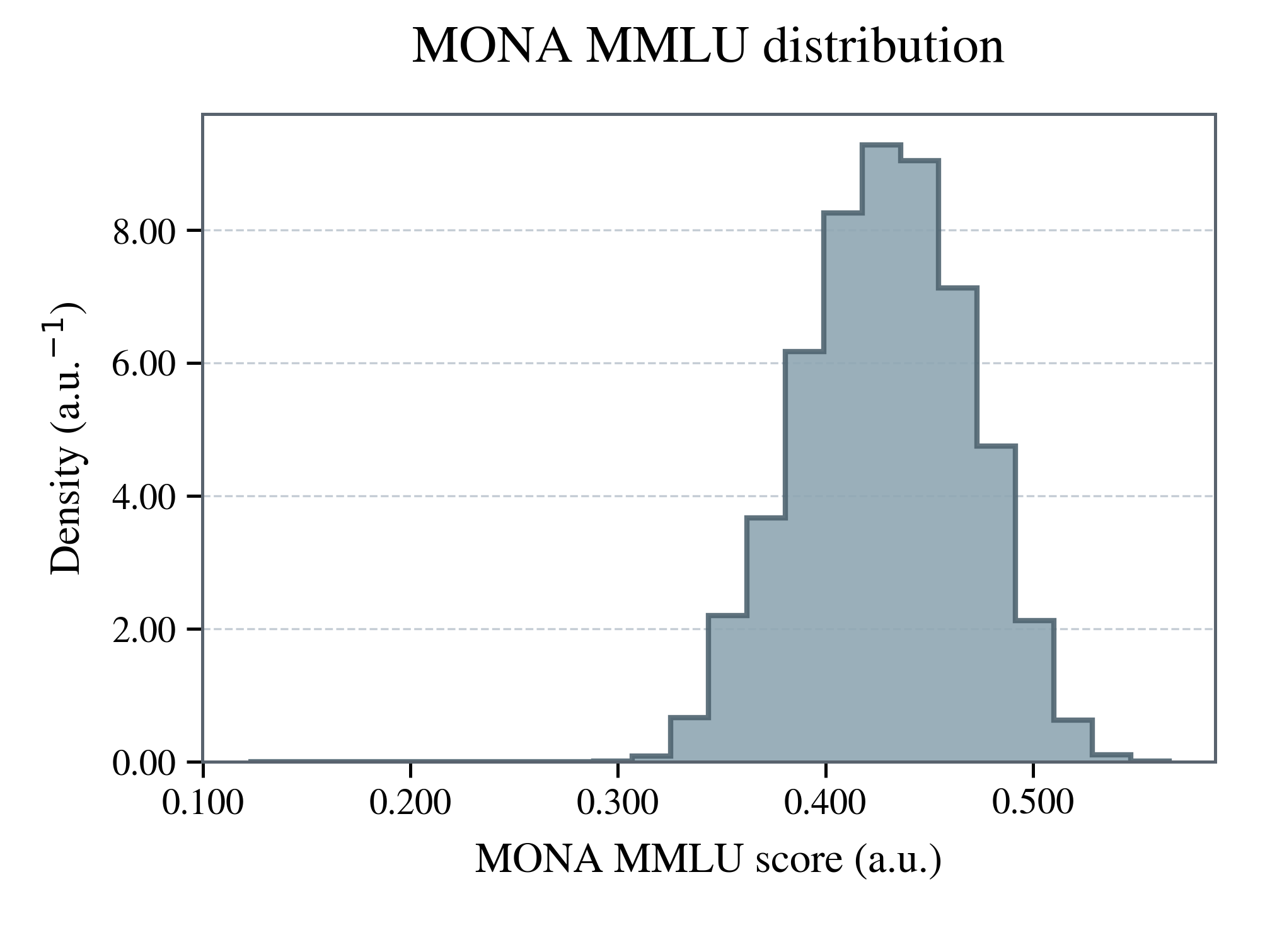} &
\includegraphics[width=0.145\textwidth]{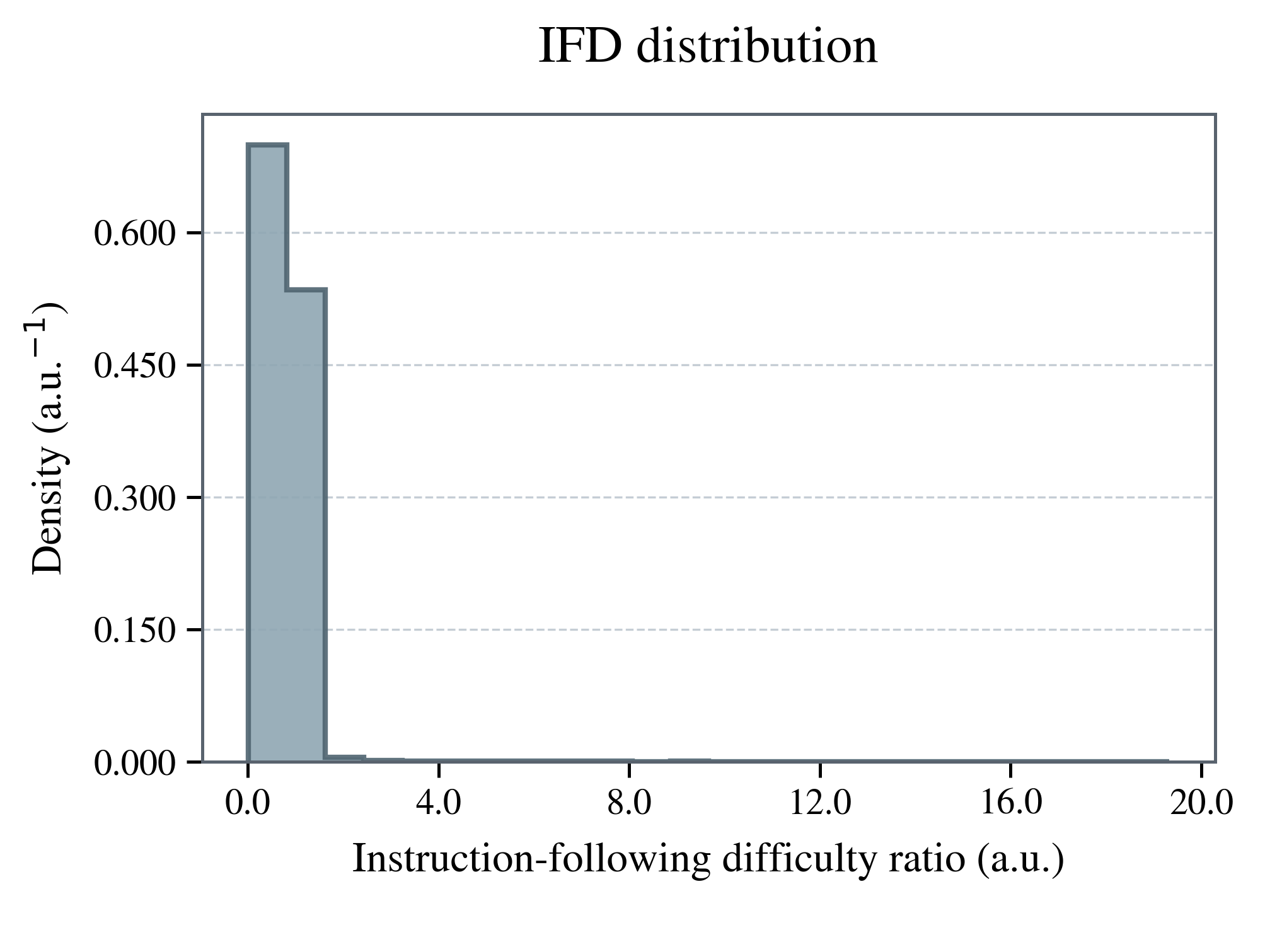} &
\includegraphics[width=0.145\textwidth]{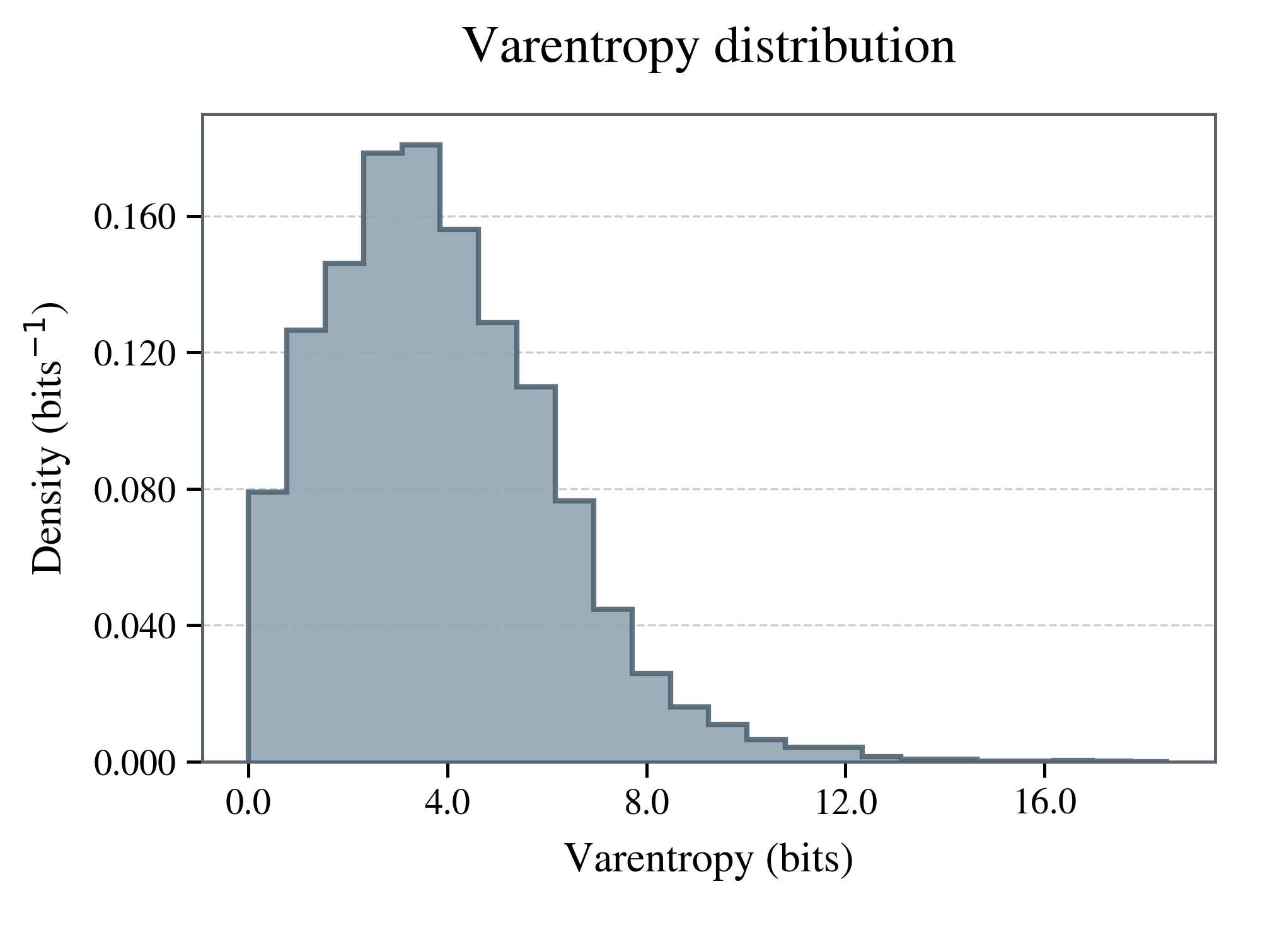} \\
\end{tabular}
\caption{Qualitative metric-distribution comparison for nine anonymous evaluated recipes. Each row corresponds to one recipe, and the six panels in the row correspond to subset-level metric views. Aggregate scores for the same recipes are reported in Table~\ref{tab:recipe_metric_distribution_scores}.}
\label{fig:recipe_metric_distributions}
\end{figure*}

Figure~\ref{fig:recipe_metric_distributions} compares nine evaluated recipes from the same 1.5B search process. Each row corresponds to one evaluated recipe, and the six panels report the marginal distributions of MONA-style relevance scores for GPQA, GSM8K, BBH, and MMLU, followed by IFD and entropy. The purpose of this analysis is not to identify a single best metric, but to examine whether individual metric distributions are sufficient to explain the downstream performance of a selected subset. Table~\ref{tab:recipe_metric_distribution_scores} reports the corresponding aggregate scores for these recipes.

The main observation is that several evaluated recipes exhibit highly similar marginal distributions but obtain different aggregate scores after full evaluation. In the MONA-based columns, the GPQA, GSM8K, BBH, and MMLU distributions are all concentrated in narrow ranges and mostly preserve similar unimodal shapes across recipes. The IFD distributions are also strongly concentrated near the lower end of the axis for nearly all recipes. Even entropy, which shows more variation in tail shape than the other metrics, does not provide a clear separation among all recipes. These patterns indicate that data can look similar under individual selection signals while still leading to different downstream outcomes.

The distributional evidence supports the central design choice of AutoSelection: data selection should be treated as multi-view selection rather than single-score filtering. Since similar one-dimensional distributions can correspond to different downstream performance, the search process needs to consider how operators, thresholds, and ordering jointly shape the selected subset. Full evaluation remains necessary because it observes the actual effect of a recipe after fine-tuning, while the metric distributions serve as contextual signals that guide exploration and exploitation under a limited evaluation budget.

\begin{table}[h]
\centering
\caption{Aggregate scores for the nine anonymous recipes visualized in the metric-distribution figure.}
\label{tab:recipe_metric_distribution_scores}
\begin{tabular}{lrrr}
\toprule
Recipe & Iteration & Retained examples & Score \\
\midrule
Recipe 1 & 3 & 9,136 & 32.75 \\
Recipe 2 & 9 & 40,471 & 37.70 \\
Recipe 3 & 19 & 22,326 & 38.05 \\
Recipe 4 & 4 & 63,252 & 38.36 \\
Recipe 5 & 7 & 24,085 & 38.72 \\
Recipe 6 & 13 & 66,748 & 39.88 \\
Recipe 7 & 17 & 75,044 & 40.48 \\
Recipe 8 & 16 & 73,716 & 41.06 \\
Recipe 9 & 15 & 69,634 & 42.28 \\
\bottomrule
\end{tabular}
\end{table}

\newpage
\clearpage
\section{Prompt Templates Used in Search and Evaluation}
\label{app:prompt_templates}

This appendix reports the runtime prompt templates used by AutoSelection.

\subsection{Search-agent prompts}
\label{app:search_agent_prompts}

The search loop uses four LLM-facing prompt templates.
The Summarizer converts verified history into concise search guidance, the Proposer generates a candidate pool from the current recipe and guidance, the Ranker chooses among surrogate-ranked candidates, and the Reseeder retunes restart parameters after stagnation.

\begin{lstlisting}[style=promptstyle,caption={Summarizer prompt template.},label={lst:prompt_summarizer}]
You are a data science evaluation analyst. Analyze the following evaluated-recipe history from an automated data selection search system.

The system is searching for the best data recipe to train an LLM. Each row is one evaluated recipe: the recipe produced a selected subset, then a model was trained and evaluated.

{experiment_history_table}

KEY CONTEXT:
- Higher scores are better (aggregated accuracy across benchmarks)
- "Operators" are data filtering/mixing steps applied sequentially
- "Samples" is the number of training samples after filtering
- The pool has {estimated_total_pool_size} total samples

TASK: Produce exactly 3-5 concise, actionable findings. Each finding should be:
1. A specific observation (not vague)
2. Backed by data from the table
3. Actionable (suggests what to try or avoid)

Format each finding as a numbered line. Be direct and quantitative.
These are HYPOTHESES based on limited data, not proven facts.

Example format:
1. More data consistently helps: recipe_A (12K samples, 22.2%) > recipe_C (3K samples, 18.5%). Avoid aggressive filtering.
2. operator_X at rate 0.3 hurts benchmark_Y: recipe_B dropped from 15% to 0.9%. Try higher rates or skip it.
\end{lstlisting}

\begin{lstlisting}[style=promptstyle,caption={Proposer prompt template.},label={lst:prompt_proposer}]
You are an expert Data-Centric AI Search Controller optimizing a data recipe.

YOUR GOAL: Propose {n_candidates} DISTINCT mutated recipe configurations that resolve current risks and explore different valid subspaces. Some should be conservative, some more aggressive.

{operator_catalog}
{registered_operator_note}

=== CURRENT STATE ===
Current Recipe:
{current_recipe_steps}

Current Metric Score: {score}

{state_vector_section}
{benchmark_diagnostic_section}
{pool_context_section}
{search_history_section}
{experiment_insights_section}
{union_operator_section}

=== INSTRUCTIONS ===
1. Analyze the current recipe, state vector, and search history.
2. Select operators and hyperparameters ONLY from the OPERATOR CATALOG.
3. Your output MUST be a valid JSON array of objects representing the {n_candidates} recipes. Do NOT include markdown blocks (` ```json `), just raw JSON.
4. Format:
[
  {
    "steps": [
      {
        "operator": "operator_name",
        "params": {"param1": "value", "param2": 123}
      }
    ]
  },
  ... (up to {n_candidates} distinct configurations)
]
\end{lstlisting}

\begin{lstlisting}[style=promptstyle,caption={Ranker prompt template.},label={lst:prompt_ranker}]
You are a strategic advisor for an automated data selection search system. Your task is to select the SINGLE most promising candidate recipe for real evaluation.

## SEARCH STATE
- Total data pool: {pool_size} samples
- Iterations completed: {n_iterations}
- Budget remaining: {budget_remaining}h / {budget_total}h ({budget_pct}%)
- Current best score: {best_score}% (recipe: {best_name})
- Search phase: {phase}

{parent_section}
{detailed_experiment_history_section}
{experiment_insights_section}

## CANDIDATES (ranked by GP surrogate)
NOTE: The GP surrogate predicts expected utility from an 11-D recipe encoding (operator presence + parameters).
Each candidate's pipeline has been pre-executed to obtain data state metrics (shown below for reference).

{candidate_table_with_gp_and_state_vectors}

## SELECTION CRITERIA
Consider these factors carefully:

1. Per-Task MONA Scores (PRIMARY SIGNAL):
   - score_per_task shows how relevant the selected subset is to each benchmark.
   - A candidate whose per-task MONA scores improve across multiple benchmarks is a strong positive signal, even if retain_ratio drops.
   - Compare each candidate's score_per_task against the parent's and look for improvements on weak benchmarks.
   - If a candidate improves some benchmarks but hurts others, weigh the magnitude and importance of each.
   - score_mean is the aggregate; score_per_task is the breakdown. Always prioritize the per-task view.

2. Exploration vs Exploitation Trade-off:
   - Early search: prefer high sigma candidates to gather information.
   - Late search: prefer high mu candidates to refine the best.
   - Current phase: {phase}.

3. Data Quantity Risk:
   - Recipes that aggressively filter data risk producing too few samples.
   - Historical evidence shows extreme filtering often fails catastrophically.
   - Union operators can recover data volume and are safer exploration choices.
   - Refer to the per-benchmark history to see how sample count correlates with each benchmark.

4. Operator Synergies and Redundancy:
   - Multiple filtering operators in sequence compound data loss multiplicatively.
   - Operators from the same family are often redundant.
   - Complementary operators tend to work well together.

5. Feedback Alignment:
   - Does this candidate address the patterns identified in evaluation insights?
   - Does it avoid strategies that have been shown to fail?

6. State Vector Patterns:
   - High retain_ratio with good score_mean tends to perform well.
   - High distribution_drift indicates risky distributional shift.
   - The parent's state vector shows the data profile that candidates will modify.

7. GP Model Limitations:
   - The GP has only {n_iterations} training points, so predictions carry uncertainty.
   - Do not blindly trust GP rankings, especially when scores are close.
   - Qualitative reasoning about operator interactions can add value beyond the GP.

## OUTPUT FORMAT
After thorough reasoning, output a full ranking of all presented candidates as a JSON object.
The ranking list must contain ALL candidate indices (0-based) sorted from most promising to least:

{
  "ranking": [<best_idx>, <2nd_idx>, ..., <worst_idx>],
  "confidence": "<high|medium|low>",
  "rationale": "<one-sentence explanation of why your top choice was chosen>",
  "eval_rationale": "<one-sentence reason for eval decision>"
}

Think carefully before answering. Consider each candidate's strengths and risks.
\end{lstlisting}

\begin{lstlisting}[style=promptstyle,caption={Reseeder prompt template.},label={lst:prompt_reseeder}]
You are choosing a restart operator motif for recipe search.

Use the search evidence below to select a small restart motif that is promising.
You must choose between 1 and 3 NON-TRUNCATE operators.

[OPERATOR_CATALOG]

=== SEARCH HISTORY ===
[SEARCH_HISTORY]

=== POSITIVE OPERATOR SIGNALS ===
[POSITIVE_OPERATOR_SIGNALS_JSON]

=== POSITIVE PAIR SIGNALS ===
[POSITIVE_PAIR_SIGNALS_JSON]

=== HISTORICAL SUCCESSFUL EXAMPLES ===
[HISTORICAL_SUCCESSFUL_EXAMPLES_JSON]

=== RULES ===
1. Select only operators from the allowed catalog above.
2. Prefer operators and combinations supported by the evidence.
3. Keep parameters within catalog bounds.
4. Return raw JSON only in this exact format:
[
  {"operator": "mona_filter", "params": {"fraction": 0.5}},
  {"operator": "ngram_entropy", "params": {"fraction": 0.4}}
]

Pool size reference: [POOL_SIZE]
\end{lstlisting}

\subsection{Benchmark evaluation prompts}
\label{app:evaluation_prompts}

The validation suite uses GPQA, GSM8K, BBH, and MMLU.

\begin{lstlisting}[style=promptstyle,caption={GPQA evaluation prompt scaffold.},label={lst:prompt_gpqa}]
System:
You are an expert assistant. Answer the following multiple choice question by selecting the correct option (A, B, C, or D).

Instructions:
1. Read the question carefully.
2. Consider each option.
3. Respond with ONLY the letter of the correct answer (A, B, C, or D) on the last line.
4. Format: put your final answer after "Answer:" on the last line.

Example format:
[Your reasoning]
Answer: B

Few-shot turns:
User: Question: Which of the following is NOT a function of the cell membrane?
A. Selective permeability
B. Protein synthesis
C. Cell signaling
D. Cell adhesion
Assistant: The cell membrane has multiple functions including selective permeability, cell signaling, and cell adhesion. Protein synthesis occurs at ribosomes, not at the cell membrane.
Answer: B

User: Question: What is the SI unit of electrical resistance?
A. Volt
B. Ampere
C. Ohm
D. Watt
Assistant: Electrical resistance is measured in Ohms (Omega), named after Georg Ohm. Volts measure potential difference, Amperes measure current, and Watts measure power.
Answer: C

User: Question: Which element has the highest electronegativity?
A. Oxygen
B. Nitrogen
C. Fluorine
D. Chlorine
Assistant: Fluorine has the highest electronegativity of all elements (3.98 on the Pauling scale). It is the most electronegative element because it has a small atomic radius and high effective nuclear charge.
Answer: C

Final user turn:
Question: {question}
A. {option_A}
B. {option_B}
C. {option_C}
D. {option_D}
\end{lstlisting}

\begin{lstlisting}[style=promptstyle,caption={GSM8K evaluation prompt scaffold.},label={lst:prompt_gsm8k}]
System:
You are a helpful math assistant. Solve the following math problem step by step. Show your work, then provide the final answer after '#### ' on the last line.

Few-shot turns:
User: Janet's ducks lay 16 eggs per day. She eats three for breakfast every morning and bakes muffins for her friends every day with four. She sells every duck egg at the farmers' market daily for $2. How much in dollars does she make every day at the farmers' market?
Assistant: Janet's ducks lay 16 eggs per day. She eats 3 for breakfast and uses 4 for muffins, so she uses 3 + 4 = 7 eggs. That leaves 16 - 7 = 9 eggs to sell. At $2 each, she makes 9 x $2 = $18.
#### 18

User: A robe takes 2 bolts of blue fiber and half that much white fiber. How many bolts in total does it take?
Assistant: The robe takes 2 bolts of blue fiber. It takes half that much white fiber, so 2 / 2 = 1 bolt of white fiber. In total, it takes 2 + 1 = 3 bolts.
#### 3

User: Josh decides to try flipping a house. He buys a house for $80,000 and puts $50,000 in repairs. This increased the value of the house by 150%. How much profit did he make?
Assistant: Josh spent $80,000 + $50,000 = $130,000 total. The repairs increased the value by 150%, so the increase is $80,000 x 1.5 = $120,000. The new value is $80,000 + $120,000 = $200,000. His profit is $200,000 - $130,000 = $70,000.
#### 70000

Final user turn:
{question}
\end{lstlisting}

\begin{lstlisting}[style=promptstyle,caption={BBH evaluation prompt scaffold.},label={lst:prompt_bbh}]
System:
You are an expert reasoning assistant. Answer the following question step by step, then provide your final answer.

Instructions:
1. Think through the problem carefully.
2. Show your reasoning.
3. On the last line, write your final answer after "Answer:" exactly matching the expected format.

Example format:
[Your reasoning]
Answer: (B)

Few-shot turns:
User: not ( True ) and ( True ) is
Assistant: not ( True ) evaluates to False. False and ( True ) evaluates to False.
Answer: False

User: In the following sentences, explain the antecedent of the pronoun.
Sentence: The nurse notified the patient that his shift would be ending in an hour.
Options:
(A) The nurse's shift
(B) The patient's shift
(C) Ambiguous
Assistant: The pronoun 'his' could refer to either the nurse or the patient. However, given the context of notifying about a shift ending, 'his' most likely refers to the nurse's shift.
Answer: (A)

User: If we list all the natural numbers below 10 that are multiples of 3 or 5, we get 3, 5, 6 and 9. The sum of these multiples is 23.
Find the sum of all the multiples of 3 or 5 below 20.
Assistant: Multiples of 3 below 20: 3, 6, 9, 12, 15, 18
Multiples of 5 below 20: 5, 10, 15
Combined (no duplicates): 3, 5, 6, 9, 10, 12, 15, 18
Sum = 3+5+6+9+10+12+15+18 = 78
Answer: 78

Final user turn:
{input}
\end{lstlisting}
\begin{lstlisting}[style=promptstyle,caption={MMLU evaluation prompt scaffold.},label={lst:prompt_mmlu}]
System:
You are a knowledgeable assistant. Answer the following multiple choice question by selecting the correct option (A, B, C, or D).

Instructions:
1. Read the question carefully.
2. Consider each option.
3. Respond with your reasoning, then provide the letter of the correct answer after "Answer:" on the last line.

Example format:
[Your reasoning]
Answer: B

Few-shot turns:
User: Question: What is the capital of France?
A. London
B. Berlin
C. Paris
D. Madrid
Assistant: Paris is the capital and largest city of France.
Answer: C

User: Question: Which planet is known as the Red Planet?
A. Venus
B. Mars
C. Jupiter
D. Saturn
Assistant: Mars is commonly known as the Red Planet due to its reddish appearance caused by iron oxide on its surface.
Answer: B

User: Question: What is the powerhouse of the cell?
A. Nucleus
B. Ribosome
C. Mitochondria
D. Golgi apparatus
Assistant: Mitochondria are often called the 'powerhouse of the cell' because they generate most of the cell's supply of ATP, the main energy currency.
Answer: C

Final user turn:
Question: {question}
A. {choice_A}
B. {choice_B}
C. {choice_C}
D. {choice_D}
\end{lstlisting}
\begin{lstlisting}[style=promptstyle,caption={GraphWiz and NLGraph evaluation prompt scaffold.},label={lst:prompt_graph}]
You are a graph reasoning assistant.

Solve the graph problem carefully. On the last line, output only:
Answer: Yes
or
Answer: No

Final user turn:
{question}
\end{lstlisting}

\end{document}